\documentclass[11pt]{article}

\usepackage{etex}
\usepackage{amsmath}
\usepackage{amsfonts}
\usepackage{amsthm}
\usepackage{array}
\usepackage{multirow}
\usepackage{float}
\usepackage{hyperref}
\usepackage[mathscr]{euscript}
\usepackage{subcaption}
\usepackage{bookmark}
\usepackage{url}
\usepackage{algorithm}
\usepackage{algpseudocode}
\usepackage{natbib}
\usepackage{nicefrac}       
\usepackage{microtype}      
\usepackage{graphicx} 
\usepackage{amsthm}
\usepackage{amssymb}

\usepackage{xcolor}
\usepackage{color}
\usepackage{pdfpages,caption}
\newcommand{\be}[1]{\begin{equation}\label{#1}}
\newcommand{\benon}{\begin{equation*}}  
\newcommand{\ee}{\end{equation}}
\newcommand{\eenon}{\end{equation*}}
\newcommand{\bemuln}[1]{\begin{multline}\label{#1}}
\newcommand{\bemul}{\begin{multline*}}
\newcommand{\emuln}{\end{multline}}
\newcommand{\emul}{\end{multline*}}
\newcommand{\bee}{\begin{eqnarray*}}
\newcommand{\eee}{\end{eqnarray*}}
\newcommand{\been}[1]{\begin{eqnarray}\label{#1}}
\newcommand{\eeen}{\end{eqnarray}}
\newcommand{\began}[1]{\begin{gather}\label{#1}}
\newcommand{\bega}{\begin{gather*}}
\newcommand{\egan}{\end{gather}}
\newcommand{\ega}{\end{gather*}}
\newcommand{\bealn}[1]{\begin{align}\label{#1}}
\newcommand{\beal}{\begin{align*}}
\newcommand{\ealn}{\end{align}}
\newcommand{\eal}{\end{align*}}
\newcommand{\bealatn}[2]{\begin{alignat}{#1}\label{#2}}
\newcommand{\bealat}{\begin{alignat*}}
\newcommand{\ealatn}{\end{alignat}}
\newcommand{\ealat}{\end{alignat*}}
\newcommand{\bexalatn}[1]{\begin{xalignat}\label{#1}}
\newcommand{\bexalat}{\begin{xalignat*}}
\newcommand{\exalatn}{\end{xalignat}}
\newcommand{\exalat}{\end{xalignat*}}
\newcommand{\mbb}{\mathbb}

	\newtheorem{theorem}{Theorem}[section]

	\newtheorem{lem}[theorem]{Lemma}
	\newtheorem{col}[theorem]{Corollary}
	\newtheorem{defi}{Definition}
	\newtheorem{ass}{Assumption} 
	 
\def\ba{{\mathbf a}}
\def\bb{{\mathbf b}}

\def\bg{{\mathbf g}}

\def\bu{{\mathbf u}}
\def\bv{{\mathbf v}}
\def\bw{{\mathbf w}}
\def\bx{{\mathbf x}}  
\def\by{{\mathbf y}}
\def\bz{{\mathbf z}}

\def\bI{{\mathbf I}}

\def\bM{{\mathbf M}}

\def\bX{{\mathbf X}}

\def\bZ{{\mathbf Z}}

\def\texitem#1{\par\smallskip\noindent\hangindent 25pt
               \hbox to 25pt {\hss #1 ~}\ignorespaces}

\newcommand{\bzero}{{\mathbf{0}}}

\newcommand{\scrA}{\mathcal{A}}
\newcommand{\scrB}{\mathcal{B}}

\newcommand{\scrF}{\mathcal{F}}

\newcommand{\scrH}{\mathcal{H}}

\newcommand{\scrP}{\mathcal{P}}

\newcommand{\scrR}{\mathcal{R}}

\newcommand{\scrT}{\mathcal{T}}

\newcommand{\scrZ}{\mathcal{Z}}

\newcommand{\bbeta}{\boldsymbol{\beta}}
\newcommand{\bGamma}{\boldsymbol{\Gamma}}

\newcommand{\btheta}{\boldsymbol{\theta}}

\newcommand{\bnu}{{\boldsymbol{\nu}}}

\newcommand{\bSigma}{{\boldsymbol{\Sigma}}}

\newcommand{\vertiii}[1]{{\left\vert\kern-0.25ex\left\vert\kern-0.25ex\left\vert #1 
		\right\vert\kern-0.25ex\right\vert\kern-0.25ex\right\vert}}

\begin{document}

\title{A Robust Learning Algorithm for Regression Models Using Distributionally Robust Optimization}

\author{Ruidi Chen\footnote{Ruidi Chen is with Division of Systems Engineering, Boston University, Boston, MA 02446, USA.
		Email: rchen15@bu.edu.} \ and
	Ioannis Ch. Paschalidis\footnote{Ioannis Ch. Paschalidis is with Dept. of Electrical and Computer Engineering,
		Division of Systems Engineering,
		and Dept. of Biomedical Engineering, Boston University, 8 St. Mary's St.,
		Boston, MA 02215, USA.
		Email: yannisp@bu.edu.}}

\date{}

\maketitle

\begin{abstract}
  We present a {\em Distributionally Robust Optimization (DRO)} approach
  to estimate a robustified regression plane in a linear regression setting, when the observed samples are potentially contaminated with adversarially corrupted outliers. 
  Our approach mitigates the impact of outliers through hedging against a 
  family of distributions on the observed data, some of which assign very low probabilities to the outliers. The set of distributions under consideration are close to the empirical
  distribution in the sense of the Wasserstein metric. We show that this DRO formulation can be relaxed to a convex optimization problem which encompasses a
  class of models. By selecting proper norm spaces for the Wasserstein metric, we are able to recover several commonly used regularized regression models. We provide new insights into the
  regularization term and give guidance on the selection of the
  regularization coefficient from the standpoint of a confidence
  region. We establish two types of performance guarantees for the
  solution to our formulation under mild conditions. One is related to
  its out-of-sample behavior (prediction bias), and the other concerns the discrepancy
  between the estimated and true regression planes (estimation bias). Extensive numerical
  results demonstrate the superiority of our approach to a host of regression models, in terms of the prediction and estimation accuracies. We also consider the application of our robust learning procedure to outlier detection, and show that our approach achieves a much higher AUC (Area Under the ROC Curve) than M-estimation \citep{huber1964robust, huber1973robust}.
\end{abstract}


\section{Introduction} \label{s1}

Consider a linear regression model with response $y \in \mbb{R}$, predictor vector $\bx \in \mbb{R}^{m-1}$, regression coefficient
$\bbeta^* \in \mbb{R}^{m-1}$ and error $\epsilon \in \mbb{R}$:
\begin{equation*}
y = \bx' \bbeta^* + \epsilon.
\end{equation*}
Given samples $(\bx_i, y_i), i=1, \ldots, N$, we are interested in estimating $\bbeta^*$. The {\em Ordinary Least Squares (OLS)} minimizes the sum of squared residuals $\sum_{i=1}^N (y_i - \bx_i' \bbeta)^2$, and works well if all the $N$ samples are generated from the underlying true model. However, when faced with adversarial perturbations in the training data, the OLS estimator will deviate from the true regression plane to accommodate the noise. Alternatively, one can choose to minimize the sum of absolute residuals $\sum_{i=1}^N |y_i - \bx_i' \bbeta|$, as done in {\em Least Absolute Deviation (LAD)}, to mitigate the influence of large residuals. Another commonly used approach for hedging against outliers is M-estimation \citep{huber1964robust, huber1973robust}, which minimizes a symmetric loss function $\rho(\cdot)$ of the residuals in the form $\sum_{i=1}^N \rho(y_i - \bx_i' \bbeta)$, that downweights the influence of samples with large absolute residuals. Several choices for $\rho(\cdot)$ include the Huber function \citep{huber1964robust, huber1973robust}, the Tukey's Biweight function \citep{PJ05}, the logistic function \citep{coleman1980system}, the Talwar function \citep{hinich1975simple}, and the Fair function \citep{fair1974robust}. 

Both LAD and M-estimation are not resistant to large deviations in the predictors. For contamination present in the predictor space, high breakdown value methods are required. Examples include the {\em Least Median of Squares (LMS)} \citep{rousseeuw1984least} which minimizes the median of the absolute residuals, the {\em Least Trimmed Squares (LTS)} \citep{rousseeuw1985multivariate} which minimizes the sum of the $q$ smallest squared residuals, and S-estimation \citep{rousseeuw1984robust} which has a higher statistical efficiency than LTS with the same breakdown value. A combination of the high breakdown value method and M-estimation is the MM-estimation \citep{yohai1987high}. It has a higher statistical efficiency than S-estimation. We refer the reader to the book of \citet{PJ05} for an elaborate description of these robust regression methods. 
 
The aforementioned robust estimation procedures focus on modifying the objective function in a heuristic way with the intent of minimizing the effect of outliers. A more rigorous line of research explores the underlying stochastic program that leads to the sample-based estimation procedures. For example, the OLS objective can be viewed as minimizing the expected squared residual under the uniform empirical distribution over the samples. It has been well recognized that optimizing under the empirical distribution yields estimators that are sensitive to perturbations in the data and suffer from overfitting. The reason is that, when the data $(\bx, y)$ is adversarially corrupted by outliers, the observed samples are not representative enough to encode the true underlying uncertainty of the data. But on the other hand, the samples are typically the only information available. Instead of equally weighting all the samples as in the empirical distribution, we may wish to include more informative distributions that ``drive out'' the corrupted samples. One way to realize this is to hedge the expected loss against a family of
distributions that include the true data-generating mechanism with a high confidence, which is called {\em Distributionally Robust Optimization (DRO)}. DRO minimizes the worst-case expected loss over
a probabilistic ambiguity set $\scrP$ that is constructed from the observed samples and characterized by certain known
properties of the true data-generating distribution. For example,
\cite{San11} study the distributionally robust least squares problem
with $\scrP$ defined through either moment constraints, norm bounds
with moment constraints, or a confidence region over a reference
probability measure. Compared to the single distribution-based stochastic optimization, DRO often results in better
out-of-sample performance due to its distributional robustness.

The existing literature on DRO can be split into two main branches
according to the way in which $\scrP$ is defined. One is through a
moment ambiguity set, which contains all distributions that satisfy
certain moment constraints \citep[see][]{popescu2007robust, Ye10,
  goh2010distributionally, zymler2013distributionally, Sim14}. In many
cases it leads to a tractable DRO problem but has been criticized for
yielding overly conservative solutions \citep{wang2016likelihood}. The
other is to define $\scrP$ as a ball of distributions using some
probabilistic distance functions such as the $\phi$-divergences
\citep{bayraksan2015data}, which include the Kullback-Leibler (KL)
divergence \citep{Hu13, jiang2015data} as a special case, the Prokhorov
metric \citep{erdougan2006ambiguous}, and the Wasserstein distance
\citep{Pey15, gao2016distributionally, zhao2015data, luo2017decomposition, blanchet2016quantifying}. Deviating from the stochastic setting, there are also some works focusing on deterministic robustness. \cite{LAU97} consider the least squares problem with unknown
but bounded, non-random disturbance and solve it in polynomial
time. \citet{Cara08} study the robust linear regression problem with
norm-bounded feature perturbation and show that it is equivalent to the
$\ell_1$-regularized regression. See \citet{yang2013unified, bertsimas2017characterization} which also use a deterministic robustness.

In this paper we consider a DRO problem with $\scrP$ containing
distributions that are close to the discrete empirical distribution in
the sense of Wasserstein distance. The reason for choosing the
Wasserstein metric is two-fold. On one hand, the Wasserstein ambiguity
set is rich enough to contain both continuous and discrete relevant
distributions, while other metrics such as the KL divergence, exclude
all continuous distributions if the nominal distribution is discrete
\citep{Pey15, gao2016distributionally}. Furthermore, considering
distributions within a KL distance from the empirical, does not allow
for probability mass outside the support of the empirical
distribution. On the other hand, measure concentration results
guarantee that the Wasserstein set contains the true data-generating
distribution with high confidence for a sufficiently large sample size
\citep{Four14}. Moreover, the Wasserstein metric takes into account the closeness between support points while other metrics such as the $\phi$-divergence only consider the probabilities on these points. The image retrieval example in
\citet{gao2016distributionally} suggests that the probabilistic ambiguity set constructed based on the KL divergence prefers the pathological distribution to the true
distribution, whereas the Wasserstein
distance does not exhibit such a problem. The reason lies in that
$\phi$-divergence does not incorporate a notion of closeness between two
points, which in the context of image retrieval represents the
perceptual similarity in color.

Our DRO problem minimizes the worst-case absolute residual over a Wasserstein ball of distributions, and could be relaxed to the following form:
\begin{equation} \label{eq1} 
\inf\limits_{\bbeta} \frac{1}{N}\sum\limits_{i=1}^N|y_i - \bx_i' \bbeta| + \epsilon\|(-\bbeta, 1)\|_*,
\end{equation}
where $\epsilon$ is the radius of the Wasserstein ball, and $\|\cdot\|_*$ is the dual norm of the norm space where the Wasserstein metric is defined on. Formulation (\ref{eq1}) incorporates a wide class of
models whose specific form depends on the notion of transportation cost embedded in the
Wasserstein metric (see Section \ref{s2}).
Although the Wasserstein DRO formulation simply reduces to regularized regression models, we want to emphasize a few new insights brought by this methodology. First, the regularization term controls the conservativeness of the Wasserstein set, or the amount of ambiguity in the data, which differentiates itself from the heuristically added regularizers in traditional regression models that serve the purpose of preventing overfitting, error/variance reduction, or sparsity recovery. Second, the regularization term is determined by the dual norm of the regression coefficient, which controls the {\em growth rate} of the $\ell_1$-loss function, and the radius of the Wasserstein set. This connection provides guidance on the selection of the regularization coefficient and may lead to significant computational savings compared to cross-validation. DRO essentially enables new and more accurate interpretations of the regularizer, and establishes its dependence on the {\em growth rate} of the loss, the underlying metric space and the reliability of the observed samples.

The connection between robustness and regularization has been established in several works. The earliest one may be credited to \cite{LAU97}, who show that minimizing the worst-case squared residual within a Frobenius norm-based perturbation set is equivalent to Tikhonov regularization. In more recent works, using properly selected uncertainty sets, \citet{Cara08} has shown the equivalence between robust linear regression with feature perturbations and the {\em Least Absolute Shrinkage and Selection Operator (LASSO)}. \citet{yang2013unified} extend this to more general LASSO-like procedures, including versions of the grouped LASSO. \citet{bertsimas2017characterization} give a comprehensive characterization of the conditions under which robustification and regularization are equivalent for regression models with deterministic norm-bounded perturbations on the features. For classification problems,
\citet{xu2009robustness} show the equivalence between the regularized
support vector machines (SVMs) and a robust optimization formulation, by
allowing potentially correlated disturbances in the
covariates. \citet{shafieezadeh2015distributionally} consider a robust
version of logistic regression under the assumption that the probability
distributions under consideration lie in a Wasserstein ball, and they
show that the regularized logistic regression is a special case of this
robust formulation. Recently, \citet{shafieezadeh2017regularization, gao2017wasserstein} have provided a unified framework for connecting the Wasserstein DRO with regularized learning procedures, for various regression and classification models.  

Our work is motivated by the problem of identifying patients who receive an abnormally high radiation exposure in CT exams, given the patient characteristics and exam-related variables \citep{ctradiation}. This could be casted as an outlier detection problem; specifically, estimating a robustified regression plane that is immunized against outliers and learns the underlying true relationship between radiation dose and the relevant predictors. We focus on robust learning of the parameter in regression models under distributional perturbations residing within a Wasserstein ball. While the applicability of the Wasserstein DRO methodology is not restricted to regression analysis \citep{sinha2017certifiable, gao2017wasserstein, shafieezadeh2017regularization}, or a particular form of the loss function (as long as it satisfies certain smoothness conditions \citep{gao2017wasserstein}), we focus on the absolute residual loss in linear regression in light of our motivating application and for the purpose of enhancing robustness. Our contributions may be summarized as follows: 
\begin{enumerate}
\item We develop a DRO approach to robustify linear regression using an
  $\ell_1$ loss function and an ambiguity set around the empirical
  distribution of the training samples defined based on the Wasserstein
  metric. The formulation is general enough to include any norm-induced
  Wasserstein metric and incorporate additional regularization
  constraints on the regression coefficients (e.g., $\ell_1$-norm
  constraints). 
   It provides an intuitive connection between the amount of
   ambiguity allowed and a regularization penalty term in the robust
   formulation, which provides a natural way to adjust the latter.  


\item We establish novel performance guarantees on both the out-of-sample loss (prediction bias)
  and the discrepancy between the estimated
  and the true regression coefficients (estimation bias). Our guarantees manifest the role of the regularizer, which is related to the dual norm of the regression coefficients, in bounding the biases and are in concert with the theoretical foundation that leads to the regularized problem. The generalization error bound, in particular, builds a connection between the loss function and the form of the regularizer via Rademacher complexity, providing a rigorous explanation for the commonly observed good out-of-sample performance of regularized regression. On the other hand, the estimation error bound corroborates the validity of the $\ell_1$-loss function, which tends to incur a lower estimation bias than other candidates such as the $\ell_2$ and $\ell_{\infty}$ losses.  
  Our results are novel in the
  robust regression setting and different from earlier work in the DRO
  literature, enabling new perspectives and interpretations of the norm-based regularization, and providing justifications for the $\ell_1$-loss based learning algorithms.

\item We empirically explore three important aspects of the Wasserstein DRO formulation, including the advantages of the $\ell_1$-loss function, the selection of a proper norm for the Wasserstein metric, and the implication of penalizing the {\em extended regression coefficient} $(-\bbeta, 1)$, through comparing with a series of regression models on a number of synthetic datasets. We show the superiority of the Wasserstein DRO approach, presenting a thorough analysis, under four different experimental setups. We also consider the application of our methodology to outlier detection, and compare with M-estimation in terms of the ability of identifying outliers ({\em ROC (Receiver Operating Characteristic)} curves). The Wasserstein DRO formulation achieves significantly higher {\em AUC (Area Under Curve)} values.
\end{enumerate}

The rest of the paper is organized as follows. In Section \ref{s2}, we
introduce the Wasserstein metric and derive the general Wasserstein DRO
formulation in a linear regression framework. Section \ref{s3}
establishes performance guarantees for both the general formulation and
the special case where the Wasserstein metric is defined on the
$\ell_1$-norm space. The numerical experimental results are presented in
Section \ref{s4}. We conclude the paper in Section \ref{s5}.

\textbf{Notational conventions:} We use boldfaced lowercase letters to
denote vectors, ordinary lowercase letters to denote scalars, boldfaced
uppercase letters to denote matrices, and calligraphic capital letters
to denote sets. $\mathbb{E}$ denotes expectation and $\mathbb{P}$
probability of an event. All vectors are column vectors. For space
saving reasons, we write $\bx=(x_1, \ldots, x_{\text{dim}(\bx)})$ to
denote the column vector $\bx$, where $\text{dim}(\bx)$ is the dimension
of $\bx$. We use prime to denote the transpose of a vector, $\|\cdot\|$
for the general norm operator, $\|\cdot\|_2$ for the $\ell_2$ norm,
$\|\cdot\|_1$ for the $\ell_1$ norm, and $\|\cdot\|_{\infty}$ for the
infinity norm. $\scrP(\scrZ)$ denotes the set of probability measures
supported on $\scrZ$. $\mathbf{e}_i$ denotes the $i$-th unit vector, $\mathbf{e}$ the vector of ones,
$\bzero$ a vector of zeros, and $\bI$ the identity matrix. Given a norm $\|\cdot\|$ on $\mbb{R}^m$, the dual norm
$\|\cdot\|_*$ is defined as: $\|\btheta\|_* \triangleq \sup_{\|\bz\|\le 1}\btheta'\bz$. For a function $h(\bz)$, its convex conjugate $h^*(\cdot)$ is defined as:
$h^*(\btheta) \triangleq
\sup_{\bz \in \text{dom} \ h}\ \{\btheta'\bz-h(\bz)\},$
where $\text{dom} \ h$ denotes the domain of the function $h$.

\section{Problem Statement and Justification of Our Formulation} \label{s2}

\begin{figure}[h]
	\centering
	\includegraphics[height = 2in]{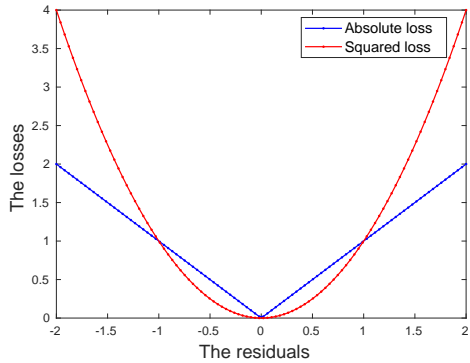}
	\caption{The comparison between $\ell_1$ and $\ell_2$ loss functions.}
	\label{absloss}
\end{figure}

Consider a linear regression problem where we are given a predictor/feature vector $\bx \in \mbb{R}^{m-1}$, and a response variable $y \in \mbb{R}$. Our goal is to obtain an
accurate estimate of the regression plane that is robust with respect to the adversarial perturbations in the data. We consider an $\ell_1$-loss function $h_{\bbeta}(\bx, y) \triangleq |y - \bx'\bbeta|$, motivated by the observation that the absolute loss function is more robust to large residuals than the squared loss (see Fig. \ref{absloss}). Moreover, the estimation error analysis presented in Section \ref{limit} suggests that the $\ell_1$-loss function leads to a smaller estimation bias than others. Our Wasserstein DRO problem using the $\ell_1$-loss function is formulated as:
\begin{equation} \label{dro}
\inf\limits_{\bbeta\in \scrB}\sup\limits_{\mbb{Q}\in \Omega}
\mbb{E}^{\mbb{Q}}\big[ |y-\bx'\bbeta|\big], 
\end{equation}
where $\bbeta$ is the regression coefficient vector that belongs
to some set $\scrB$. $\scrB$ could be $\mbb{R}^{m-1}$,
or $\scrB = \{\bbeta: \|\bbeta\|_1\le l\}$ if we wish to induce
sparsity, with $l$ being some pre-specified number. $\mbb{Q}$ is the
probability distribution of $(\bx, y)$, belonging to some set $\Omega$ which is
defined as:
\begin{equation*}
\Omega \triangleq \{\mbb{Q}\in \scrP(\scrZ): W_p(\mathbb{Q},\ \hat{\mathbb{P}}_N) \le \epsilon\},
\end{equation*}
where $\scrZ$ is the set of possible values for $(\bx, y)$; $\scrP(\scrZ)$ is
the space of all probability distributions supported on $\scrZ$;
$\epsilon$ is a pre-specified radius of the Wasserstein ball; and
$W_p(\mbb{Q},\ \hat{\mbb{P}}_N)$ is the order-$p$ Wasserstein distance
between $\mbb{Q}$ and $\hat{\mbb{P}}_N$ (see definition in (\ref{wass_p})), with $\hat{\mbb{P}}_N$ the uniform empirical distribution over samples. The formulation
in (\ref{dro}) is robust since it minimizes over the regression
coefficients the worst case expected loss, that is, the expected loss
maximized over all probability distributions in the ambiguity set
$\Omega$.

Before deriving a tractable reformulation for (\ref{dro}), let us first define the Wasserstein metric.
Let $(\scrZ, s)$ be a metric space where $\scrZ$ is a set and $s$ is a
metric on $\scrZ$. The Wasserstein metric of order $p \ge 1$ defines the
distance between two probability distributions $\mbb{Q}_1$ and
$\mbb{Q}_2$ in the following way:
\begin{equation} \label{wass_p}
W_p (\mbb{Q}_1, \mbb{Q}_2) \triangleq \Biggl(\min\limits_{\Pi \in \scrP(\scrZ \times \scrZ)} \biggl\{\int_{\scrZ \times \scrZ} \bigl(s((\bx_1, y_1), (\bx_2, y_2))\bigr)^p \ \Pi \bigl(d(\bx_1, y_1), d(\bx_2, y_2)\bigr)\biggr\}\Biggr)^{1/p},
\end{equation}
where $\Pi$ is the joint distribution of $(\bx_1, y_1)$ and $(\bx_2, y_2)$ with
marginals $\mbb{Q}_1$ and $\mbb{Q}_2$, respectively. The Wasserstein
distance between $\mbb{Q}_1$ and $\mbb{Q}_2$ represents the cost of an
optimal mass transportation plan, where the cost is measured through the
metric $s$. The order $p$ should be selected in such a way as to ensure that the worst-case expected loss is meaningfully defined, i.e.,
\begin{equation} \label{finitesup}
	\mbb{E}^{\mbb{Q}}\big[ h_{\bbeta}(\bx, y) \big] < \infty, \ \forall \mbb{Q}\in \Omega.
\end{equation}
Notice that the ambiguity set $\Omega$ is centered at the empirical distribution $\hat{\mbb{P}}_N$ and has radius $\epsilon$. It may be desirable to translate (\ref{finitesup}) into:
\begin{equation} \label{finitedif}
\Bigl|\mbb{E}^{\mbb{Q}}\big[ h_{\bbeta}(\bx, y)\big] - \mbb{E}^{\hat{\mbb{P}}_N}\big[ h_{\bbeta}(\bx, y)\big]\Bigr|< \infty, \ \forall \mbb{Q}\in \Omega.
\end{equation}
We want to relate (\ref{finitedif}) with the Wasserstein distance $W_p(\mathbb{Q},\ \hat{\mathbb{P}}_N)$, which is no larger than $\epsilon$ for all $\mbb{Q}\in \Omega$. The LHS of (\ref{finitedif}) could be written as:
\begin{equation} \label{finitedifexp}
\begin{aligned}
	  & \Bigl|\mbb{E}^{\mbb{Q}}\big[ h_{\bbeta}(\bx, y)\big] -  \mbb{E}^{\hat{\mbb{P}}_N}\big[ h_{\bbeta}(\bx, y)\big]\Bigr|  \\
	  = & \biggl|\int_{\scrZ} h_{\bbeta}(\bx_1, y_1) \mbb{Q}(d(\bx_1, y_1)) - \int_{\scrZ} h_{\bbeta}(\bx_2, y_2) \hat{\mbb{P}}_N(d(\bx_2, y_2)) \biggr| \\
	 = & \biggl|\int_{\scrZ} h_{\bbeta}(\bx_1, y_1) \int_{\scrZ}\Pi_0(d(\bx_1, y_1), d(\bx_2, y_2)) - \int_{\scrZ} h_{\bbeta}(\bx_2, y_2) \int_{\scrZ}\Pi_0(d(\bx_1, y_1), d(\bx_2, y_2)) \biggr| \\
	 \le & \int_{\scrZ \times \scrZ} \bigl|h_{\bbeta}(\bx_1, y_1)-h_{\bbeta}(\bx_2, y_2)\bigr| \Pi_0(d(\bx_1, y_1), d(\bx_2, y_2)),
\end{aligned}
\end{equation}
where $\Pi_0$ is the joint distribution of $(\bx_1, y_1)$ and $(\bx_2, y_2)$ with
marginals $\mbb{Q}$ and $\hat{\mbb{P}}_N$, respectively. Comparing (\ref{finitedifexp}) with (\ref{wass_p}), we see that for (\ref{finitedif}) to hold, the following quantity which characterizes the {\em growth rate} of the loss function needs to be bounded:
\begin{equation} \label{growthrate}
\text{GR}_{h_{\bbeta}}((\bx_1, y_1), (\bx_2, y_2)) \triangleq \frac{\bigl|h_{\bbeta}(\bx_1, y_1) - h_{\bbeta}(\bx_2, y_2)\bigr|}{\bigl(s((\bx_1, y_1), (\bx_2, y_2))\bigr)^p},  \ \forall (\bx_1, y_1), (\bx_2, y_2) \in \scrZ.
\end{equation}
A formal definition of the growth rate is due to \citet{gao2016distributionally}, which takes the limit of (\ref{growthrate}) as $s((\bx_1, y_1), (\bx_2, y_2)) \rightarrow \infty$, to eliminate its dependence on $(\bx, y)$. One important aspect they have pointed out is that when the growth rate of the loss
function is infinite, strong duality for the worst-case problem $\sup_{\mbb{Q}\in \Omega}
\mbb{E}^{\mbb{Q}}\big[ h_{\bbeta}(\bx, y)\big]$ fails to hold, in which case the DRO problem (\ref{dro}) becomes
intractable. Assuming that the metric $s$ is induced by some norm $\|\cdot\|$, the bounded {\em growth rate} requirement is expressed as follows:
\begin{equation} \label{gr}
\begin{split}
& \limsup_{\|(\bx_1, y_1)-(\bx_2, y_2)\| \rightarrow \infty} \frac{|h_{\bbeta}(\bx_1, y_1) - h_{\bbeta}(\bx_2, y_2)|}{\|(\bx_1, y_1)-(\bx_2, y_2)\|^p} \le \limsup_{\|(\bx_1, y_1)-(\bx_2, y_2)\| \rightarrow \infty} \frac{|y_1 - \bx_1' \bbeta - (y_2 - \bx_2' \bbeta)|}{\|(\bx_1, y_1)-(\bx_2, y_2)\|^p} \\
 \le & \limsup_{\|(\bx_1, y_1)-(\bx_2, y_2)\| \rightarrow \infty} \frac{\|(\bx_1, y_1)-(\bx_2, y_2)\| \|(-\bbeta, 1)\|_*}{\|(\bx_1, y_1)-(\bx_2, y_2)\|^p} < \infty,\\
\end{split}
\end{equation}
where $\|\cdot\|_*$ is the dual norm of $\|\cdot\|$, and the second inequality is due to the Cauchy-Schwarz inequality. Notice that by taking $p=1$, (\ref{gr}) is equivalently translated into the condition that $\|(-\bbeta, 1)\|_* < \infty$, which we will see in Section \ref{s3} is an essential requirement to guarantee a good generalization performance for the Wasserstein DRO estimator. The growth rate essentially reveals the underlying metric space used by the Wasserstein distance. Taking $p>1$ leads to zero growth rate in the limit of (\ref{gr}), which is not desirable since it removes the Wasserstein ball structure from our formulation and renders it an optimization problem over a singleton distribution. This will be made more clear in the following analysis. We thus choose the order-$1$ Wasserstein metric with $s$ being induced by some norm $\|\cdot\|$ to define our DRO problem. 
  



Next, we will discuss how to convert (\ref{dro}) into a tractable formulation. Suppose we have $N$ independently and identically distributed
realizations of $(\bx, y)$, denoted by $(\bx_i,y_i), i = 1, \ldots, N$. We make the assumption that $(\bx, y)$ comes from a mixture of two distributions, with probability $q$ from the outlying distribution
$\mathbb{P}_{out}$ and with probability $1-q$ from the true distribution
$\mathbb{P}$. Recall that $\hat{\mathbb{P}}_N$ is the discrete uniform
distribution over the $N$ samples. Our goal is to generate estimators that are consistent with the true distribution $\mathbb{P}$. We claim that when $q$ is small, if the Wasserstein ball radius $\epsilon$ is chosen judiciously, the true distribution $\mathbb{P}$ will be included in the set $\Omega$ while the outlying distribution $\mathbb{P}_{out}$ will be excluded. To see this, consider a simple example where $\mathbb{P}$ is a discrete distribution that assigns equal probability to $10$ data points equally spaced between $0.1$ and $1$, and $\mathbb{P}_{out}$ assigns probability $0.5$ to two data points $1$ and $2$. We generate $100$ samples and plot the Wasserstein distances from $\hat{\mathbb{P}}_N$ for both $\mathbb{P}$ and $\mathbb{P}_{out}$.
\begin{figure}[H]
	\centering
	\includegraphics[height = 2in]{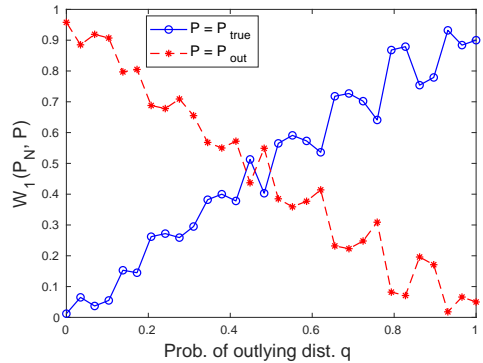}
	\caption{The order-$1$ Wasserstein distances from the empirical distribution.}
	\label{wassdis}
\end{figure}
From Fig. \ref{wassdis} we observe that for $q$ below $0.5$, the true distribution $\mathbb{P}$ is closer to $\hat{\mathbb{P}}_N$ whereas the outlying distribution $\mathbb{P}_{out}$ is further away. If the radius $\epsilon$ is chosen between the red ($\ast-$) and blue ($\circ-$) lines, the Wasserstein ball that we are hedging against will exclude the outlying distribution and the resulting estimator will be robust to the adversarial perturbations. Moreover, as $q$ becomes smaller, the gap between the red and blue lines becomes larger. One implication from this observation is that as the data becomes purer, the radius of the Wasserstein ball tends to be smaller, and the confidence in the observed samples is higher. For large $q$ values, the DRO formulation seems to fail. However, as outliers are defined to be the data points that do not conform to the majority of data, we can safely claim that $\mathbb{P}_{out}$ is the distribution of the minority and $q$ is always below $0.5$. 

We now look at the inner supremum in (\ref{dro}). \citet[Theorem 6.3]{Pey15} show that when the set
$\scrZ$ is closed and convex, and the loss function $h_{\bbeta}(\bx, y)$ is convex in $(\bx, y)$, 
\begin{equation} \label{kk}
\sup\limits_{\mbb{Q}\in \Omega}\mbb{E}^{\mbb{Q}}[h_{\bbeta}(\bx, y)]\le
\kappa\epsilon+\frac{1}{N}\sum\limits_{i=1}^Nh_{\bbeta}(\bx_i, y_i),  \ \forall \epsilon \geq 0,
\end{equation}
where $\kappa(\bbeta)=\sup\{\|\btheta\|_*: h_{\bbeta}^*(\btheta)<\infty\}$, with $h_{\bbeta}^*(\cdot)$ the convex conjugate function of $h_{\bbeta}(\bx, y)$. Through (\ref{kk}), we can relax
problem (\ref{dro}) by minimizing the right hand side of (\ref{kk})
instead of the worst-case expected loss. Moreover, as shown in
\citet{Pey15}, (\ref{kk}) becomes an equality when
$\scrZ=\mathbb{R}^{m}$. In Theorem \ref{kappa}, we compute the value of
$\kappa(\bbeta)$ for the specific $\ell_1$ loss function we use. The proof of this Theorem and all results hereafter are included in Appendix A.

\begin{theorem} \label{kappa} Define $\kappa(\bbeta) = \sup\{\|\btheta\|_*:
	h_{\bbeta}^*(\btheta)<\infty\}$, where $\|\cdot\|_*$ is the dual norm
	of $\|\cdot\|$, and $h_{\bbeta}^*(\cdot)$ is the conjugate
	function of $h_{\bbeta}(\cdot)$.  When the
	loss function is $h_{\bbeta}(\bx, y) = |y-\bx'\bbeta|$, we have $\kappa(\bbeta) =
	\|(-\bbeta, 1)\|_*$.
\end{theorem}

Due to Theorem \ref{kappa}, (\ref{dro}) could be formulated as
the following optimization problem:
\begin{equation} \label{qcp}
\inf\limits_{\bbeta\in \scrB}\epsilon\|(-\bbeta, 1)\|_*+\frac{1}{N}\sum\limits_{i=1}^N|y_i - \bx_i'\bbeta|.
\end{equation}
Note that the regularization term of (\ref{qcp}) is the product of the {\em growth rate} of the loss and the Wasserstein ball radius. The growth rate is closely related to the way the Wasserstein metric defines the transportation costs on the data $(\bx, y)$. As mentioned earlier, a zero growth rate diminishes the effect of the Wasserstein distributional uncertainty set, and the resulting formulation would simply be an empirical loss minimization problem.
The parameter $\epsilon$ controls the conservativeness of the
formulation, whose selection depends on the sample size, the dimensionality of the data, and the
confidence that the Wasserstein ball contains the true distribution
\citep[see eq. (8) in][]{Pey15}.  Roughly speaking, when the sample size
is large enough, and for a fixed confidence level, $\epsilon$ is
inversely proportional to $N^{1/m}$.

Formulation (\ref{qcp}) incorporates a class of models whose specific form depends
on the norm space we choose, which could be application-dependent and
practically useful. For example, when the Wasserstein metric $s$ is induced by $\|\cdot\|_2$ and the set
$\scrB$ is the intersection of a polyhedron with convex quadratic
inequalities, (\ref{qcp}) is a convex quadratic problem which can
be solved to optimality very efficiently. Specifically, it could be
converted to:
\begin{equation}  \label{l2qcp}
\begin{aligned}
\min\limits_{\substack{a, \ b_1, \ldots, b_N, \ \bbeta}} \quad & a\epsilon+\frac{1}{N}\sum_{i=1}^N b_i\\
\text{s.t.} \quad & \|\bbeta\|_2^2 + 1 \le a^2,\\
& y_i - \bx_i'\bbeta \le b_i, \  i=1,\ldots,N,\\
& -(y_i - \bx_i'\bbeta) \le b_i, \  i=1,\ldots,N,\\
& a, \ b_i\ge 0, \ i=1,\ldots,N,\\
& \bbeta \in \scrB.
\end{aligned}
\end{equation}
When the Wasserstein metric is defined using $\|\cdot\|_1$ and the set
$\scrB$ is a polyhedron, (\ref{qcp}) is a linear programming problem:
\begin{equation}  \label{l1qcp}
\begin{aligned}
\min\limits_{\substack{a, \ b_1, \ldots, b_N, \ \bbeta}} \quad & a\epsilon+\frac{1}{N}\sum_{i=1}^N b_i\\
\text{s.t.} \quad & a \ge \bbeta'\mathbf{e}_i, \ i=1,\ldots,m-1,\\
& a \ge -\bbeta'\mathbf{e}_i, \ i=1,\ldots,m-1,\\
& y_i - \bx_i'\bbeta \le b_i, \  i=1,\ldots,N,\\
& -(y_i - \bx_i'\bbeta) \le b_i, \  i=1,\ldots,N,\\
& a \ge 1,\\
& b_i\ge 0, \ i=1,\ldots,N,\\
& \bbeta \in \scrB.
\end{aligned}
\end{equation}
More generally, when the coordinates of $(\bx, y)$ differ
from each other substantially, a properly chosen, positive definite
weight matrix $\bM \in \mbb{R}^{m \times m}$ could scale correspondingly different coordinates of
$(\bx, y)$ by using the $\bM$-weighted norm:
\begin{equation*}
\|(\bx, y)\|_{\bM} = \sqrt{(\bx, y)' \bM (\bx, y)}.
\end{equation*}     
It
can be shown that (\ref{qcp}) in this case becomes:
\begin{equation} \label{wqcp}
\inf\limits_{\bbeta\in \scrB}\epsilon \sqrt{(-\bbeta, 1)' \bM^{-1} (-\bbeta, 1)} +\frac{1}{N}\sum\limits_{i=1}^N|y_i - \bx_i'\bbeta|.
\end{equation}

We note that this Wasserstein DRO framework could be applied to a broad class of loss functions and the tractable reformulations have been derived in \citet{shafieezadeh2017regularization, gao2017wasserstein} for regression and classification models. We adopt the absolute residual loss in this paper to enhance the robustness of the formulation, which is the focus of our work and serves the purpose of estimating robust parameters that are immunized against perturbations/outliers. Notice that
(\ref{qcp}) coincides with the regularized LAD models \citep{pollard1991asymptotics, wang2006regularized}, except that we are regularizing a variant of the regression coefficient. We would like to highlight
several novel viewpoints that are brought by the Wasserstein DRO framework and justify the value and novelty of
(\ref{qcp}). First, (\ref{qcp}) is obtained as an
outcome of a fundamental DRO formulation, which enables new interpretations of the regularizer from the standpoint of distributional robustness, and provides rigorous theoretical foundation on why the $\ell_2$-regularizer prevents overfitting to the training data. The regularizer
could be seen as a control over the amount of ambiguity in the data 
and reveals the reliability of the
contaminated samples. Second, the geometry of the Wasserstein ball is embedded in the regularization term, which penalizes the regression coefficient on the dual Wasserstein space, with the magnitude of penalty being the radius of the ball. This offers an
intuitive interpretation and provides guidance on how to set the regularization coefficient. Moreover, different from the traditional regularized LAD models that directly penalize the regression coefficient $\bbeta$, we regularize the vector $(-\bbeta, 1)$, where the $1$ takes into account the transportation cost along the $y$ direction. Penalizing only on $\bbeta$ corresponds to an infinite transportation cost along $y$. Our model is more general in this sense, and establishes the connection between the metric space on data and the form of the regularizer.

\section{Performance Guarantees}  \label{s3}

Having obtained a tractable reformulation for the Wasserstein DRO problem, we next establish guarantees on the predictive power and estimation quality for the solution to (\ref{qcp}). Two types of results will be presented in this section, one of which bounds the prediction bias of the estimator on new, future data (given in Section \ref{out}). The other one that bounds
the discrepancy between the estimated and true regression planes (estimation bias), is given
in Section \ref{limit}.
\subsection{Out-of-Sample Performance} \label{out} 
In this subsection we investigate generalization characteristics of the
solution to (\ref{qcp}), which involves measuring the error generated by
our estimator on a new random sample $(\bx, y)$. We would like to obtain
estimates that not only explain the observed samples well, but, more
importantly, possess strong generalization abilities. The derivation is
mainly based on {\em Rademacher complexity} \citep[see][]{Peter02}, which is a
measurement of the complexity of a class of functions. We would like to emphasize the applicability of such a proof technique to general loss functions, as long as their empirical Rademacher complexity could be bounded. The bound we derive for the prediction bias
depends on both the sample average loss (the training error) and the dual norm of the regression coefficient (the regularizer), which corroborates the validity and necessity of our regularized formulation. Moreover, the generalization result also builds a connection between the loss function and the form of the regularizer via Rademacher complexity, which enables new insights into the regularization term and explains the commonly observed good out-of-sample performance of regularized regression in a rigorous way. 
We first make several mild assumptions that are needed for the generalization result.

\begin{ass} \label{a1} The norm of the uncertainty parameter $(\bx, y)$ is
  bounded above almost surely, i.e., $\|(\bx, y)\|\le R.$
\end{ass}
\begin{ass} \label{a2} The dual norm of $(-\bbeta, 1)$ is bounded
	above within the feasible region, namely,
	\begin{equation*}
	\sup\limits_{\bbeta\in \scrB}\|(-\bbeta, 1)\|_*=\bar{B}.
	\end{equation*}
\end{ass}
Under these two assumptions, the absolute loss could be bounded via the Cauchy-Schwarz inequality.
\begin{lem} \label{l1}
	For every feasible $\bbeta$, it follows
	\begin{equation*}
	|y - \bx'\bbeta|\le \bar{B}R, \quad \text{almost surely}.
	\end{equation*}
\end{lem} 

With the above result, the idea is to bound the generalization error using the empirical {\em Rademacher complexity} of
the following class of loss functions: 
\begin{equation*}
\scrH=\{(\bx, y) \mapsto h_{\bbeta}(\bx, y): h_{\bbeta}(\bx, y)= |y - \bx'\bbeta|,\
\bbeta\in \scrB\}. 
\end{equation*}
We need to show that the empirical Rademacher complexity of
$\scrH$, denoted by $\scrR_N(\scrH)$, is upper bounded. The following result, similar to Lemma 3 in
\citet{Dim14}, provides a bound that is inversely proportional to the square root of the sample size.
\begin{lem} \label{radcom}
	\begin{equation*}
	\scrR_N(\scrH)\le \frac{2\bar{B}R}{\sqrt{N}}.
	\end{equation*}
\end{lem}

Let $\hat{\bbeta}$ be an optimal solution to (\ref{qcp}), obtained using
the samples $(\bx_i, y_i)$, $i=1,\ldots,N$. Suppose we draw a new i.i.d.\
sample $(\bx,y)$. In Theorem \ref{t2} we establish bounds on the
error $|y - \bx'\hat{\bbeta}|$.
 \begin{theorem} \label{t2} Under Assumptions \ref{a1} and \ref{a2}, for any
	$0<\delta<1$, with probability at least $1-\delta$ with respect to the
	sampling,
	\begin{equation} \label{exp}
	\mathbb{E}[|y - \bx'\hat{\bbeta}|]\le
	\frac{1}{N}\sum_{i=1}^N
	|y_i - \bx_i'\hat{\bbeta}|+\frac{2\bar{B}R}{\sqrt{N}}+ 
	\bar{B}R\sqrt{\frac{8\log(2/\delta)}{N}}\ ,
	\end{equation}
	and for any $\zeta>\frac{2\bar{B}R}{\sqrt{N}}+
	\bar{B}R\sqrt{\frac{8\log(2/\delta)}{N}}$,
	\begin{equation} \label{prob}
	\mathbb{P}\biggl(|y - \bx'\hat{\bbeta}| \ge
	\frac{1}{N}\sum_{i=1}^N |y_i - \bx_i'\hat{\bbeta}|+\zeta\biggr) 
	\le \frac{\frac{1}{N}\sum_{i=1}^N
		|y_i - \bx_i'\hat{\bbeta}|+\frac{2\bar{B}R}{\sqrt{N}}+ 
		\bar{B}R\sqrt{\frac{8\log(2/\delta)}{N}}}{\frac{1}{N}\sum_{i=1}^N
		|y_i - \bx_i'\hat{\bbeta}|+\zeta}. 
	\end{equation}
\end{theorem}

There are two probability measures in the statement of Theorem
\ref{t2}. One is related to the new data $(\bx, y)$, while the other is
related to the samples $(\bx_1, y_1),\ldots,(\bx_N, y_N)$. The expectation in
(\ref{exp}) (and the probability in (\ref{prob})) is taken w.r.t. the new
data $(\bx, y)$. For a given set of samples, (\ref{exp}) (and (\ref{prob}))
holds with probability at least $1-\delta$ w.r.t.\ the measure of
samples. Theorem \ref{t2} essentially says that given typical samples,
the expected loss on new data using our Wasserstein DRO estimator could
be bounded above by the average sample loss plus extra terms that depend on the supremum of $\|(-\bbeta, 1)\|_*$ (our regularizer), and are
proportional to $1/\sqrt{N}$. This result validates the dual norm-based regularized regression from the perspective of generalization ability, and could be generalized to any bounded loss function. It also provides implications on the form of the regularizer. For example, if given an $\ell_2$-loss function, the dependency on $\bar{B}$ for the generalization error bound will be of the form $\bar{B}^2$, which suggests using $\|(-\bbeta, 1)\|_*^2$ as a regularizer, reducing to a variant of ridge regression \citep{hoerl1970ridge} for $\|\cdot\|_2$ induced Wasserstein metric. 

We also note that the upper bounds in (\ref{exp}) and (\ref{prob}) do not depend
on the dimension of $(\bx, y)$. This dimensionality-free characteristic
implies direct applicability of our Wasserstein approach to
high-dimensional settings and is particularly useful in many real
applications where, potentially, hundreds of features may be present. Theorem \ref{t2} also provides guidance on the number of
samples that are needed to achieve satisfactory out-of-sample performance. 
\begin{col} \label{samplesize1}
  Suppose $\hat{\bbeta}$ is the optimal solution to (\ref{qcp}). For a
  fixed confidence level $\delta$ and some threshold parameter $\tau \ge
  0$, to guarantee that the percentage difference between the expected
  absolute loss on new data and the sample average loss is less than
  $\tau$, that is,
	\begin{equation*}
	\frac{\mathbb{E}[|y - \bx'\hat{\bbeta}|]-\frac{1}{N}\sum_{i=1}^N |y_i - \bx_i'\hat{\bbeta}|}{\bar{B}R}\le \tau,
	\end{equation*} 
	the sample size $N$ must satisfy
	\begin{equation} \label{N1}
	N\ge \biggl[\frac{2(1+\sqrt{2\log(2/\delta)}\ )}{\tau}\biggr]^2.
	\end{equation}
\end{col}

\begin{col} \label{samplesize2}
  Suppose $\hat{\bbeta}$ is the optimal solution to (\ref{qcp}). For a
  fixed confidence level $\delta$, some $\tau \in (0, 1)$ and $\gamma
  \ge 0$, to guarantee that
	\begin{equation*}
	\mathbb{P}\biggl(\frac{|y - \bx'\hat{\bbeta}|-\frac{1}{N}\sum_{i=1}^N |y_i - \bx_i'\hat{\bbeta}|}{\bar{B}R}\ge \gamma\biggr)\le \tau,
	\end{equation*}
	the sample size $N$ must satisfy
	\begin{equation} \label{N2}
	N\ge 
	\biggl[\frac{2(1+\sqrt{2\log(2/\delta)}\ )}{\tau\cdot\gamma+\tau-1}\biggr]^2,
	\end{equation}
	provided that \  $\tau\cdot\gamma+\tau-1>0$.
\end{col}

In Corollaries \ref{samplesize1} and \ref{samplesize2}, the sample size is inversely proportional to both
$\delta$ and $\tau$, which is reasonable
since the more confident we want to be, the more samples we
need. Moreover, the smaller $\tau$ is, the stricter a requirement we
impose on the performance, and thus more samples are needed.

\subsection{Discrepancy between Estimated and True 
Regression Planes} \label{limit} 

In addition to the generalization performance, we are also interested in the accuracy of the estimator. In this section we seek to bound the difference between the estimated
and true regression coefficients, under a certain distributional assumption on $(\bx, y)$. Throughout the
section we will use $\hat{\bbeta}$ to denote the estimated regression
coefficients, obtained as an optimal solution to (\ref{infty}), and
$\bbeta^*$ for the true (unknown) regression coefficients.
The bound we will derive turns out to be related to the Gaussian width (see definition in the Appendix)
of the unit ball in $\|\cdot\|_{\infty}$, the sub-Gaussian norm of the
uncertainty parameter $(\bx, y)$, as well as the geometric structure of the
true regression coefficients. We note that this proof technique may be applied to several other loss functions, e.g., $\ell_2$ and $\ell_{\infty}$ losses, with slight modifications. However, we will see that the $\ell_1$-loss function incurs a relatively low estimation bias compared to others, further demonstrating the superiority of our absolute error minimization formulation. 

To facilitate the analysis, we will use the following equivalent form of problem (\ref{qcp}):
\begin{equation} \label{infty}
\begin{aligned}
\min_{\bbeta} & \quad \|(-\bbeta, 1)\|_* \\
\text{s.t.} & \quad \|(-\bbeta, 1)'\bZ\|_1\le \gamma_N,\\
& \quad \bbeta\in \scrB,
\end{aligned}
\end{equation}
where $\bZ=[(\bx_1, y_1), \ldots, (\bx_N, y_N)]$ is the matrix with columns $(\bx_i, y_i), 
i = 1, \ldots, N$, and $\gamma_N$ is some exogenous parameter related to
$\epsilon$. One can show that for properly chosen $\gamma_N$, (\ref{infty}) produces the same solution with (\ref{qcp}) \citep{bertsekas1999nonlinear}. (\ref{infty}) is similar to (11) in
\citet{chen2016alternating}, with the difference lying in that we impose
a constraint on the error instead of the gradient, and we consider a more general notion of norm on the coefficient. On the other hand,
due to their similarity, we will follow the line of development in
\citet{chen2016alternating}. Still, our analysis is self-contained and
the bound we obtain is in a different form, which provides meaningful
insights into our specific problem. We list below the assumptions that are needed to bound
the estimation error.

\begin{ass} \label{2norm}
	The $\ell_2$ norm of $(-\bbeta, 1)$ is bounded
	above within the feasible region, namely,
	\begin{equation*}
	\sup\limits_{\bbeta\in \scrB}\|(-\bbeta, 1)\|_2=\bar{B}_2.
	\end{equation*}
\end{ass}

\begin{ass}[Restricted Eigenvalue Condition] \label{RE} 
	For some set $\scrA(\bbeta^*)=\text{cone}\{\bv|\ \\ \|(-\bbeta^*, 1) + \bv\|_* \le\|(-\bbeta^*, 1)\|_*\}\cap \mathbb{S}^m$ and some positive scalar $\underline{\alpha}$, where $\mathbb{S}^m$ is the unit sphere in the $m$-dimensional Euclidean space,
	\begin{equation*}
	\inf\limits_{\bv\in \scrA(\bbeta^*)}\bv'\bZ\bZ'\bv\ge\underline{\alpha}.
	\end{equation*}
\end{ass}

\begin{ass} \label{adm}
	The true coefficient $\bbeta^*$ is a feasible solution to (\ref{infty}), i.e.,
	\begin{equation*}
	\|\bZ'(-\bbeta^*, 1)\|_1\le \gamma_N, \quad \bbeta^*\in \scrB. 
	\end{equation*}
\end{ass}

\begin{ass} \label{subgaussian}
	$(\bx, y)$ is a centered sub-Gaussian random vector (see definition in the Appendix), i.e., it has
        zero mean and satisfies the following condition: 
	\begin{equation*}
	\vertiii{(\bx, y)}_{\psi_2}=\sup\limits_{\bu\in
          \mathbb{S}^{m}}\vertiii{(\bx, y)'\bu}_{\psi_2}\le \mu. 
	\end{equation*}
\end{ass}

\begin{ass} \label{eigen}
	The covariance matrix of $(\bx, y)$ has bounded positive
        eigenvalues. Set $\bGamma=\mathbb{E}[(\bx, y) (\bx, y)']$; then, 
	\begin{equation*}
	0<\lambda_{\text{min}} \triangleq \lambda_{\text{min}}(\bGamma)\le\lambda_{\text{max}}(\bGamma)\triangleq \lambda_{\text{max}}<\infty.
	\end{equation*}
\end{ass}

Notice that both $\underline{\alpha}$ in Assumption \ref{RE} and $\gamma_N$ in Assumption \ref{adm} are related to the random observation matrix $\bZ$. A probabilistic description for these two quantities will be provided later. We next present a preliminary result, similar to Lemma 2 in
\citet{chen2016alternating}, that bounds the $\ell_2$-norm of the estimation
bias in terms of a quantity that is related to the geometric structure
of the true coefficients. This result gives a rough idea on the factors that affect the estimation error, and shows the advantages of using the $\ell_1$-loss from the perspective of its dual norm. The bound derived in Theorem \ref{mainresult} is 
crude in the sense that it is a function of several random parameters that are related to the random observation matrix $\bZ$. This randomness will be described in a probabilistic way in the subsequent analysis. 
  
\begin{theorem} \label{mainresult} Suppose the true regression coefficient
  vector is $\bbeta^*$ and the solution to (\ref{infty}) is
  $\hat{\bbeta}$. For the set $\scrA(\bbeta^*)=\text{cone}\{\bv|\ \|(-\bbeta^*, 1) + \bv\|_* \le\|(-\bbeta^*, 1)\|_*\}\cap \mathbb{S}^m$,
  under Assumptions \ref{a1}, \ref{RE}, and \ref{adm}, we have:
\begin{equation} \label{l2norm}
\|\hat{\bbeta}-\bbeta^*\|_2\le
\frac{2R\gamma_N}{\underline{\alpha}}\Psi(\bbeta^*), 
\end{equation}
where $\Psi(\bbeta^*)=\sup_{\bv\in \scrA(\bbeta^*)}\|\bv\|_*$.
\end{theorem}

Notice that the bound in (\ref{l2norm}) does not explicitly depend on the sample size $N$. If we change to
the $\ell_2$-loss function, problem (\ref{infty}) will become: 
\begin{equation*} 
\begin{aligned}
\min_{\bbeta} & \quad \|(-\bbeta, 1)\|_* \\
\text{s.t.} & \quad \|(-\bbeta, 1)'\bZ\|_2\le \gamma_N,\\
& \quad \bbeta\in \scrB.
\end{aligned}
\end{equation*}
The proof of Theorem \ref{mainresult} still applies with slight modification. We will find out that in the case of $\ell_2$-loss, the estimation error bound is in the following form:
\begin{equation*} 
\|\hat{\bbeta}-\bbeta^*\|_2\le
\frac{2R\sqrt{N}\gamma_N}{\underline{\alpha}}\Psi(\bbeta^*).
\end{equation*}
Similarly, the $\ell_{\infty}$-loss, which considers only the maximum absolute loss among the samples, turns (\ref{infty}) into:
\begin{equation*} 
\begin{aligned}
\min_{\bbeta} & \quad \|(-\bbeta, 1)\|_* \\
\text{s.t.} & \quad \|(-\bbeta, 1)'\bZ\|_{\infty} \le \gamma_N,\\
& \quad \bbeta\in \scrB.
\end{aligned}
\end{equation*}
The corresponding bound becomes:
\begin{equation*}
\|\hat{\bbeta}-\bbeta^*\|_2\le
\frac{2RN\gamma_N}{\underline{\alpha}}\Psi(\bbeta^*).
\end{equation*}
We see that by using either $\ell_2$ or $\ell_{\infty}$-loss, an explicit dependency on $N$ is introduced. As a result, the estimation error bounds become worse. The reason is that for the $\ell_1$-loss function, its dual norm operator only picks out the maximum absolute coordinate and thus avoids the dependence on the dimension, which in our case is the sample size (see Eq.(\ref{re2})), whereas other norms, e.g., $\ell_2$-norm, sum over all the coordinates and thus introduce a dependence on $N$.

As mentioned earlier, (\ref{l2norm}) provides a random upper bound, revealed in $\underline{\alpha}$ and $\gamma_N$, that depends on the randomness in $\bZ$. We therefore would like to replace these two parameters by non-random quantities. The $\underline{\alpha}$ acts as the minimum eigenvalue of the matrix $\bZ \bZ'$ restricted to a subspace of $\mbb{R}^m$, and thus a proper substitute should be related to the minimum eigenvalue of the covariance matrix of $(\bx, y)$, i.e., the $\bGamma$ matrix (cf. Assumption \ref{eigen}), given that $(\bx, y)$ is zero mean. See Lemmas \ref{alphalem}, \ref{gaussianwidthlem} and \ref{alphacol} for the derivation.

\begin{lem} \label{alphalem} Consider the set $\scrA_{\bGamma}=\{\bw\in
  \mathbb{S}^{m}|\bGamma^{-1/2}\bw \in \text{cone}(\scrA(\bbeta^*))\}$,
  where $\scrA(\bbeta^*)$ is defined as in Theorem~\ref{mainresult}, and
  $\bGamma=\mathbb{E}[(\bx, y)(\bx, y)']$. Under Assumptions \ref{subgaussian} and
  \ref{eigen}, when the sample size $N\ge C_1\bar{\mu}^4
  (w(\scrA_{\bGamma}))^2$, where
  $\bar{\mu}=\mu\sqrt{\frac{1}{\lambda_{\text{min}}}}$, and
  $w(\scrA_{\bGamma})$ is the Gaussian width of $\scrA_{\bGamma}$, with
  probability at least $1-\exp(-C_2N/\bar{\mu}^4)$, we have
\begin{equation*}
  \bv'\bZ\bZ'\bv\ge \frac{N}{2}\bv'\bGamma\bv, \quad \forall \ \bv\in
  \scrA(\bbeta^*), 
\end{equation*}
where $C_1$ and $C_2$ are positive constants.
\end{lem}

Note that the sample size requirement stated in Lemma \ref{alphalem}
depends on the Gaussian width of $\scrA_{\bGamma}$, where $\scrA_{\bGamma}$ relates to $\scrA(\bbeta^*)$. The
following lemma shows that their Gaussian widths are also related. This relation is built upon the square root of the eigenvalues of $\bGamma$, which measures the extent to which $\scrA_{\bGamma}$ expands $\scrA(\bbeta^*)$.
\begin{lem}[Lemma 4 in
  \citet{chen2016alternating}] \label{gaussianwidthlem} Let $\mu_0$ be
  the $\psi_2$-norm of a standard Gaussian random vector $\bg \in
  \mathbb{R}^{m}$, and $\scrA_{\bGamma}$, $\scrA(\bbeta^*)$ be defined
  as in Lemma \ref{alphalem}. Then, under Assumption \ref{eigen},
	\begin{equation*}
	w(\scrA_{\bGamma})\le C_3\mu_0\sqrt{\frac{\lambda_{\text{max}}}{\lambda_{\text{min}}}}\Bigl(w(\scrA(\bbeta^*))+3\Bigr),
	\end{equation*}
for some positive constant $C_3$.
\end{lem}

Combining Lemmas \ref{alphalem} and \ref{gaussianwidthlem}, and expressing the covariance matrix $\bGamma$ using its eigenvalues, we arrive at the following result.
\begin{col} \label{alphacol} Under Assumptions \ref{subgaussian} and
  \ref{eigen}, and the conditions in Lemmas \ref{alphalem} and
  \ref{gaussianwidthlem}, when $N\ge \bar{C_1}\bar{\mu}^4
  \mu_0^2\cdot\frac{\lambda_{\text{max}}}{\lambda_{\text{min}}}\Bigl(w(\scrA(\bbeta^*))+3\Bigr)^2$,
 with probability at least $1-\exp(-C_2N/\bar{\mu}^4)$,
\begin{equation*}
\bv'\bZ\bZ'\bv\ge \frac{N\lambda_{\text{min}}}{2}, \quad \forall \
\bv\in \scrA(\bbeta^*), 
\end{equation*}
where $\bar{C_1}$ and $C_2$ are positive constants.
\end{col}

Next we derive the smallest possible value of $\gamma_N$
such that $\bbeta^*$ is feasible. The derivation uses the dual norm operator of the $\ell_1$-loss, resulting in a bound that depends on the Gaussian width of the unit ball in the dual norm space ($\|\cdot\|_\infty$). See Lemma
\ref{gammalem} for details.
\begin{lem} \label{gammalem}
Under Assumptions \ref{2norm} and \ref{subgaussian}, for any feasible
$\bbeta$, with probability at least
$1-C_4\exp(-\frac{C_5^2(w(\scrB_u))^2}{4\rho^2})$, 
\begin{equation*}
\|(-\bbeta, 1)'\bZ\|_1\le C\mu \bar{B}_2w(\scrB_u),
\end{equation*}
where $\scrB_u$ is the unit ball of norm $\|\cdot\|_\infty$,
$\rho=\sup_{\bv\in \scrB_u}\|\bv\|_2$, and $C_4, C_5, C$ positive constants.
\end{lem}

We note that for other loss functions, e.g., the $\ell_2$ and $\ell_{\infty}$ losses, similar results can be obtained, where $\scrB_u$ is defined to be the unit $\|\cdot\|_*^{\text{loss}}$-ball in $\mbb{R}^m$, with $\|\cdot\|_*^{\text{loss}}$ being the dual norm of the loss. Combining Theorem \ref{mainresult}, Corollary \ref{alphacol} and Lemma
\ref{gammalem}, we have the following main performance guarantee result
that bounds the estimation bias of the solution to (\ref{infty}).
\begin{theorem}
  Under Assumptions \ref{a1}, \ref{2norm}, \ref{RE}, \ref{adm},
  \ref{subgaussian}, \ref{eigen}, and the conditions of
  Theorem~\ref{mainresult}, Corollary \ref{alphacol} and Lemma
  \ref{gammalem}, when $N\ge \bar{C_1}\bar{\mu}^4
  \mu_0^2\cdot\frac{\lambda_{\text{max}}}{\lambda_{\text{min}}}
  \Bigl(w(\scrA(\bbeta^*))+3\Bigr)^2$,  
  with probability at least
  $1-\exp(-C_2N/\bar{\mu}^4)-C_4\exp(-C_5^2 (w(\scrB_u))^2/(4\rho^2))$,
\begin{equation} \label{finalbound} \|\hat{\bbeta}-\bbeta^*\|_2\le
  \frac{\bar{C}R\bar{B}_2\mu}{N\lambda_{\text{min}}}
  w(\scrB_u)\Psi(\bbeta^*).
\end{equation}
\end{theorem}
 
From (\ref{finalbound}) we see that the bias is decreased as the sample size increases and the
uncertainty embedded in $(\bx, y)$ (revealed in $R$ and $\mu$) is reduced. The estimation error bound depends on the geometric structure of the true
coefficients, defined using the dual norm space of the Wasserstein metric, the Gaussian width of the unit $\|\cdot\|_*^{\text{loss}}$-ball in $\mbb{R}^m$, and the minimum eigenvalue of the covariance matrix of $(\bx, y)$, with a convergence rate $1/N$ for the $\ell_1$-loss we applied. As mentioned earlier, other loss functions may incur a dependence on $N$ in the numerator of the bound, thus resulting in a slower convergence rate, which substantiates the benefit of using an $\ell_1$-loss function.
\section{Simulation Experiments on Synthetic Datasets} \label{s4}
In this section we will explore the robustness of the Wasserstein formulation in terms of its {\em Absolute Deviation (AD)} loss function and the dual norm regularizer on the {\em extended regression coefficient} $(-\bbeta, 1)$. Recall that our Wasserstein formulation is in the following form:
\begin{equation} \label{qcp2}
\inf\limits_{\bbeta\in \scrB} \frac{1}{N}\sum\limits_{i=1}^N|y_i - \bx_i'\bbeta| + \epsilon\|(-\bbeta, 1)\|_*.
\end{equation}
We will focus on the following three aspects of this formulation:
\begin{enumerate}
	\item How to choose a proper norm $\|\cdot\|$ for the Wasserstein metric?
	\item Why do we penalize the extended regression coefficient $(-\bbeta, 1)$ rather than $\bbeta$?
	\item What is the advantage of the AD loss compared to the {\em Squared Residuals (SR)} loss?
\end{enumerate} 

To answer Question 1, we will connect the choice of $\|\cdot\|$ for the Wasserstein metric with the characteristics/structures of the data $(\bx, y)$. Specifically, we will design two sets of experiments, one with a dense regression coefficient $\bbeta^*$, where all coordinates of $\bx$ play a role in determining the value of the response $y$, and another with a sparse $\bbeta^*$ implying that only a few predictors are relevant/important in predicting $y$. Two Wasserstein formulations will be tested and compared, one induced by the $\|\cdot\|_2$ (Wasserstein $\ell_2$), which leads to an $\ell_2$-regularizer in (\ref{qcp2}), and the other one induced by the $\|\cdot\|_{\infty}$ (Wasserstein $\ell_{\infty}$) and resulting in an $\ell_1$-regularizer in (\ref{qcp2}). Intuitively, and based on the past experience in implementing the regularization techniques, the Wasserstein $\ell_2$ should outperform the Wasserstein $\ell_{\infty}$ in the dense setting, while in the sparse setting, the reverse is true. Researchers have well identified the sparsity inducing property of the $\ell_1$-regularizer and provided a nice geometrical interpretation for it \citep{friedman2001elements}. Here, we try to offer a different explanation from the perspective of the Wasserstein DRO formulation, through projecting the sparsity of $\bbeta^*$ onto the $(\bx, y)$ space and establishing a {\em sparse} distance metric that only extracts a subset of coordinates from $(\bx, y)$ to measure the closeness between samples. 

For the second question, we first note that if the Wasserstein metric is induced by the following metric $s_c$:
\begin{equation*}
	s_c(\bx, y) = \|(\bx, cy)\|_2,
\end{equation*}
for a positive constant $c$, then as $c \rightarrow \infty$, the resulting Wasserstein DRO formulation becomes:
\begin{equation*}
\inf\limits_{\bbeta\in \scrB} \frac{1}{N}\sum\limits_{i=1}^N|y_i - \bx_i'\bbeta| + \epsilon\|\bbeta\|_2,
\end{equation*}
which is the $\ell_2$-regularized LAD. This can be proved by recognizing that $s_c(\bx, y) = \|(\bx, y)\|_{\bM}$, with $\bM \in \mbb{R}^{m \times m}$ a diagonal matrix whose diagonal elements are $(1, \ldots, 1, c^2)$, and then applying (\ref{wqcp}). Alternatively, if we let 
$$s_c(\bx, y) = \|(\bx, cy)\|_{\infty},$$
it can be shown that as $c \rightarrow \infty$, the corresponding Wasserstein formulation becomes:
\begin{equation*}
\inf\limits_{\bbeta\in \scrB} \frac{1}{N}\sum\limits_{i=1}^N|y_i - \bx_i'\bbeta| + \epsilon\|\bbeta\|_1,
\end{equation*}
which is the $\ell_1$-regularized LAD (see proof in the Appendix). It follows that regularizing over $\bbeta$ implies an infinite transportation cost along $y$. In other words, for two data points $(\bx_1, y_1)$ and $(\bx_2, y_2)$, if $y_1 \neq y_2$, then they are considered to be infinitely far away. By contrast, our Wasserstein formulation, which regularizes over the extended regression coefficient $(-\bbeta, 1)$, stems from a finite cost along $y$ that is equally weighted with $\bx$. We will see the disadvantages of penalizing only $\bbeta$ in the analysis of the experimental results.

To answer Question 3, we will compare with several commonly used regression models that employ the SR loss function, e.g., ridge regression \citep{hoerl1970ridge}, LASSO \citep{tibshirani1996regression}, and {\em Elastic Net (EN)} \citep{zou2005regularization}. We will also compare against M-estimation \citep{huber1964robust, huber1973robust}, which uses a variant of the SR loss and is equivalent to solving a weighted least squares problem, where the weights are determined by the residuals. These models will be compared under two different experimental setups, one involving adversarial perturbations in both $\bx$ and $y$, and the other with perturbations only in $\bx$. The purpose is to investigate the behavior of these approaches when the noise in $y$ is substantially reduced. As shown by Fig. \ref{absloss}, compared to the SR loss, the AD loss is less vulnerable to large residuals, and hence, it is advantageous in the scenarios where large perturbations appear in $y$. We are interested in studying whether its performance is consistently good when the corruptions appear mainly in $\bx$.

We next describe the data generation process. Each training sample has a probability $q$ of being drawn from the outlying distribution, and a probability $1-q$ of being drawn from the true (clean) distribution. Given the true regression coefficient $\bbeta^*$, we generate the training data as follows:
\begin{itemize}
	\item Generate a uniform random variable on $[0, 1]$. If it is no larger than $1-q$, generate a clean sample as follows:
	\begin{enumerate}
		\item Draw the predictor $\bx \in \mbb{R}^{m-1}$ from the normal distribution $N_{m-1} (\mathbf{0}, \bSigma)$, where $\bSigma$ 
		is the covariance matrix of $\bx$, which is just the top left block of the matrix $\bGamma$ in Assumption \ref{eigen}. Specifically, $\bGamma=\mathbb{E}[(\bx, y) (\bx, y)']$ is equal to
		\begin{equation*}
		\bGamma = 
		\begin{pmatrix}
		& \bSigma   & \bSigma \bbeta^* \\
		& (\bbeta^*)'\bSigma & (\bbeta^*)' \bSigma \bbeta^* + \sigma^2
		\end{pmatrix},
		\end{equation*}
		with $\sigma^2$ being the variance of the noise term. In our implementation, $\bSigma$ has diagonal elements equal to $1$ (unit variance) and off-diagonal elements equal to $\rho$, with $\rho$ the correlation between predictors. 
		\item Draw the response variable $y$ from $N(\bx' \bbeta^*, \sigma^2)$.
	\end{enumerate}
	\item Otherwise, depending on the experimental setup, generate an outlier that is either:
	\begin{itemize}
		\item Abnormal in both $\bx$ and $y$, with outlying distribution:
		\begin{enumerate}
			\item $\bx \sim N_{m-1} (\mathbf{0}, \bSigma) + N_{m-1} (5\mathbf{e}, \mathbf{I})$, or $\bx \sim N_{m-1} (\mathbf{0}, \bSigma) + N_{m-1} (\mathbf{0}, 0.25\mathbf{I})$;
			\item $y \sim N(\bx' \bbeta^*, \sigma^2) + 5\sigma$.
		\end{enumerate}
	    \item Abnormal only in $\bx$: 
	    \begin{enumerate}
	    	\item $\bx \sim N_{m-1} (\mathbf{0}, \bSigma) + N_{m-1} (5\mathbf{e}, \mathbf{I})$;
	    	\item $y \sim N(\bx' \bbeta^*, \sigma^2)$. 
	    \end{enumerate}
	\end{itemize}	
    \item Repeat the above procedure for $N$ times, where $N$ is the size of the training set.
\end{itemize}

To test the generalization ability of various formulations, we generate a test dataset containing $M$ samples from the clean distribution. It is worth noting that only clean samples are included in the test set, since we only care about the prediction accuracy on clean data points, and our estimator is supposed to be consistent with the clean distribution and stay away from the outlying one. We are interested in studying the performance of various methods as the following factors are varied:
\begin{itemize}
	\item {\em Signal to Noise Ratio (SNR)}, defined as:
	\begin{equation*}
	\text{SNR} = \frac{(\bbeta^*)'\bSigma \bbeta^*}{\sigma^2},
	\end{equation*}
	which is equally spaced between $0.05$ and $2$ on a log scale.
	\item The correlation between predictors: $\rho$, which takes values in $(0.1, 0.2, \ldots, 0.9)$.
\end{itemize}  	
The performance metrics we use include:
\begin{itemize}
	\item {\em Mean Squared Error (MSE)} on the test dataset, which is defined to be $\sum_{i=1}^M(y_i - \bx_i'\hat{\bbeta})^2/M$, with $\hat{\bbeta}$ being the estimate of $\bbeta^*$ obtained from the training set, and $(\bx_i, y_i), \ i=1, \ldots, M,$ being the observations from the test dataset;
	\item {\em Relative Risk (RR)} of $\hat{\bbeta}$ defined as:
	\begin{equation*}
	\text{RR}(\hat{\bbeta}) \triangleq \frac{(\hat{\bbeta} - \bbeta^*)'\mathbf{\Sigma}(\hat{\bbeta} - \bbeta^*)}{(\bbeta^*)' \mathbf{\Sigma} \bbeta^*}.
	\end{equation*}
	\item {\em Relative Test Error (RTE)} of $\hat{\bbeta}$ defined as:
	\begin{equation*}
	\text{RTE}(\hat{\bbeta}) \triangleq \frac{(\hat{\bbeta} - \bbeta^*)'\bSigma (\hat{\bbeta} - \bbeta^*) + \sigma^2}{\sigma^2}.
	\end{equation*}
	\item {\em Proportion of Variance Explained (PVE)} of $\hat{\bbeta}$ defined as:
	\begin{equation*}
	\text{PVE}(\hat{\bbeta}) \triangleq 1 - \frac{(\hat{\bbeta} - \bbeta^*)'\bSigma (\hat{\bbeta} - \bbeta^*) + \sigma^2}{(\bbeta^*)' \mathbf{\Sigma} \bbeta^* + \sigma^2}.
	\end{equation*}
\end{itemize} 
For the metrics that evaluate the accuracy of the estimator, i.e., the RR, RTE and PVE, we list below two types of scores, one achieved by the best possible estimator $\hat{\bbeta} = \bbeta^*$, called the perfect score, and the other one achieved by the null estimator $\hat{\bbeta} = 0$, called the null score. 
\begin{itemize}
	\item RR: a perfect score is 0 and the null score is 1.
	\item RTE: a perfect score is 1 and the null score is SNR+1.
	\item PVE: a perfect score is $\frac{\text{SNR}}{\text{SNR}+1}$, and the null score is 0.
\end{itemize} 
 
During the training process, all the regularization parameters are tuned on a separate validation dataset. Specifically, we divide all the $N$ training samples into two sets, dataset 1 and dataset 2 (validation set). For a pre-specified range of values for the penalty parameters, dataset 1 is used to train the models and derive $\hat{\bbeta}$, and the performance of $\hat{\bbeta}$ is evaluated on dataset 2. We choose the regularization parameter that yields the minimum {\em Median Absolute Deviation (MAD)} on the validation set. Using MAD as a selection criterion serves to hedge against the potentially large noise in the validation samples. As to the range of values for the tuned parameters, we borrow ideas from \cite{hastie2017extended}, where the LASSO was tuned over $50$ values ranging from $\lambda_m = \|\bX'\by\|_{\infty}$ to a small fraction of $\lambda_m$ on a log scale, with $\bX \in \mbb{R}^{N \times (m-1)}$ the design matrix whose $i$-th row is $\bx_i'$, and $\by = (y_1, \ldots, y_N)$ the response vector. In our experiments, this range is properly adjusted for procedures that use the AD loss. Specifically, for Wasserstein $\ell_2$ and $\ell_{\infty}$, $\ell_1$- and $\ell_2$-regularized LAD, the range of values for the regularization parameter is: 
$$\sqrt{\exp\biggl(\text{lin}\Bigl(\log(0.005*\|\bX'\by\|_{\infty}),\log(\|\bX'\by\|_{\infty}),50\Bigr)\biggr)},$$
where $\text{lin}(a, b, n)$ is a function that takes in scalars $a$, $b$ and $n$ (integer) and outputs a set of $n$ values equally spaced between $a$ and $b$; the $\exp$ function is applied elementwise to a vector. The square root operator is in consideration of the AD loss that is the square root of the SR loss if evaluated on a single sample. 

The regularization coefficient $\epsilon$ in formulation (\ref{qcp}), which is the radius of the Wasserstein ball, allows for a more efficient tuning procedure. It has been noted in \citet{Pey15} that for a large enough sample size, $\epsilon$ is inversely proportional to $N^{1/m}$. This proportionality could be used as a guidance on setting $\epsilon$, where only the proportional factor needs to be tuned (using cross-validation or a separate validation dataset as described earlier). In our implementation, given the small size of the simulated datasets, we will still adopt the validation dataset approach to tune the regularization parameter. 

\subsection{Dense $\bbeta^*$, outliers in both $\bx$ and $y$} \label{densexy}
\begin{figure}[p] 
	\begin{subfigure}{.5\textwidth}
		\centering
		\includegraphics[width=1.0\textwidth]{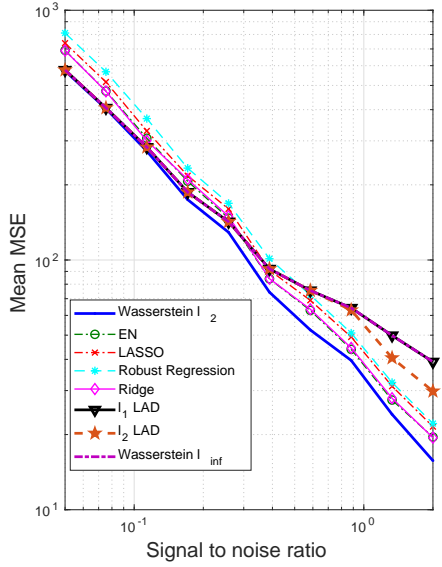}
		\caption{\small{Mean Squared Error.}}
	\end{subfigure}
	\begin{subfigure}{.5\textwidth}
		\centering
		\includegraphics[width=1.0\textwidth]{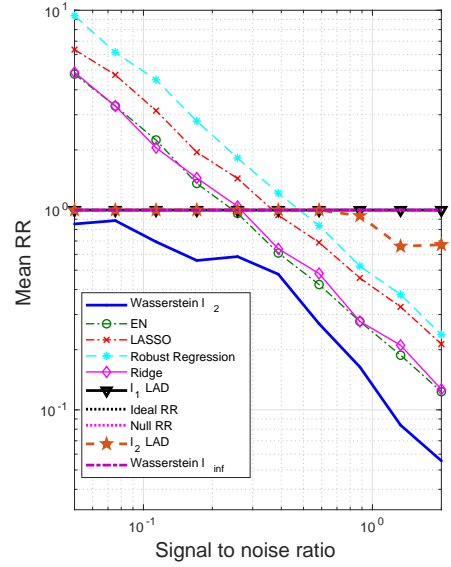}
		\caption{\small{Relative risk.}}
	\end{subfigure}
	
	\begin{subfigure}{.5\textwidth}
		\centering
		\includegraphics[width=1.0\textwidth]{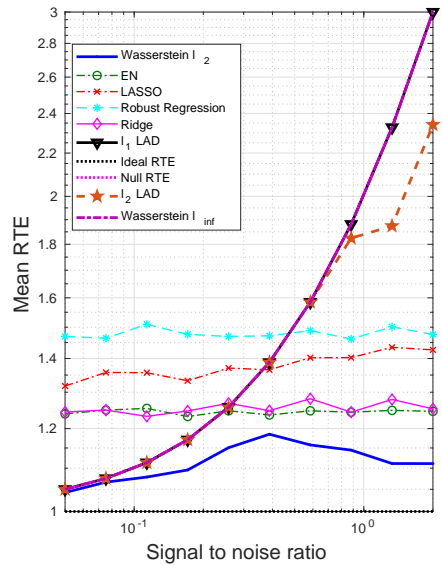}
		\caption{\small{Relative test error.}}
	\end{subfigure}%
	\begin{subfigure}{.5\textwidth}
		\centering
		\includegraphics[width=1.0\textwidth]{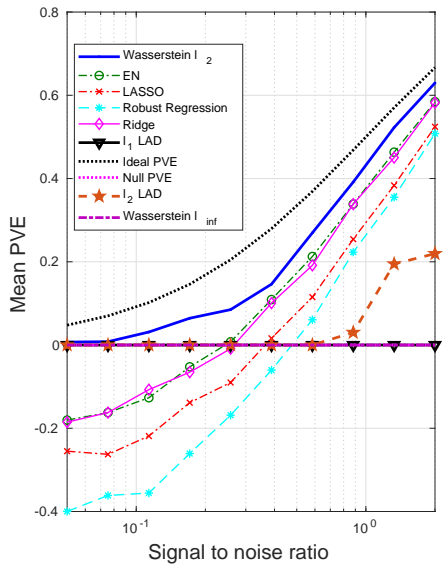}
		\caption{\small{Proportion of variance explained.}}
	\end{subfigure}
	\caption{The impact of SNR on the performance metrics: dense $\bbeta^*$, outliers in both $\bx$ and $y$.}
	\label{snr-1}
\end{figure}

\begin{figure}[p] 
	\begin{subfigure}{.5\textwidth}
		\centering
		\includegraphics[width=1.0\textwidth]{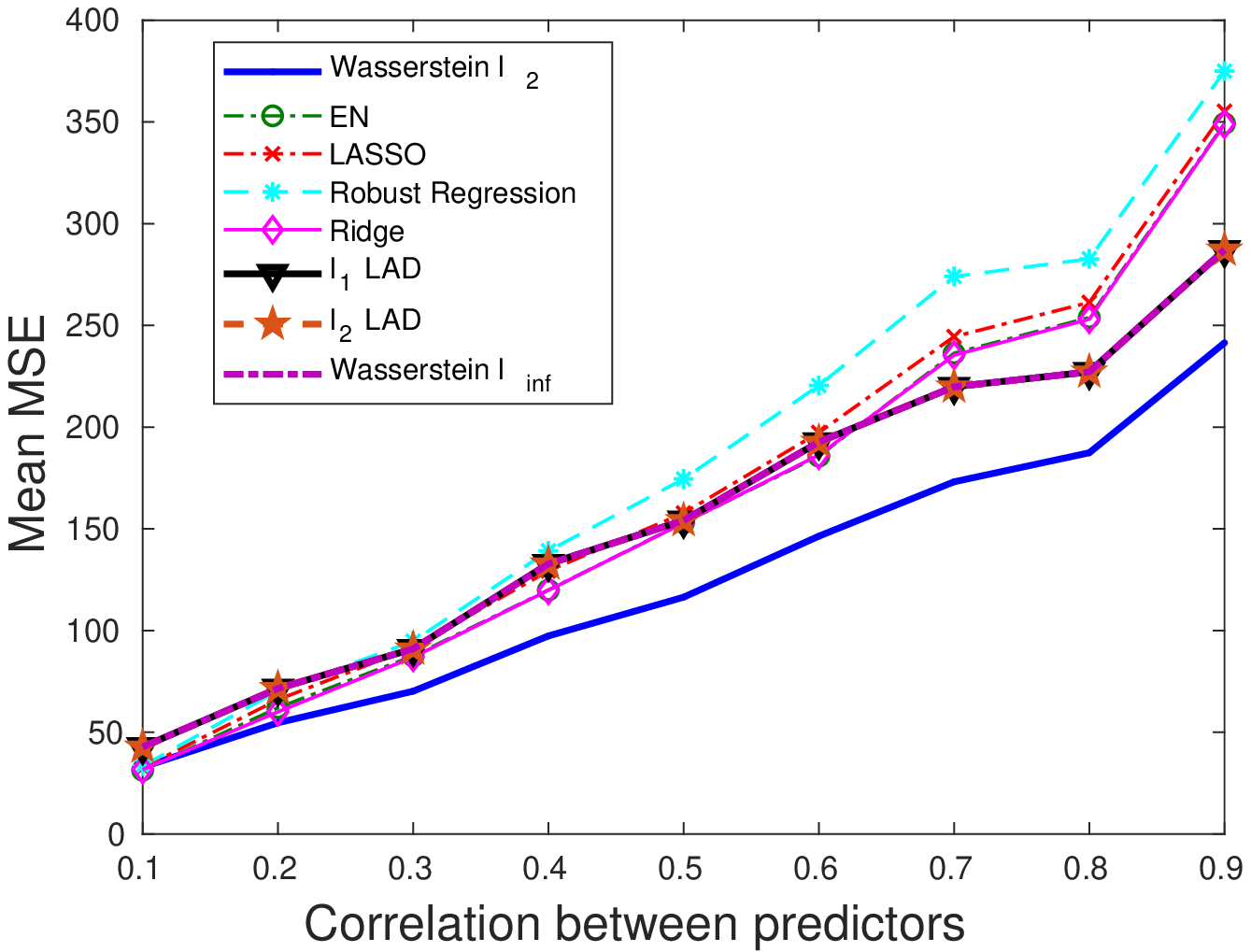}
		\caption{\small{Mean Squared Error.}}
	\end{subfigure}
	\begin{subfigure}{.5\textwidth}
		\centering
		\includegraphics[width=1.0\textwidth]{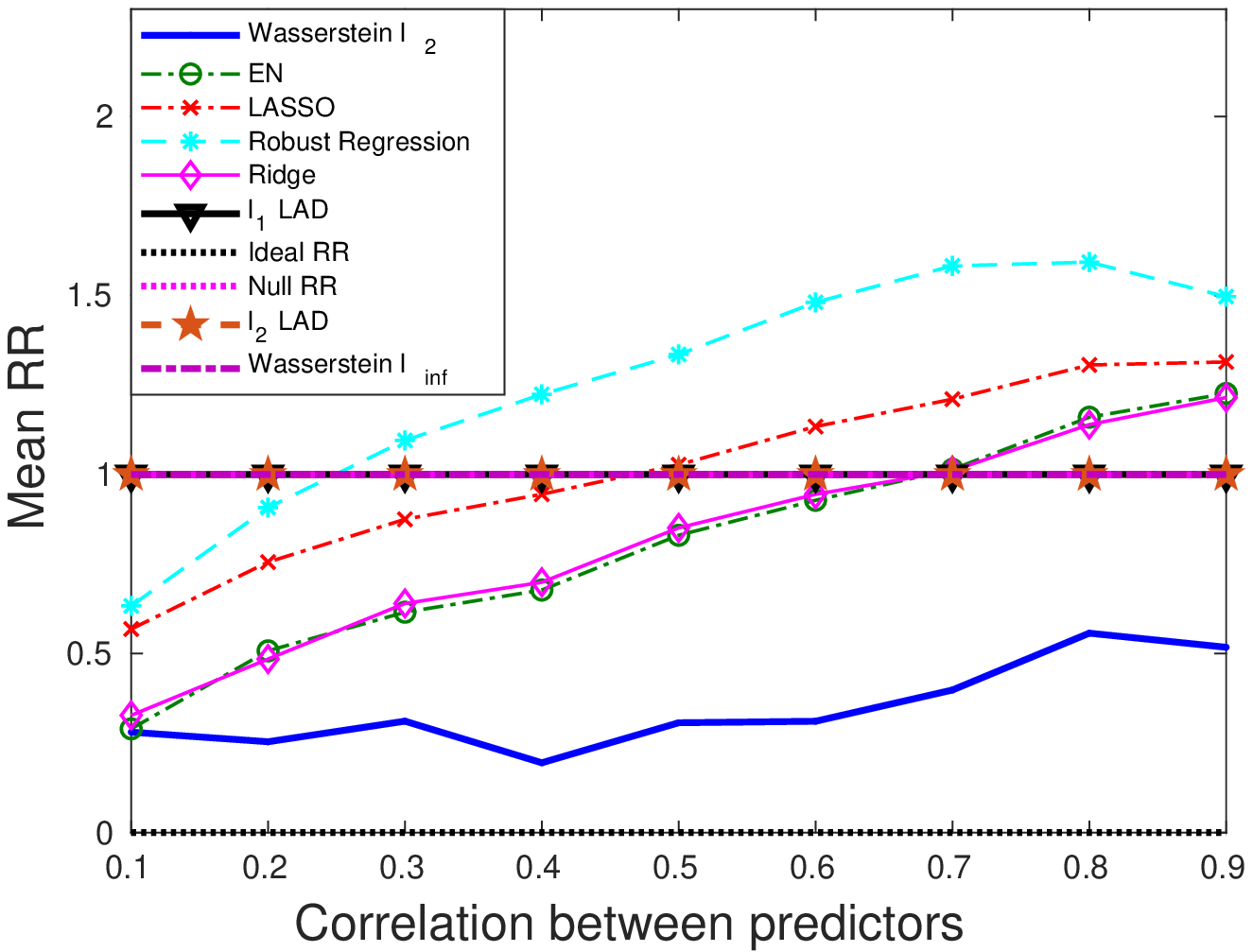}
		\caption{\small{Relative risk.}}
	\end{subfigure}
	
	\begin{subfigure}{.5\textwidth}
		\centering
		\includegraphics[width=1.0\textwidth]{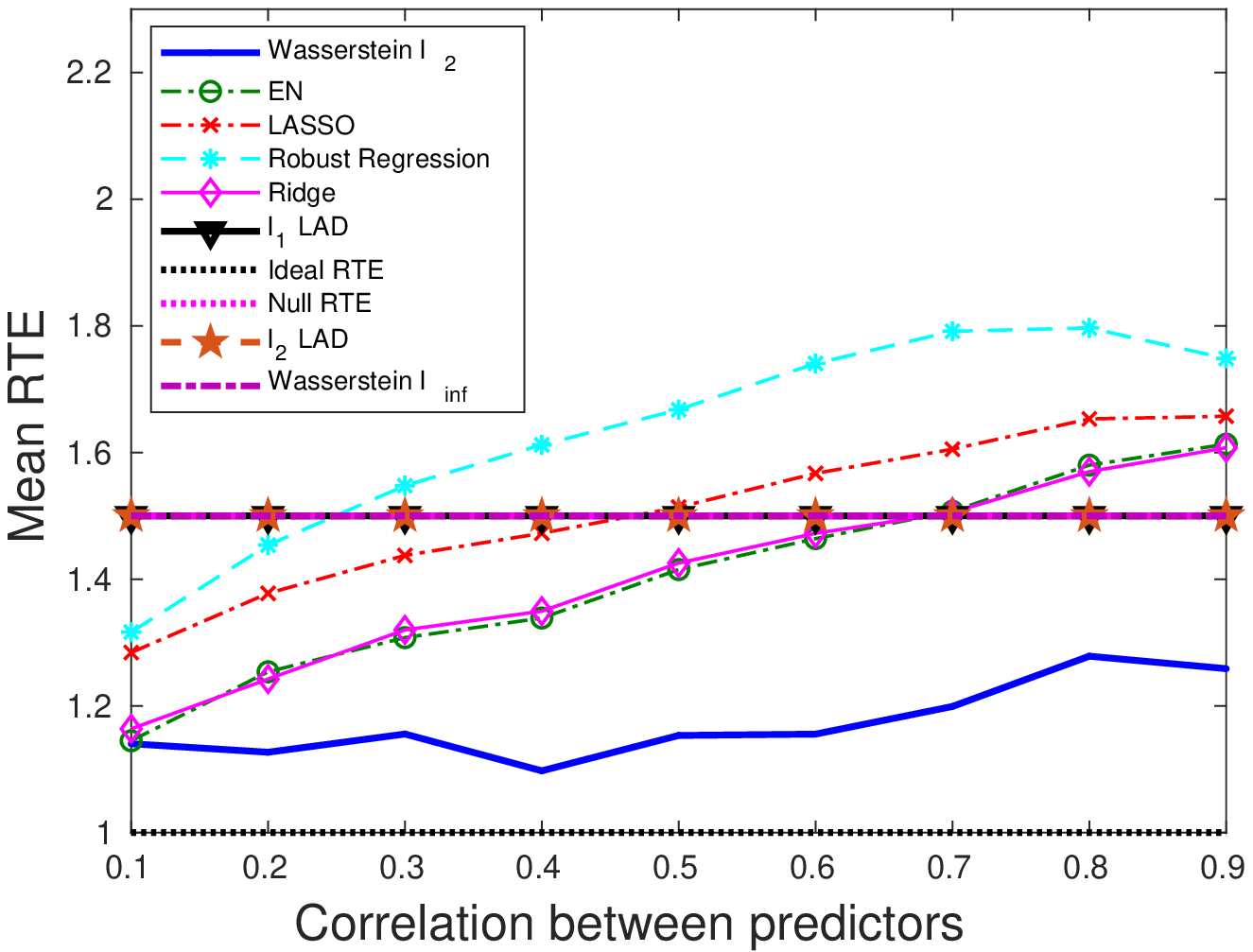}
		\caption{\small{Relative test error.}}
	\end{subfigure}%
	\begin{subfigure}{.5\textwidth}
		\centering
		\includegraphics[width=1.0\textwidth]{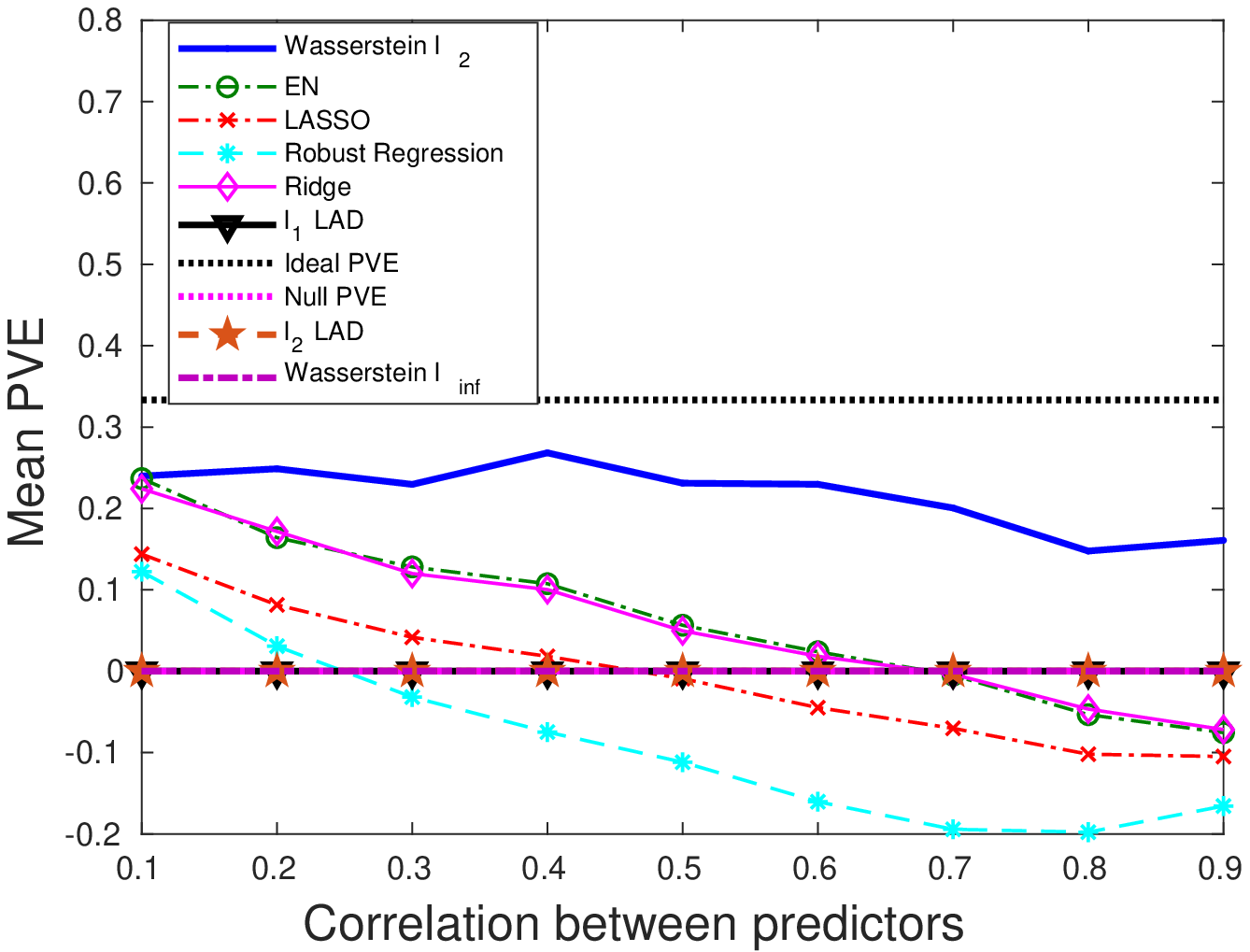}
		\caption{\small{Proportion of variance explained.}}
	\end{subfigure}
	\caption{The impact of predictor correlation on the performance metrics: dense $\bbeta^*$, outliers in both $\bx$ and $y$.}
	\label{corr-1}
\end{figure}

In this subsection, we choose a dense regression coefficient $\bbeta^*$, set the intercept $\beta_0^* = 0.3$, and the coefficient for each predictor $x_i$ to be $\beta_i^* = 0.5, i = 1, \ldots, 20$. The adversarial perturbations are present in both $\bx$ and $y$. Specifically, the outlying distribution is described by:
\begin{enumerate}
	\item $\bx \sim N_{m-1} (\mathbf{0}, \bSigma) + N_{m-1} (5\mathbf{e}, \mathbf{I})$;
	\item $y \sim N(\bx' \bbeta^*, \sigma^2) + 5\sigma$.
\end{enumerate}
We generate 10 datasets consisting of $N = 100, M = 60$ observations. The probability of a training sample being drawn from the outlying distribution is $q = 30\%$. The mean values of the performance metrics (averaged over the 10 datasets), as we vary the SNR and the correlation between predictors, are shown in Figs. \ref{snr-1} and \ref{corr-1}. Note that when SNR is varied, the correlation between predictors is set to $0.8$ times a random noise uniformly distributed on the interval $[0.2, 0.4]$. When the correlation $\rho$ is varied, the SNR is fixed to $0.5$.

It can be seen that as the SNR decreases or the correlation between the predictors increases, the estimation problem becomes harder, and the performance of all approaches gets worse. In general the Wasserstein $\ell_2$ achieves the best performance in terms of all four metrics. Specifically, 
\begin{itemize}
	\item It is better than the $\ell_2$-regularized LAD, which assumes an infinite transportation cost along $y$.
	\item It is better than the Wasserstein $\ell_{\infty}$ and $\ell_1$-regularized LAD which use the $\ell_1$-regularizer.
	\item It is better than the approaches that use the SR loss function.
\end{itemize}

Empirically we have found out that in most cases, the approaches that use the AD loss, including the $\ell_1$- and $\ell_2$-regularized LAD, and the Wasserstein $\ell_{\infty}$ formulation, drive all the coordinates of $\bbeta$ to zero, due to the relatively small magnitude of the AD loss compared to the norm of the coefficient, so that the regularizer dominates the solution. The approaches that use the SR loss, e.g., ridge regression and EN, do not exhibit such a problem, since the squared residuals weaken the dominance of the regularization term. 

Overall the $\ell_2$-regularizer outperforms the $\ell_1$-regularizer, since the true regression coefficient is dense, which implies that a proper distance metric on the $(\bx, y)$ space should take into account all the coordinates. From the perspective of the Wasserstein DRO framework, the $\ell_1$-regularizer corresponds to an $\|\cdot\|_{\infty}$-based distance metric on the $(\bx, y)$ space that only picks out the most influential coordinate to determine the closeness between data points, which in our case is not reasonable since every coordinate plays a role (reflected in the dense $\bbeta^*$). In contrast, if $\bbeta^*$ is sparse, using the $\|\cdot\|_{\infty}$ as a distance metric on $(\bx, y)$ is more appropriate. A more detailed discussion of this will be presented in Sections \ref{sparsexy} and \ref{sparsex}.

\subsection{Dense $\bbeta^*$, outliers only in $\bx$} \label{densex}
In this subsection we will experiment with the same $\bbeta^*$ as in Section \ref{densexy}, but with perturbations only in $\bx$, i.e., for a given $\bx$ of the outlier, the corresponding $y$ value is drawn in the same way as the clean samples. Our goal is to investigate the performance of the Wasserstein formulation when the response $y$ is not subjected to large perturbations. The motivation for introducing the AD loss in the Wasserstein formulation is to hedge against large residuals, as illustrated in Fig. \ref{absloss}. We are interested in comparing the AD and SR loss functions when the residuals have moderate magnitudes.
 
\begin{figure}[p] 
	\begin{subfigure}{.5\textwidth}
		\centering
		\includegraphics[width=1.0\textwidth]{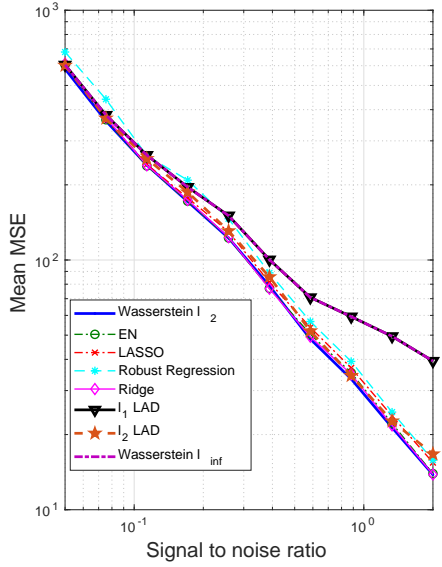}
		\caption{\small{Mean Squared Error.}}
	\end{subfigure}
	\begin{subfigure}{.5\textwidth}
		\centering
		\includegraphics[width=1.0\textwidth]{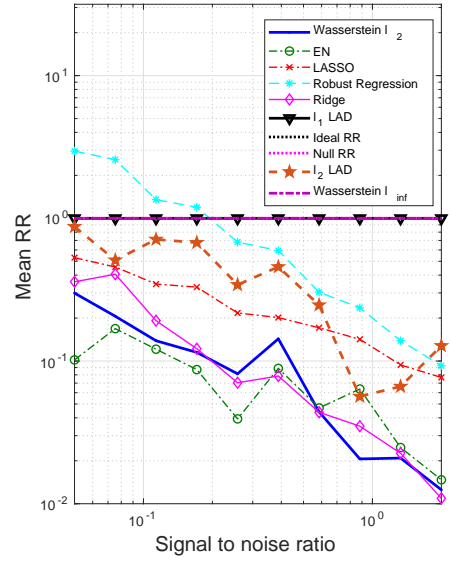}
		\caption{\small{Relative risk.}}
	\end{subfigure}
	
	\begin{subfigure}{.5\textwidth}
		\centering
		\includegraphics[width=1.0\textwidth]{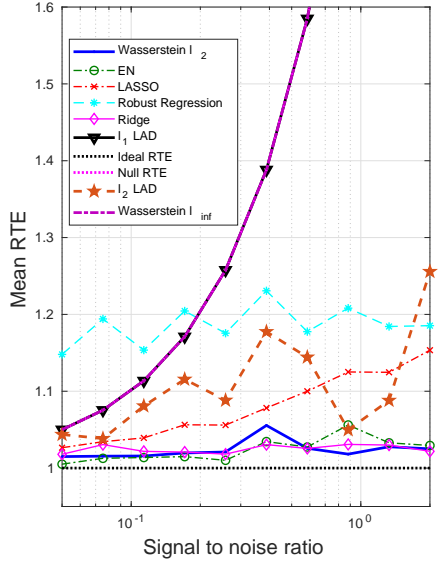}
		\caption{\small{Relative test error.}}
	\end{subfigure}%
	\begin{subfigure}{.5\textwidth}
		\centering
		\includegraphics[width=1.0\textwidth]{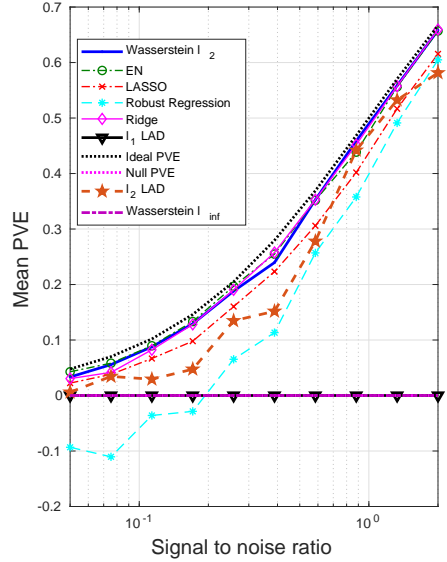}
		\caption{\small{Proportion of variance explained.}}
	\end{subfigure}
	\caption{The impact of SNR on the performance metrics: dense $\bbeta^*$, outliers only in $\bx$.}
	\label{snr-2}
\end{figure}

\begin{figure}[p] 
	\begin{subfigure}{.5\textwidth}
		\centering
		\includegraphics[width=1.0\textwidth]{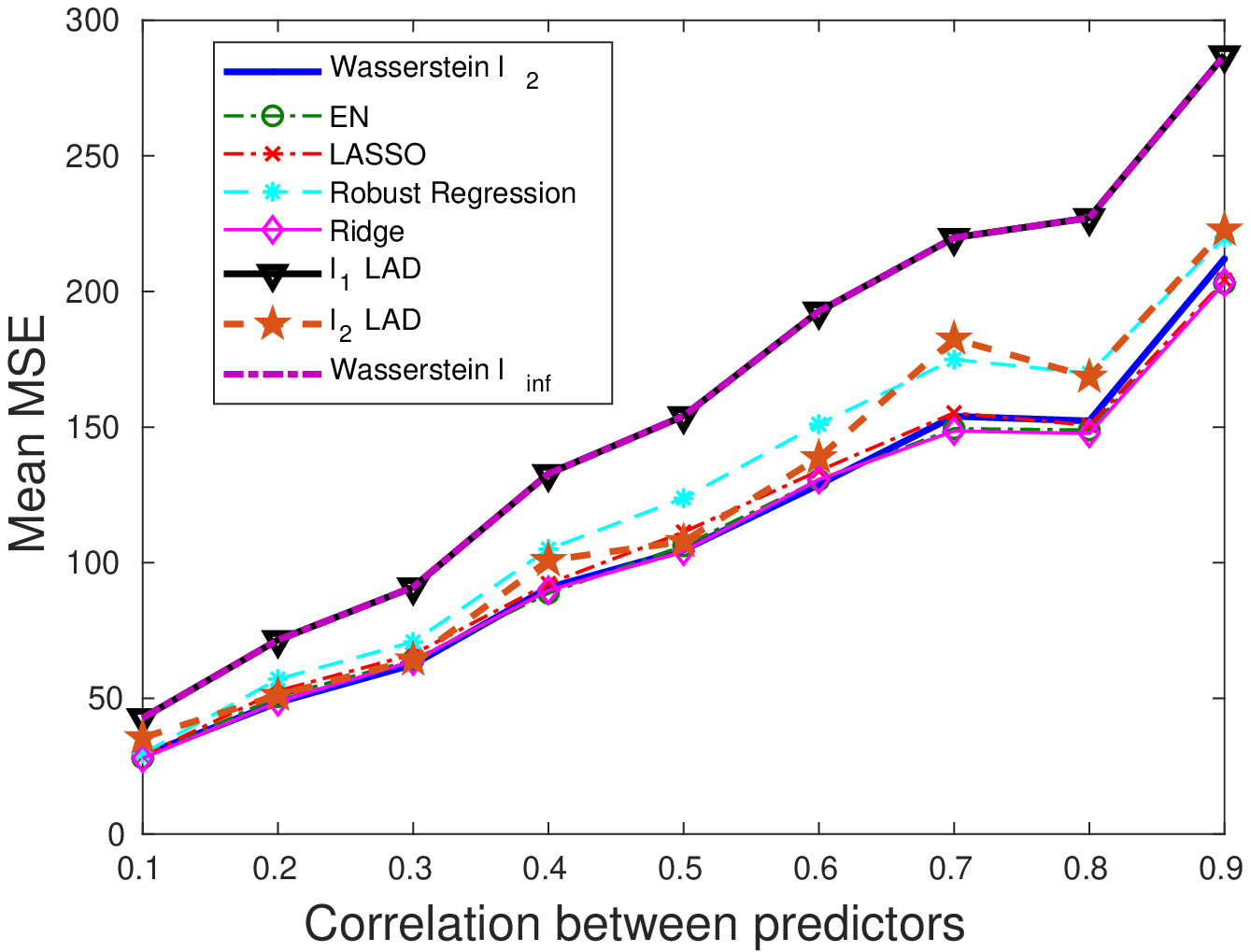}
		\caption{\small{Mean Squared Error.}}
	\end{subfigure}
	\begin{subfigure}{.5\textwidth}
		\centering
		\includegraphics[width=1.0\textwidth]{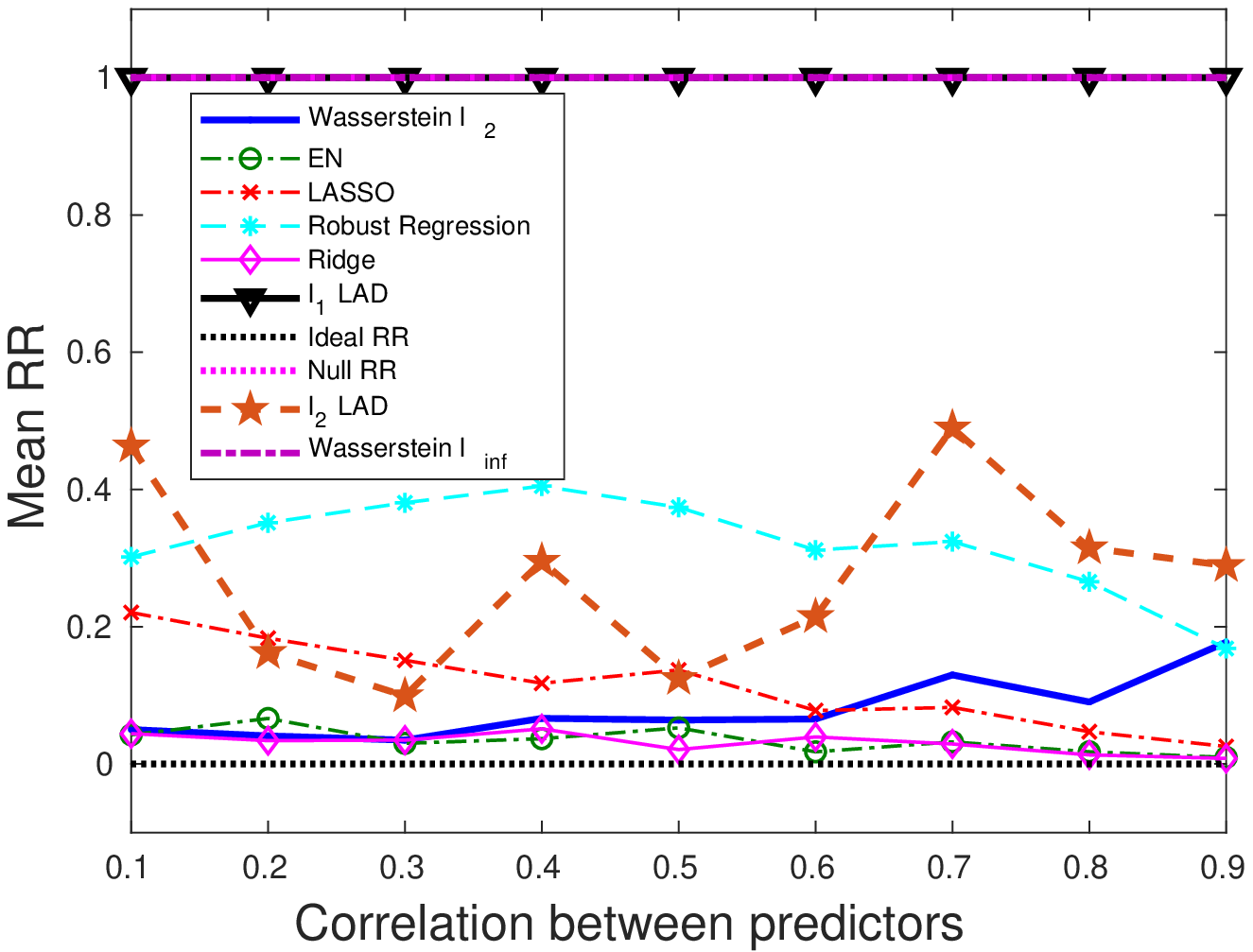}
		\caption{\small{Relative risk.}}
	\end{subfigure}
	
	\begin{subfigure}{.5\textwidth}
		\centering
		\includegraphics[width=1.0\textwidth]{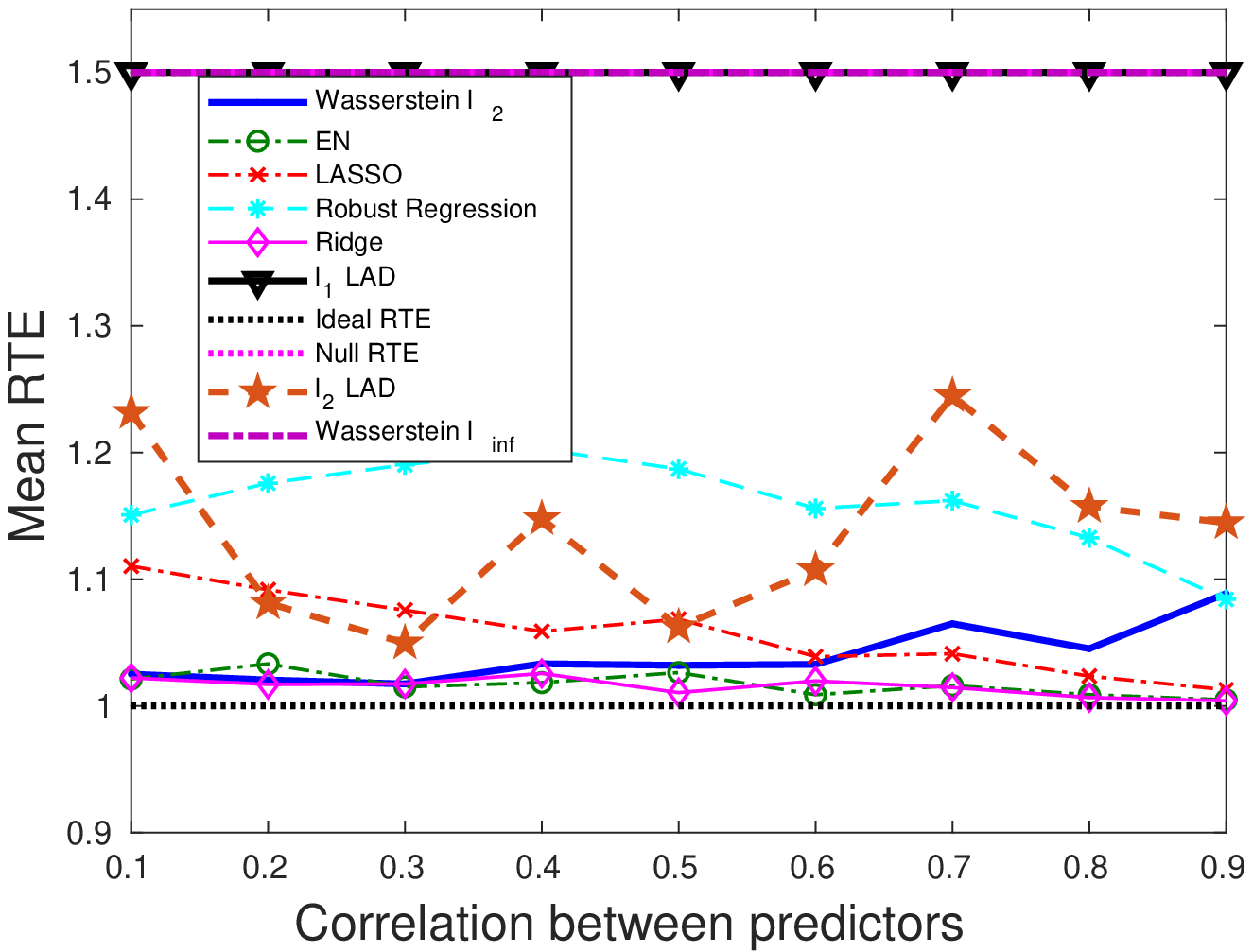}
		\caption{\small{Relative test error.}}
	\end{subfigure}%
	\begin{subfigure}{.5\textwidth}
		\centering
		\includegraphics[width=1.0\textwidth]{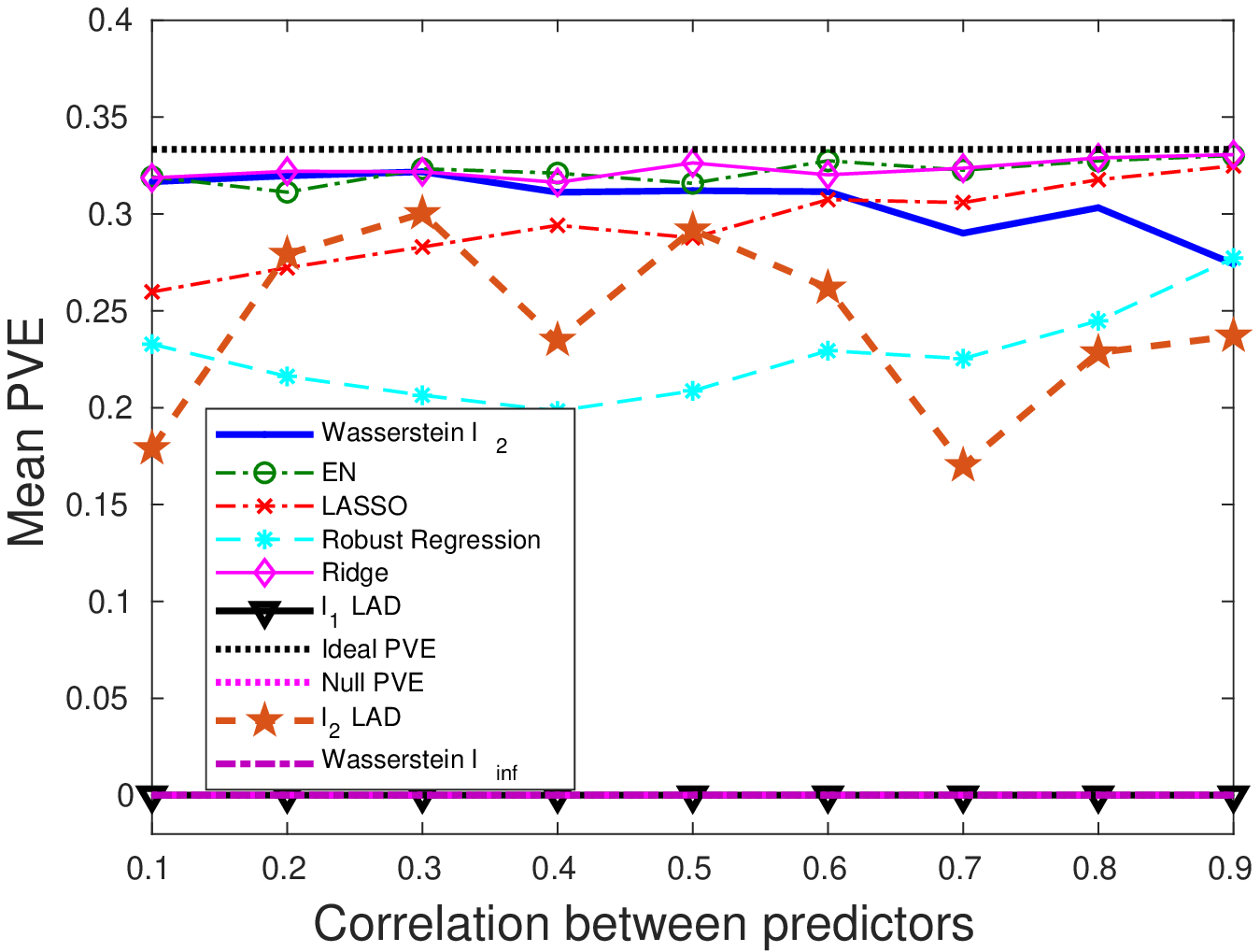}
		\caption{\small{Proportion of variance explained.}}
	\end{subfigure}
	\caption{The impact of predictor correlation on the performance metrics: dense $\bbeta^*$, outliers only in $\bx$.}
	\label{corr-2}
\end{figure}

Interestingly, we have observed that although the $\ell_1$- and $\ell_2$-regularized LAD, as well as the Wasserstein $\ell_{\infty}$ formulation, exhibit unsatisfactory performance, the Wasserstein $\ell_2$, which shares the same loss function with them, is able to achieve a comparable performance with the best among all -- EN and ridge regression (see Figs. \ref{snr-2} and \ref{corr-2}). Notably, the $\ell_2$-regularized LAD, which is just slightly different from our Wasserstein $\ell_2$ formulation, shows a much worse performance. This is because the $\ell_2$-regularized LAD implicitly assumes an infinite transportation cost along $y$, which gives zero tolerance to the variation in the response. For example, given two data points $(\bx_1, y_1)$ and $(\bx_2, y_2)$, as long as $y_1 \neq y_2$, the distance between them is infinity. Therefore, a reasonable amount of fluctuation, caused by the intrinsic randomness of $y$, would be overly exaggerated by the underlying metric used by the $\ell_2$-regularized LAD. In contrast, our Wasserstein approach uses a proper notion of norm to evaluate the distance in the $(\bx, y)$ space and is able to effectively distinguish abnormally high variations from moderate, acceptable noise.

It is also worth noting that the formulations with the AD loss, e.g., $\ell_2$- and $\ell_1$-regularized LAD, and the Wasserstein $\ell_{\infty}$, perform worse than the approaches with the SR loss. One reasonable explanation is that the AD loss, introduced primarily for hedging against large perturbations in $y$, is less useful when the noise in $y$ is moderate, in which case the sensitivity to response noise is needed. Although the AD loss is not a wise choice, penalizing the extended coefficient vector $(-\bbeta, 1)$ seems to make up, making the Wasserstein $\ell_2$ a competitive method even when the perturbations appear only in $\bx$.

\subsection{Sparse $\bbeta^*$, outliers in both $\bx$ and $y$} \label{sparsexy}
In this subsection we will experiment with a sparse $\bbeta^*$. The intercept is set to $\beta_0^* = 3$, and the coefficients for the $20$ predictors are set to $\bbeta^* = (0.05, 0, 0.006, 0, -0.007, 0, 0.008, 0, \\ \ldots, 0)$. The adversarial perturbations are present in both $\bx$ and $y$. Specifically, the distribution of outliers is characterized by:
\begin{enumerate}
	\item $\bx \sim N_{m-1} (\mathbf{0}, \bSigma) + N_{m-1} (\mathbf{0}, 0.25\mathbf{I})$;
	\item $y \sim N(\bx' \bbeta^*, \sigma^2) + 5\sigma$.
\end{enumerate}

\begin{figure}[p] 
	\begin{subfigure}{.5\textwidth}
		\centering
		\includegraphics[width=1.0\textwidth]{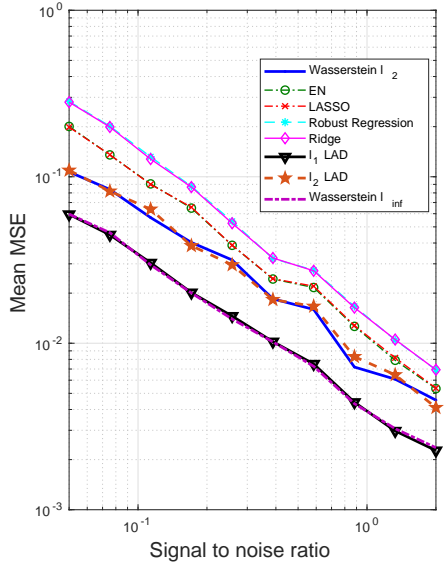}
		\caption{\small{Mean Squared Error.}}
	\end{subfigure}
	\begin{subfigure}{.5\textwidth}
		\centering
		\includegraphics[width=1.0\textwidth]{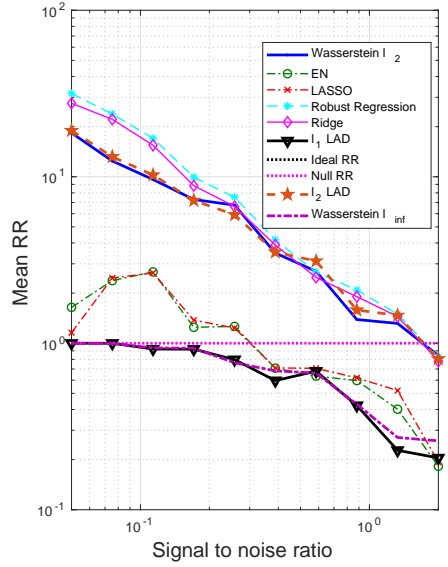}
		\caption{\small{Relative risk.}}
	\end{subfigure}
	
	\begin{subfigure}{.5\textwidth}
		\centering
		\includegraphics[width=1.0\textwidth]{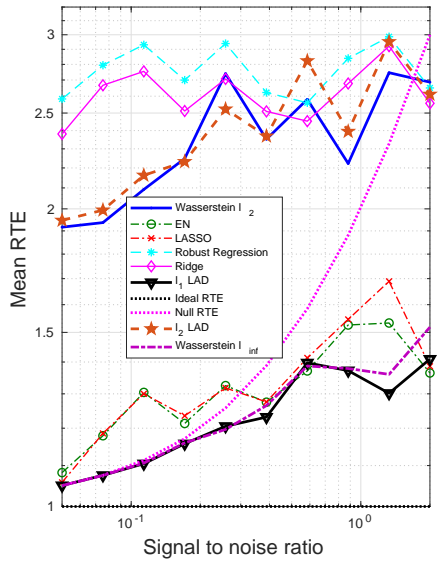}
		\caption{\small{Relative test error.}}
	\end{subfigure}%
	\begin{subfigure}{.5\textwidth}
		\centering
		\includegraphics[width=1.0\textwidth]{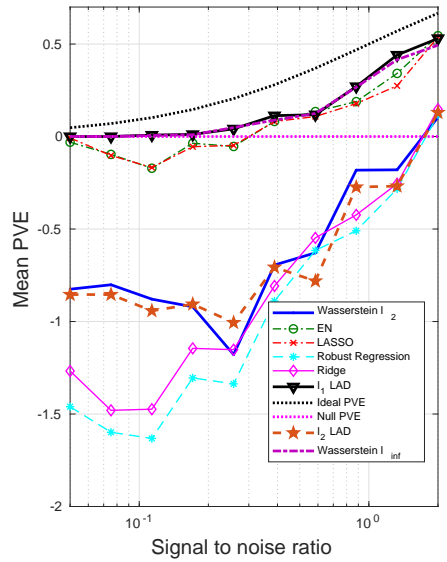}
		\caption{\small{Proportion of variance explained.}}
	\end{subfigure}
	\caption{The impact of SNR on the performance metrics: sparse $\bbeta^*$, outliers in both $\bx$ and $y$.}
	\label{snr-3}
\end{figure}

\begin{figure}[p] 
	\begin{subfigure}{.5\textwidth}
		\centering
		\includegraphics[width=1.0\textwidth]{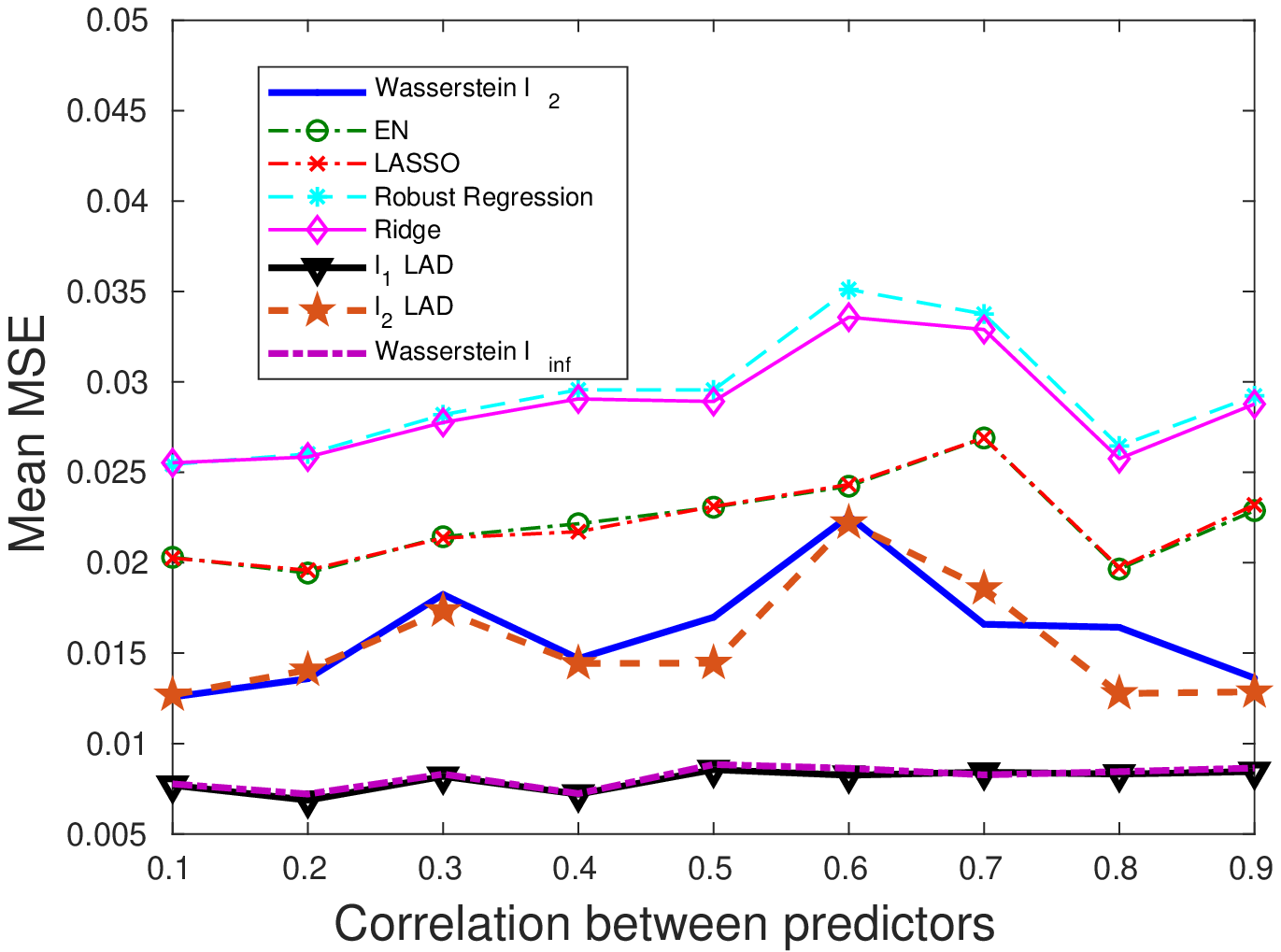}
		\caption{\small{Mean Squared Error.}}
	\end{subfigure}
	\begin{subfigure}{.5\textwidth}
		\centering
		\includegraphics[width=1.0\textwidth]{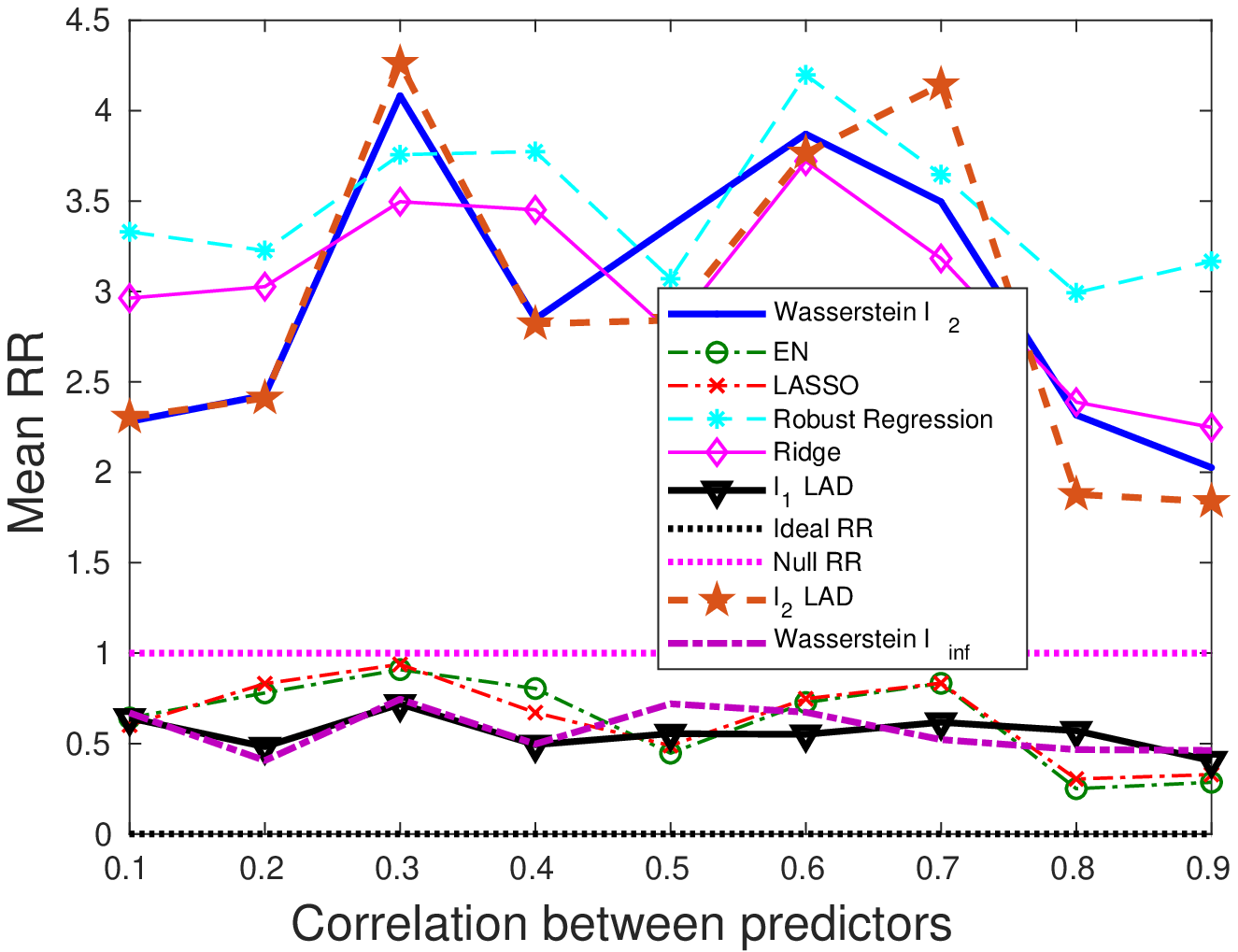}
		\caption{\small{Relative risk.}}
	\end{subfigure}
	
	\begin{subfigure}{.5\textwidth}
		\centering
		\includegraphics[width=1.0\textwidth]{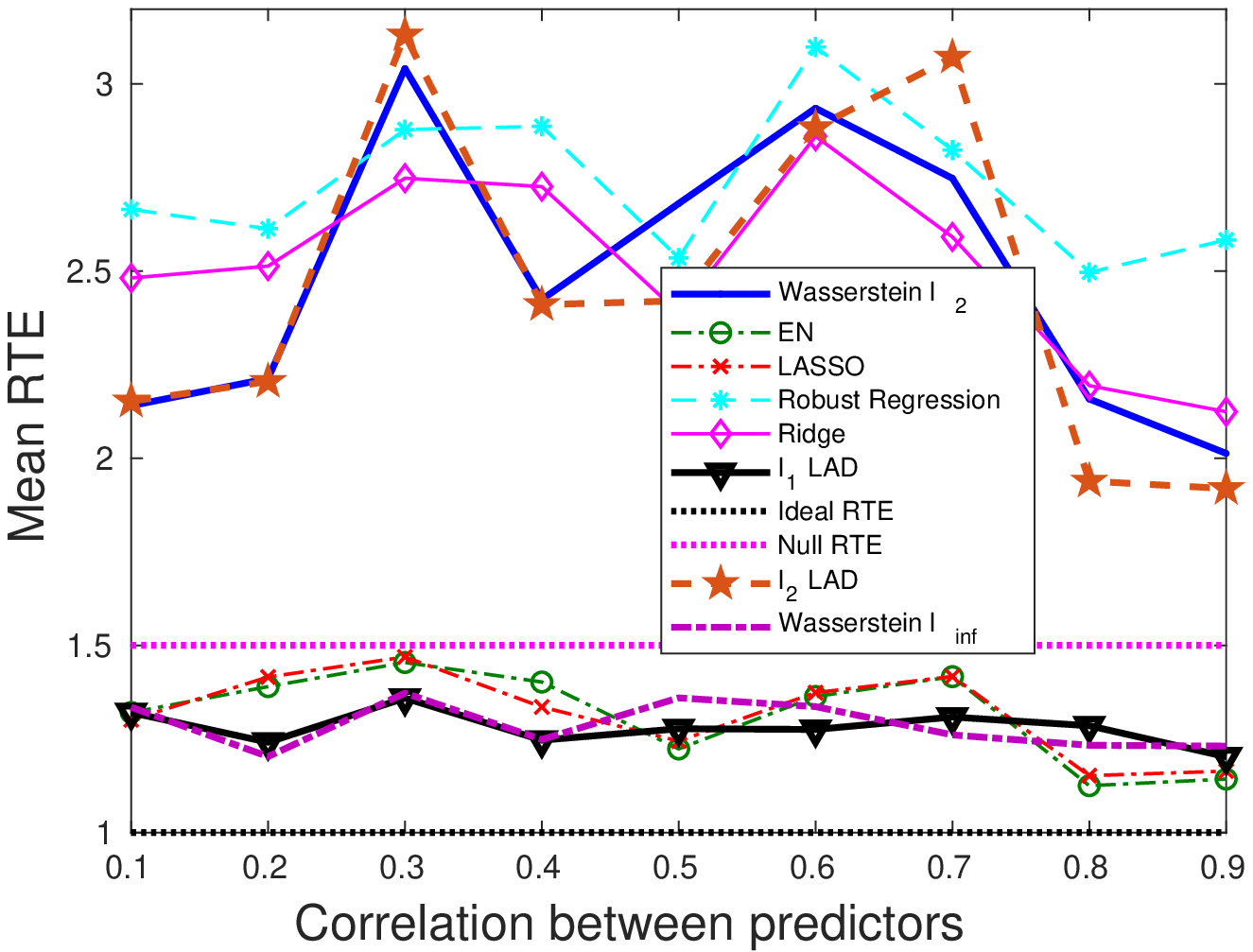}
		\caption{\small{Relative test error.}}
	\end{subfigure}%
	\begin{subfigure}{.5\textwidth}
		\centering
		\includegraphics[width=1.0\textwidth]{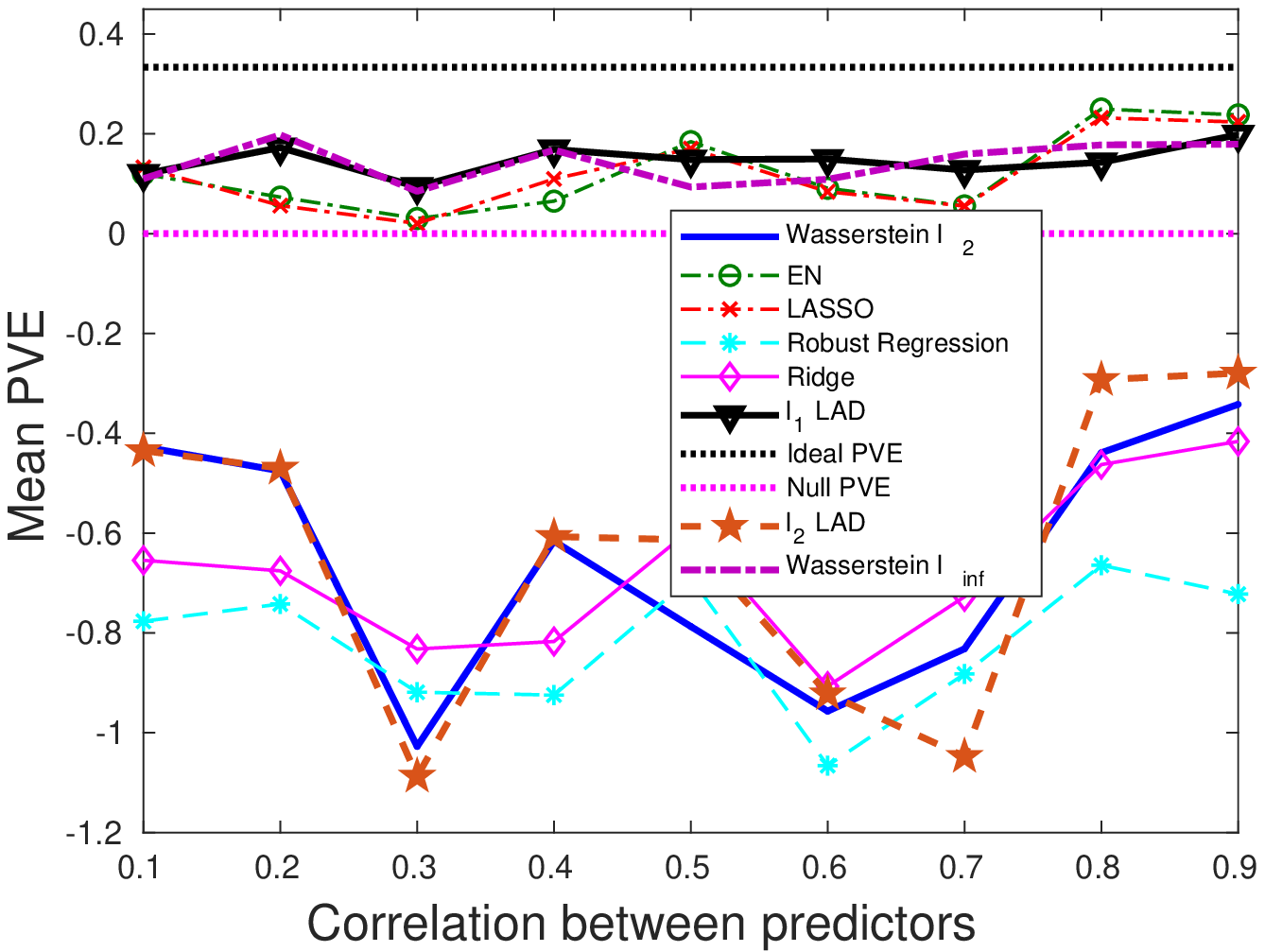}
		\caption{\small{Proportion of variance explained.}}
	\end{subfigure}
	\caption{The impact of predictor correlation on the performance metrics: sparse $\bbeta^*$, outliers in both $\bx$ and $y$.}
	\label{cor-3}
\end{figure}

Our goal is to study the impact of the sparsity of $\bbeta^*$ on the choice of the norm space for the Wasserstein metric. We know that the $\ell_1$-regularizer works better than the $\ell_2$-regularizer for sparse data, which has been validated by our results in Figs. \ref{snr-3} and \ref{cor-3}. We will see that the Wasserstein $\ell_{\infty}$ formulation significantly outperforms the Wasserstein $\ell_2$. An intuitively appealing interpretation for the sparsity inducing property of the $\ell_1$-regularizer is made available by the Wasserstein DRO framework, which we explain as follows. The sparse regression coefficient $\bbeta^*$ implies that only a few predictors are relevant to the regression model, and thus when measuring the distance in the $(\bx, y)$ space, we need a metric that only extracts the subset of relevant predictors. The $\|\cdot\|_{\infty}$, which takes only the most influential coordinate of its argument, roughly serves this purpose. Compared to the $\|\cdot\|_2$ which takes into account all the coordinates, most of which are redundant due to the sparsity assumption, $\|\cdot\|_{\infty}$ results in a better performance, and hence, the Wasserstein $\ell_{\infty}$ formulation that stems from the $\|\cdot\|_{\infty}$ distance metric on $(\bx, y)$ and induces the  $\ell_1$-regularizer is expected to outperform others.

We note that the $\ell_1$-regularized LAD achieves similar performance to ours, since replacing $\|\bbeta\|_1$ by $\|(-\bbeta, 1)\|_1$ only adds a constant term to the objective function. The generalization performance (mean MSE) of the AD loss-based formulations is consistently better than those with the SR loss, since the AD loss is less affected by large perturbations in $y$. Also note that choosing a wrong norm for the Wasserstein metric, e.g., the Wasserstein $\ell_2$, could lead to an enormous estimation error, whereas with a right norm space, we are guaranteed to outperform all others. Even when the SNR is very low, our performance is at least as good as the null estimator (see Fig. \ref{snr-3}). Although EN and LASSO achieve similar performance to ours for moderate SNR values, they have a chance of performing even worse than the null estimator when there is little signal/information to learn from.

\subsection{Sparse $\bbeta^*$, outliers only in $\bx$} \label{sparsex}
In this subsection, we will use the same sparse coefficient as in Section \ref{sparsexy}, but the perturbations are present only in $\bx$. Specifically, for outliers, their predictors and responses are drawn from the following distributions:
\begin{enumerate}
	\item $\bx \sim N_{m-1} (\mathbf{0}, \bSigma) + N_{m-1} (5\mathbf{e}, \mathbf{I})$;
	\item $y \sim N(\bx' \bbeta^*, \sigma^2)$. 
\end{enumerate}

\begin{figure}[p] 
	\begin{subfigure}{.5\textwidth}
		\centering
		\includegraphics[width=1.0\textwidth]{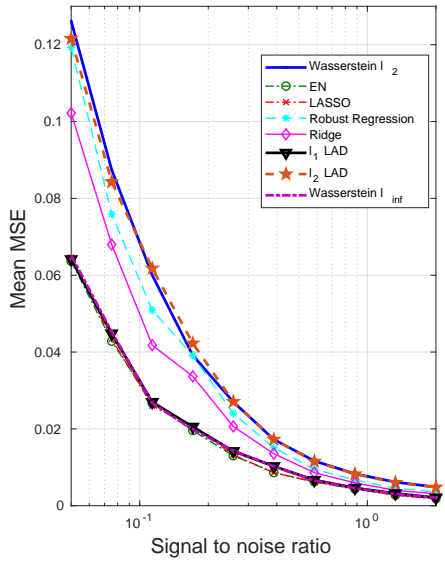}
		\caption{\small{Mean Squared Error.}}
	\end{subfigure}
	\begin{subfigure}{.5\textwidth}
		\centering
		\includegraphics[width=1.0\textwidth]{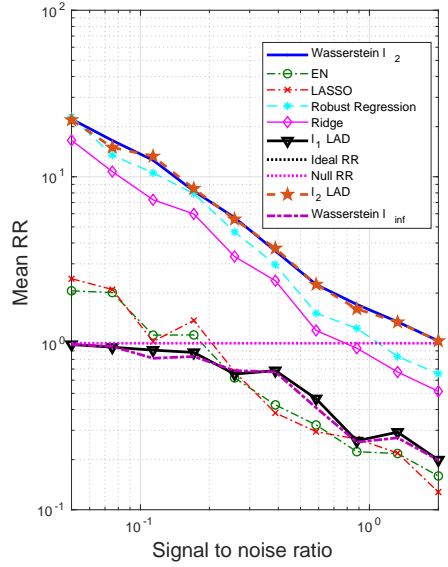}
		\caption{\small{Relative risk.}}
	\end{subfigure}
	
	\begin{subfigure}{.5\textwidth}
		\centering
		\includegraphics[width=1.0\textwidth]{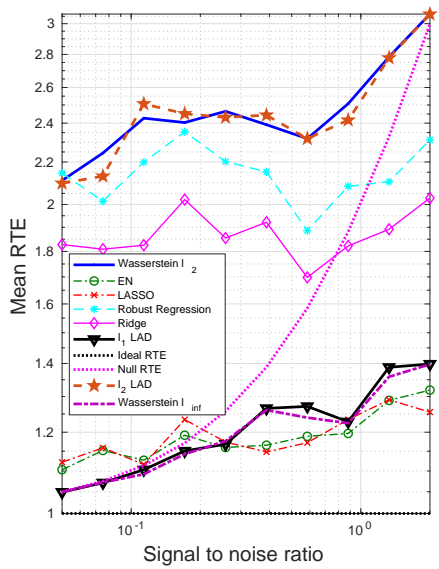}
		\caption{\small{Relative test error.}}
	\end{subfigure}%
	\begin{subfigure}{.5\textwidth}
		\centering
		\includegraphics[width=1.0\textwidth]{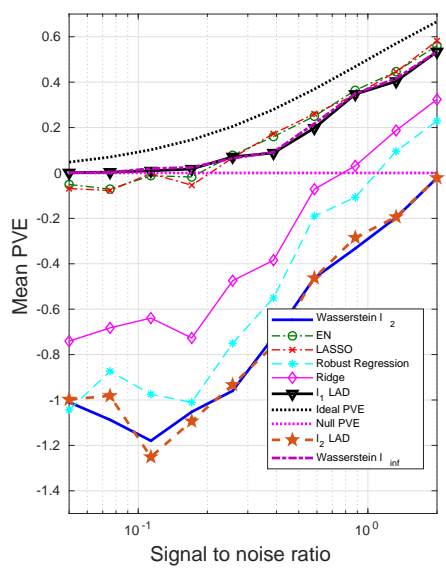}
		\caption{\small{Proportion of variance explained.}}
	\end{subfigure}
	\caption{The impact of SNR on the performance metrics: sparse $\bbeta^*$, outliers only in $\bx$.}
	\label{snr-4}
\end{figure}

Not surprisingly, the Wasserstein $\ell_{\infty}$ and the $\ell_1$-regularized LAD achieve the best performance. Notice that in Section \ref{sparsexy}, where perturbations appear in both $\bx$ and $y$, the AD loss-based formulations have smaller generalization and estimation errors than the SR loss-based formulations. When we reduce the variation in $y$, the SR loss seems superior to the AD loss, if we restrict attention to the improperly regularized ($\ell_2$-regularizer) formulations (see Fig. \ref{snr-4}). For the $\ell_1$-regularized formulations, our Wasserstein $\ell_{\infty}$ formulation, as well as the $\ell_1$-regularized LAD, is comparable with the EN and LASSO. Moreover, when there is little information to utilize (low SNR), EN and LASSO are worse than the null estimator, whereas our performance is at least as good as the null estimator.  

We summarize below our main findings from all sets of experiments we have presented:
\begin{enumerate}
	\item When a proper norm space is selected for the Wasserstein metric, the Wasserstein DRO formulation outperforms all others in terms of the generalization and estimation qualities.
	\item Penalizing the extended regression coefficient $(-\bbeta, 1)$ implicitly assumes a more reasonable distance metric on $(\bx, y)$ and thus leads to a better performance.
	\item The AD loss is remarkably superior to the SR loss when there is large variation in the response $y$.
	\item The Wasserstein DRO formulation shows a more stable estimation performance than others when the correlation between predictors is varied.
\end{enumerate}

\begin{figure}[p] 
	\begin{subfigure}{.5\textwidth}
		\centering
		\includegraphics[width=1.0\textwidth]{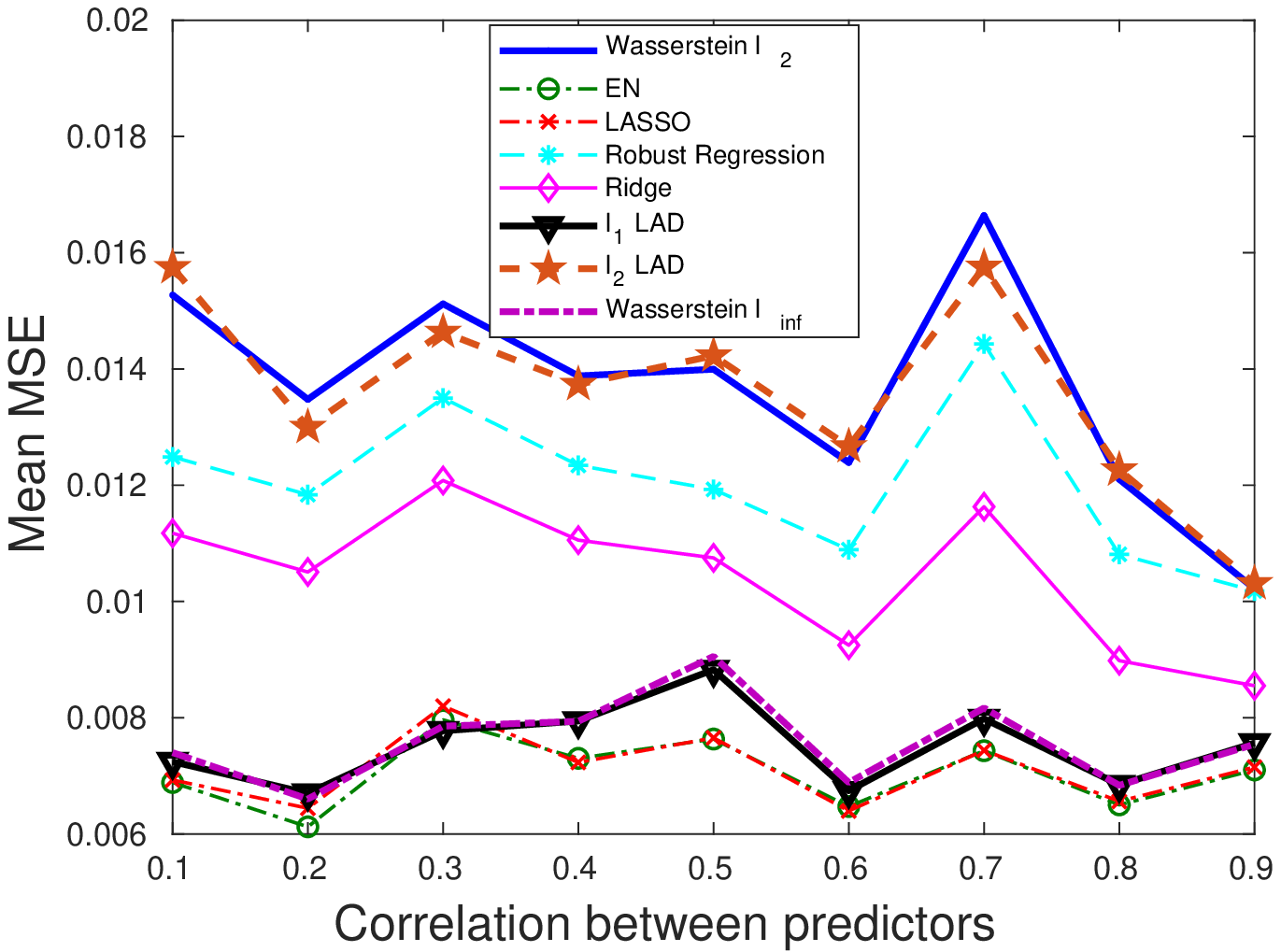}
		\caption{\small{Mean Squared Error.}}
	\end{subfigure}
	\begin{subfigure}{.5\textwidth}
		\centering
		\includegraphics[width=1.0\textwidth]{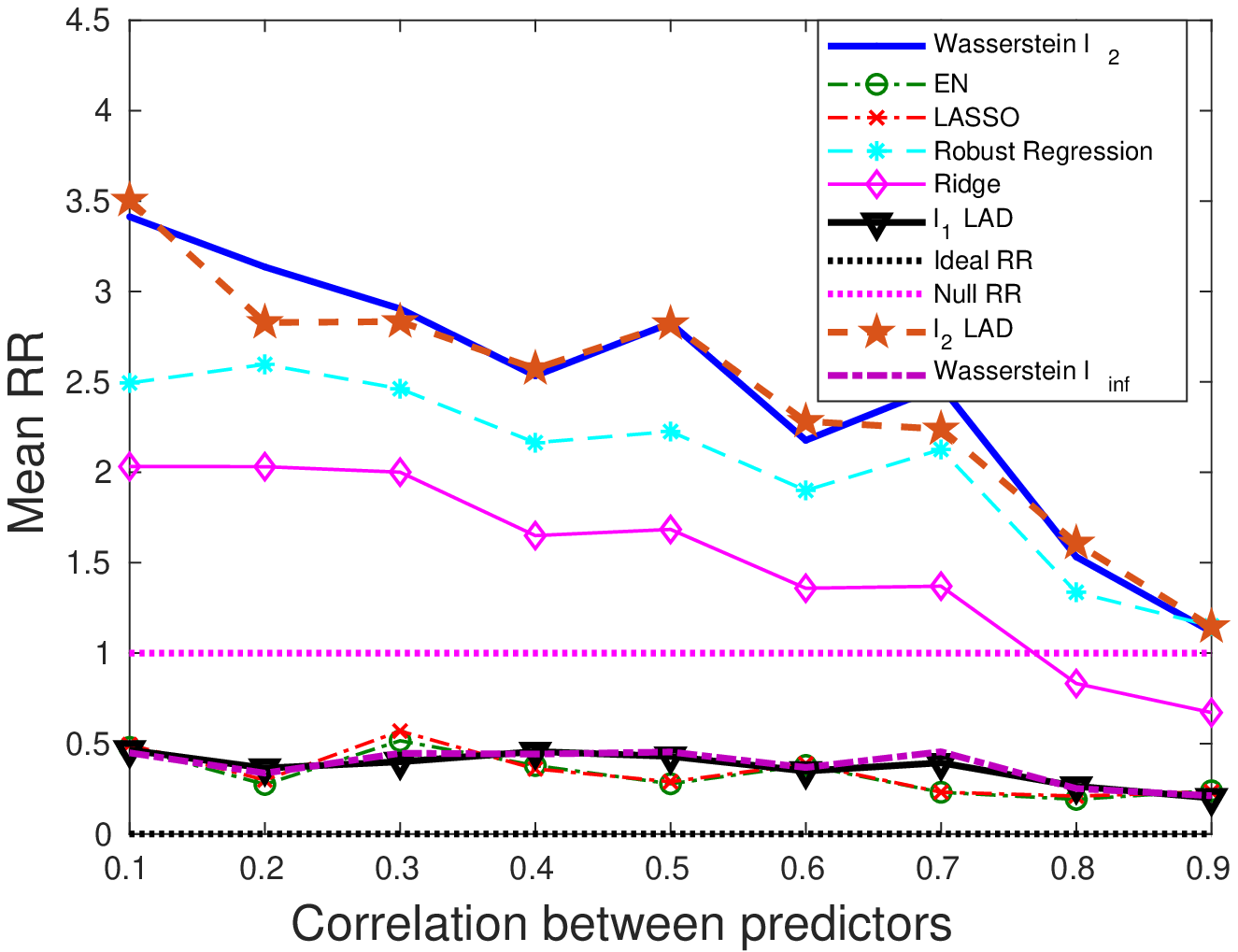}
		\caption{\small{Relative risk.}}
	\end{subfigure}
	
	\begin{subfigure}{.5\textwidth}
		\centering
		\includegraphics[width=1.0\textwidth]{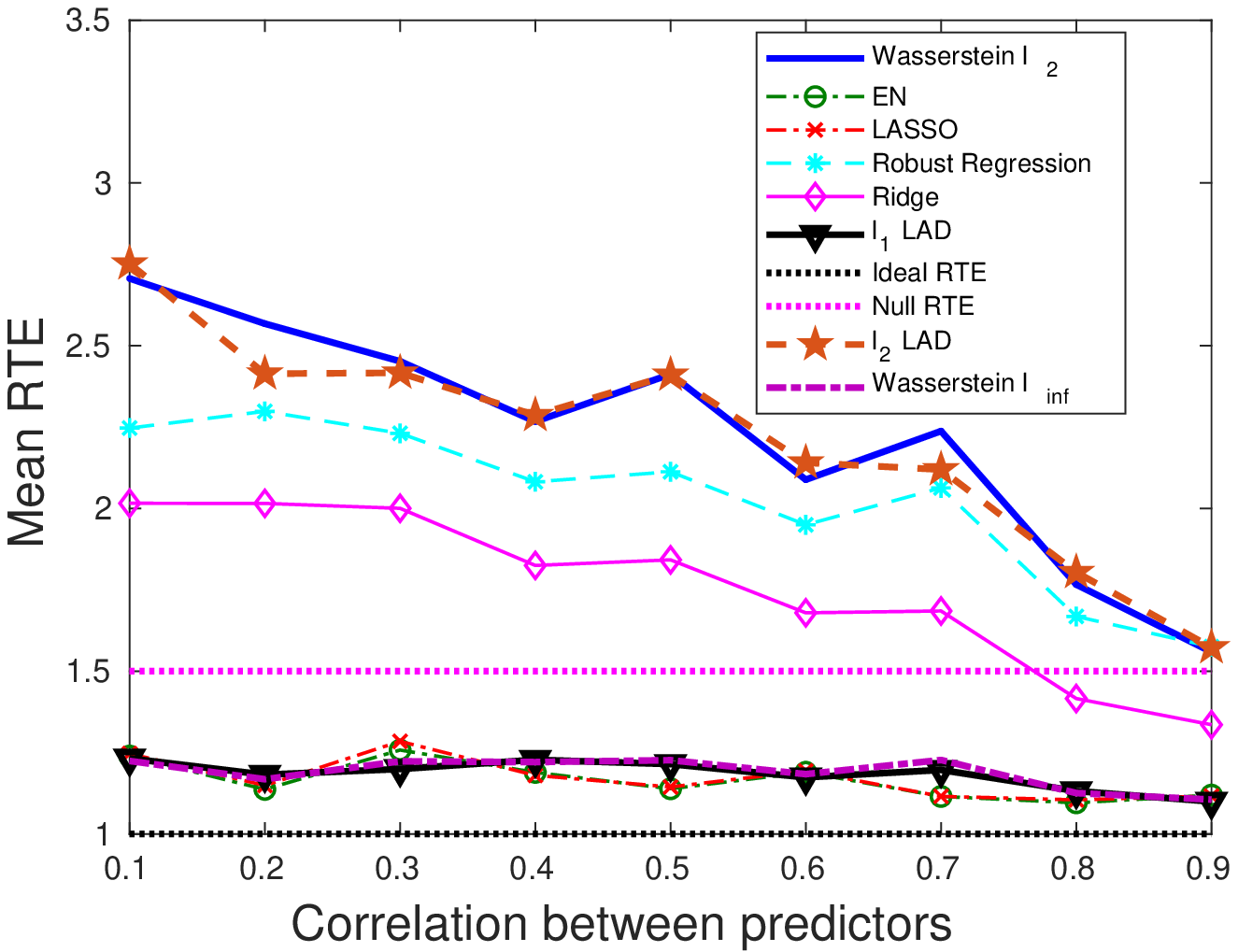}
		\caption{\small{Relative test error.}}
	\end{subfigure}%
	\begin{subfigure}{.5\textwidth}
		\centering
		\includegraphics[width=1.0\textwidth]{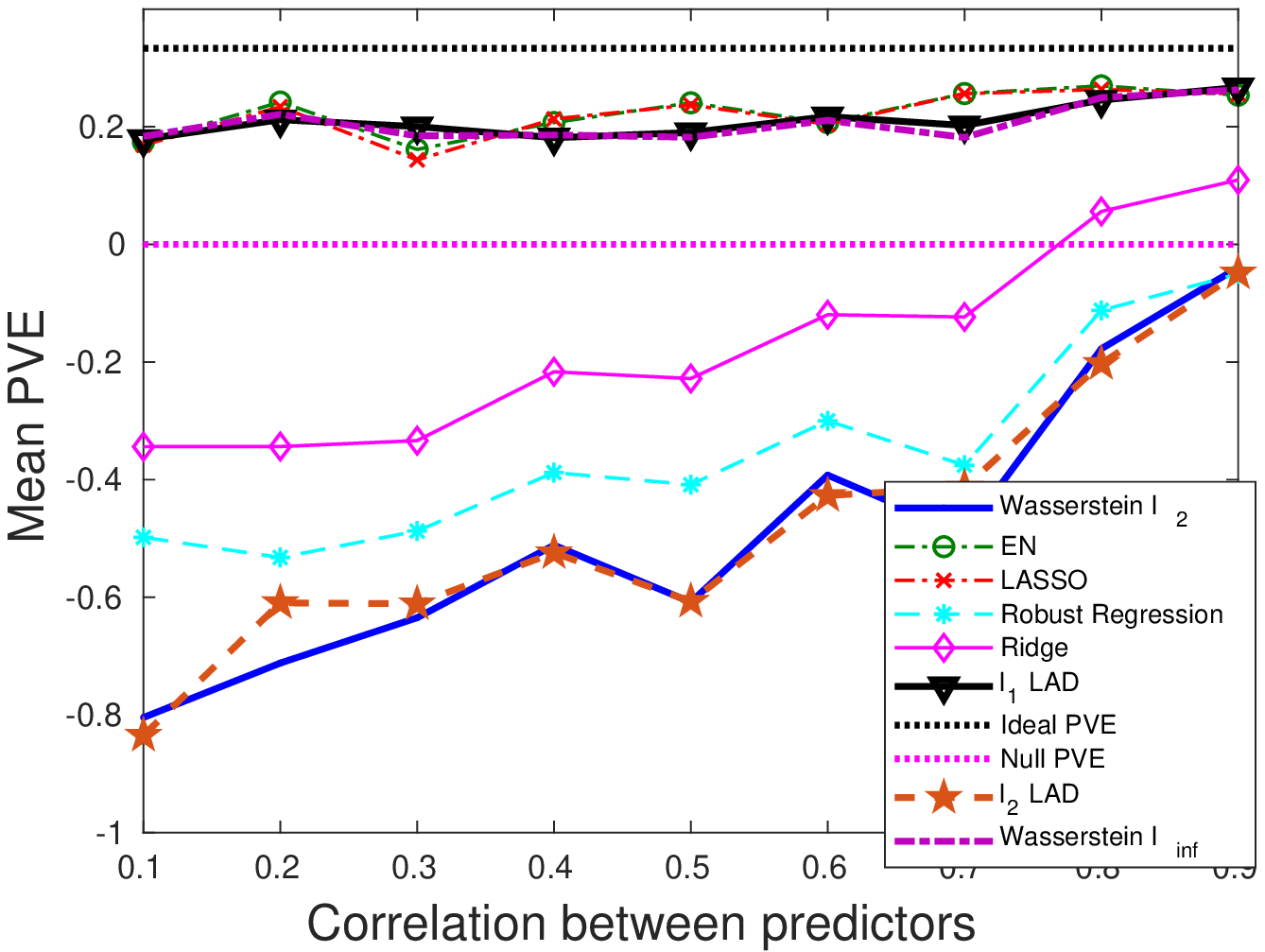}
		\caption{\small{Proportion of variance explained.}}
	\end{subfigure}
	\caption{The impact of predictor correlation on the performance metrics: sparse $\bbeta^*$, outliers only in $\bx$.}
	\label{cor-4}
\end{figure}

\subsection{An outlier detection example}
As an application, we consider an unlabeled two-class classification problem, where our goal is to identify the abnormal class of data points based on the predictor and response information using the Wasserstein formulation. We do not know a priori whether the samples are normal or abnormal, and thus classification models do not apply. The commonly used regression model for this type of problem is the M-estimation \citep{huber1964robust, huber1973robust}, against which we will compare in terms of the outlier detection capability. 

The data are generated in the same fashion as before. For clean samples, all predictors $x_1,\ldots,x_{30}$
come from a normal distribution with mean $7.5$ and standard deviation
$4.0$. The response is a linear function of the predictors with $\beta_0^*=0.3, \
\beta_1^*=\cdots=\beta_{30}^*=0.5$, plus a Gaussian distributed noise term with zero mean and standard deviation $\sigma$. The outliers concentrate in a cloud that is randomly placed in the
interior of the $\bx$-space. Specifically, their predictors are
uniformly distributed on $(u-0.125, u+0.125)$, where $u$ is a uniform
random variable on $(7.5-3\times4, 7.5+3\times4)$. The response values of the outliers
are at a $\delta_R$ distance off the regression plane.
\begin{equation*}
y=\beta_0^*+\beta_1^*x_1+\cdots+\beta_{30}^*x_{30}+\delta_R.
\end{equation*}

We will compare the performance of the Wasserstein $\ell_2$ formulation (\ref{qcp}) with the
$\ell_1$-regularized LAD and M-estimation with three cost functions
-- Huber \citep{huber1964robust,huber1973robust}, Talwar
\citep{hinich1975simple}, and Fair \citep{fair1974robust}. The performance metrics include
the {\em Receiver Operating Characteristic (ROC)} curve which
plots the true positive rate against the false positive rate, and the related {\em Area Under Curve (AUC)}. 

Notice that all the regression methods under consideration only generate an estimated regression coefficient. The
identification of outliers is based on the residual and estimated
standard deviation of the noise. Specifically,
\begin{equation*}
\text{Outlier}=
\begin{cases}
\text{YES,}&\text{if $|\text{residual}|> \text{threshold}\times\hat{\sigma}$},\\
\text{NO,}&\text{otherwise},
\end{cases}
\end{equation*} 
where $\hat{\sigma}$ is the standard deviation of residuals in the
entire training set. ROC curves are obtained through adjusting the
threshold value. 

\begin{figure}[h]
	\centering
	\includegraphics[height = 2in]{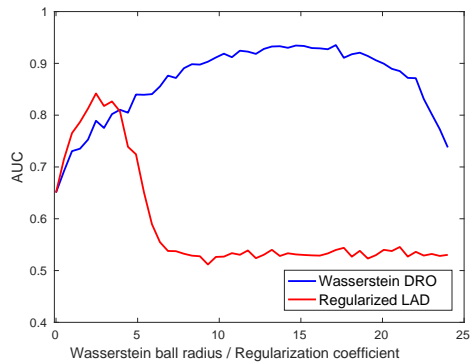}
	\caption{Out-of-sample AUC v.s.\ Wasserstein ball radius (regularization coefficient).}
	\label{radius}
\end{figure}

The regularization parameters for Wasserstein DRO and regularized LAD are tuned using a separate validation set as done in previous sections. We would like to highlight a salient advantage of our approach reflected in its robustness w.r.t. the choice of $\epsilon$. In Fig.~\ref{radius} we plot the out-of-sample AUC as the radius $\epsilon$ (regularization parameter) varies, for the $\ell_2$-induced Wasserstein DRO and the $\ell_1$-regularized LAD. For the
Wasserstein DRO curve, when
$\epsilon$ is small, the Wasserstein ball contains the true distribution
with low confidence and thus AUC is low. On the other hand, too large
$\epsilon$ makes our solution overly conservative. 
Note that the robustness of our approach,
indicated by the flatness of the Wasserstein DRO curve, constitutes
another advantage, whereas the performance of LAD dramatically
deteriorates once the regularizer deviates from the optimum. Moreover,
the maximal achievable AUC for Wasserstein DRO is significantly higher
than LAD.

\begin{figure}[h]
	\begin{subfigure}{.5\textwidth}
		\centering
		\includegraphics[width=1.0\textwidth]{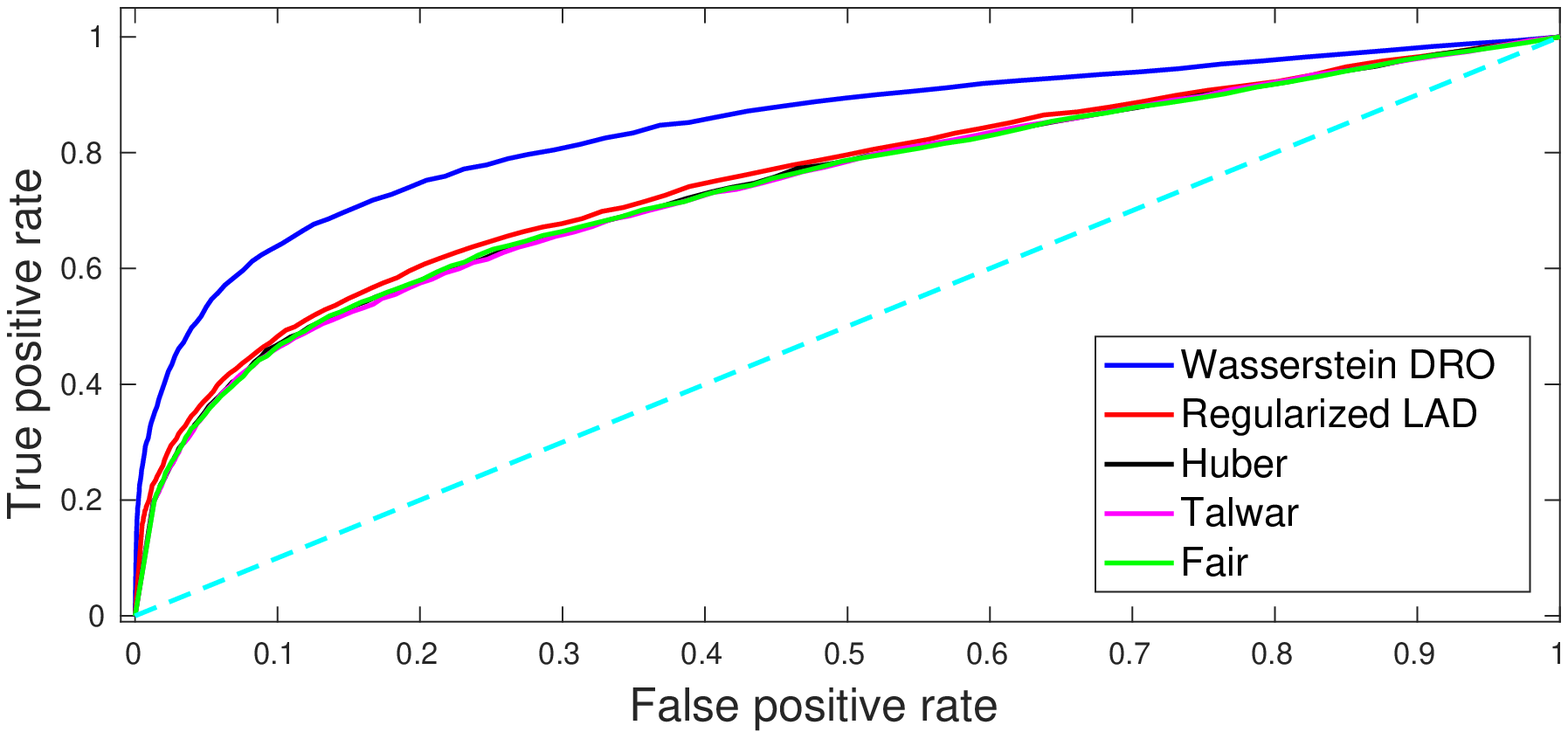}
		\caption{{\small $q=20\%, \delta_R = 3\sigma$}} 
	\end{subfigure}%
	\begin{subfigure}{.5\textwidth}
		\centering
		\includegraphics[width=1.0\textwidth]{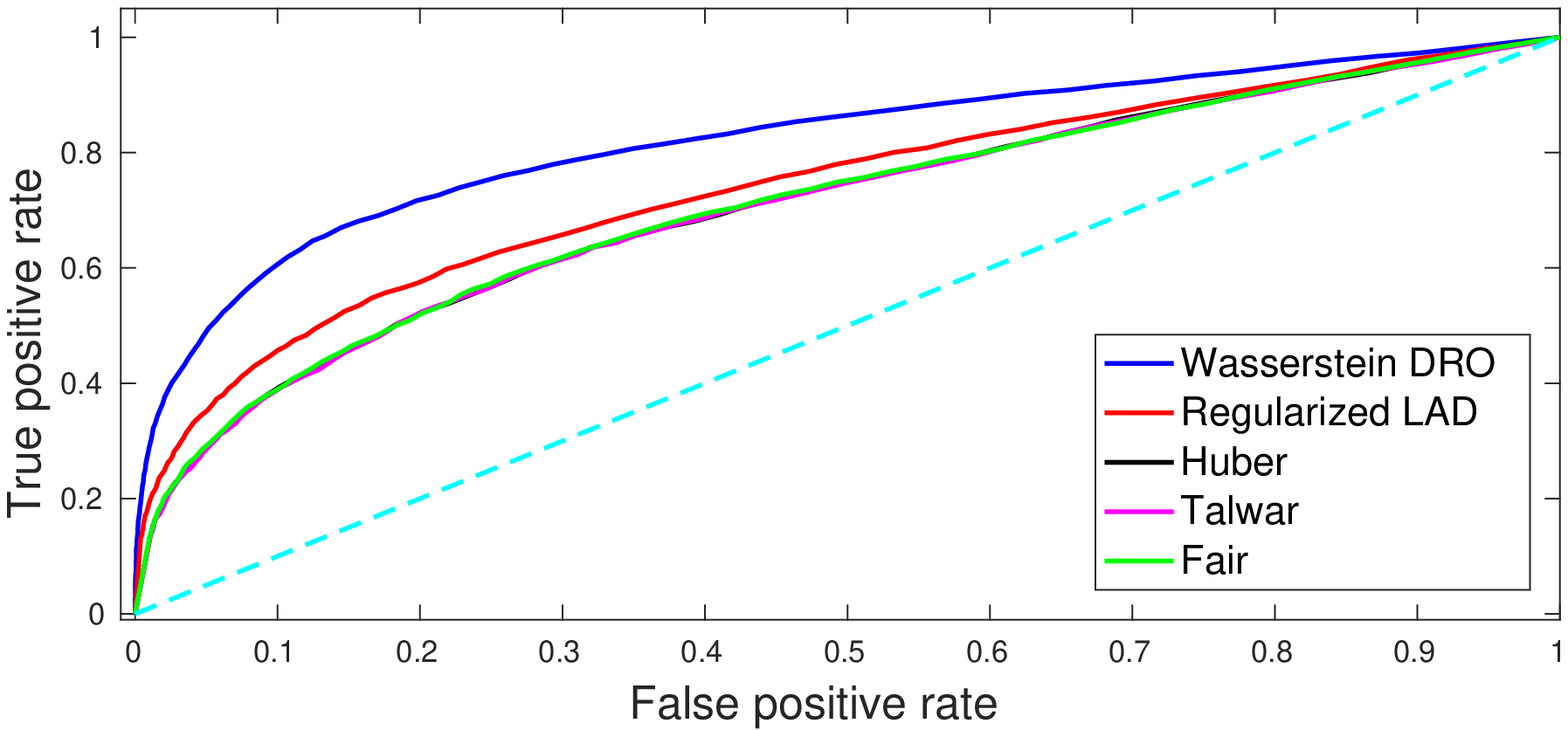}
		\caption{{\small $q=30\%, \delta_R = 3\sigma$}}
    \end{subfigure}

	\begin{subfigure}{.5\textwidth}
		\centering
		\includegraphics[width=1.0\textwidth]{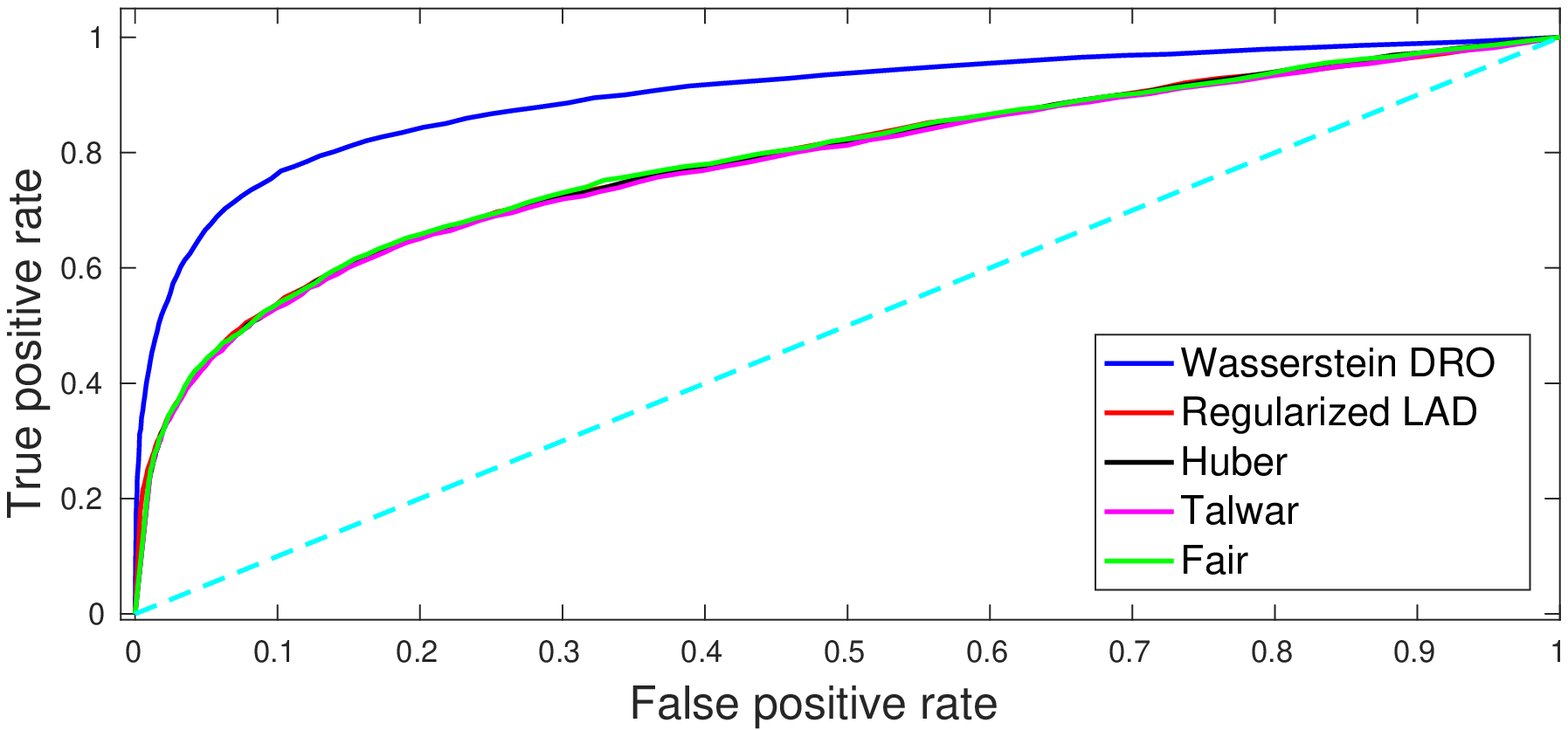}
		\caption{{\small $q=20\%, \delta_R = 4\sigma$}}
	\end{subfigure}
	\begin{subfigure}{.5\textwidth}
		\centering
		\includegraphics[width=1.0\textwidth]{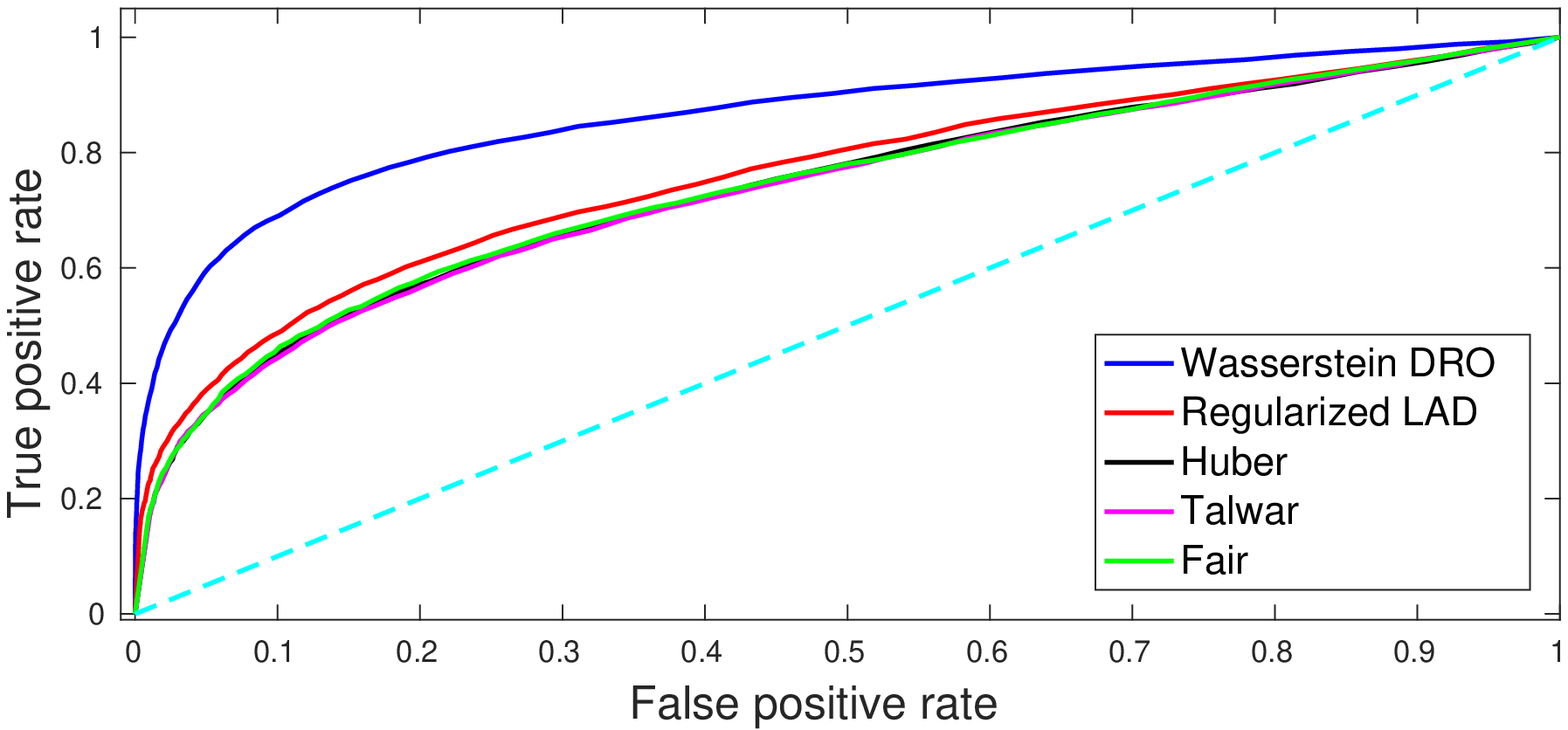}
		\caption{{\small $q=30\%, \delta_R = 4\sigma$}}
	\end{subfigure}

	\begin{subfigure}{.5\textwidth}
		\centering
		\includegraphics[width=1.0\textwidth]{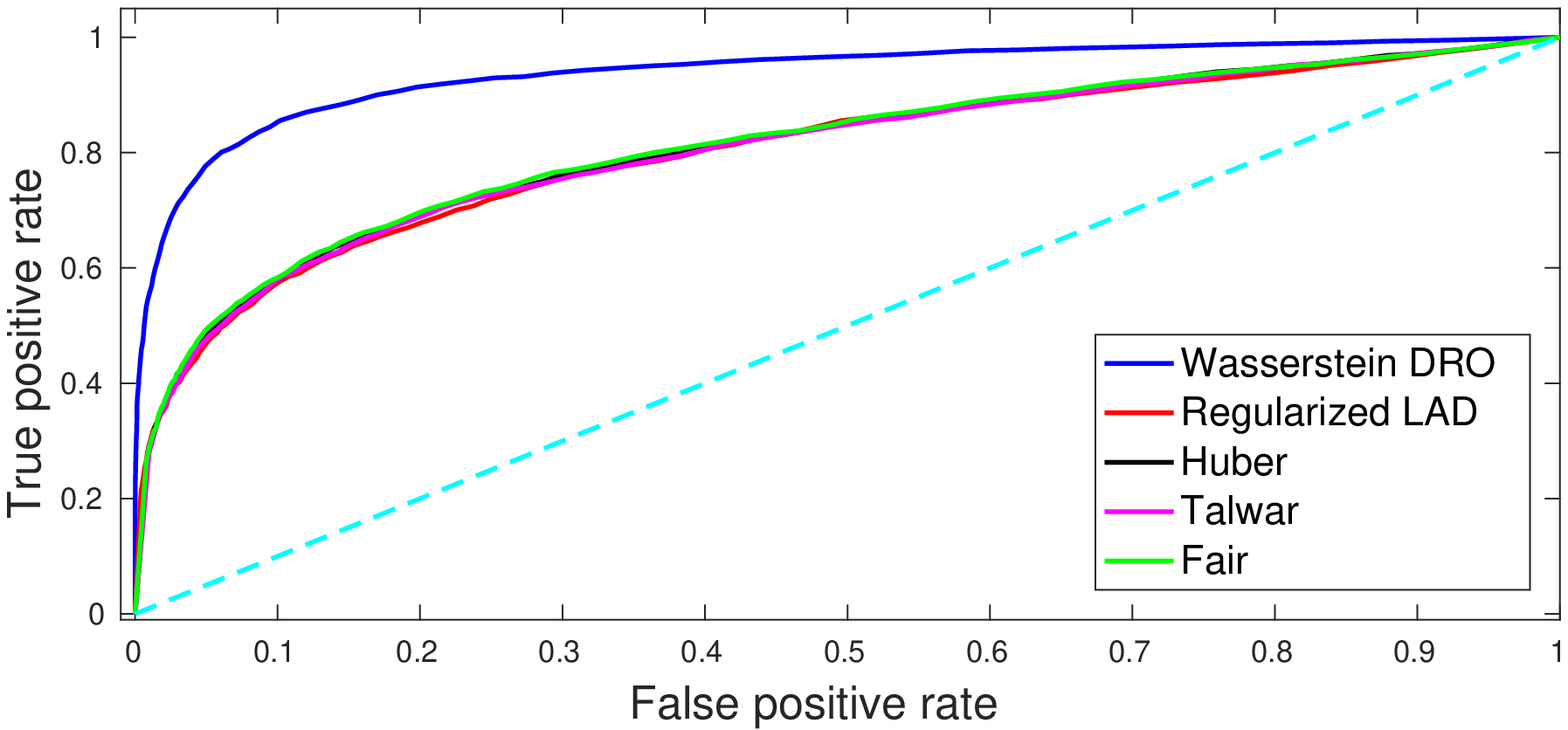}
		\caption{{\small $q=20\%, \delta_R = 5\sigma$}}
	\end{subfigure}%
	\begin{subfigure}{.5\textwidth}
		\centering
		\includegraphics[width=1.0\textwidth]{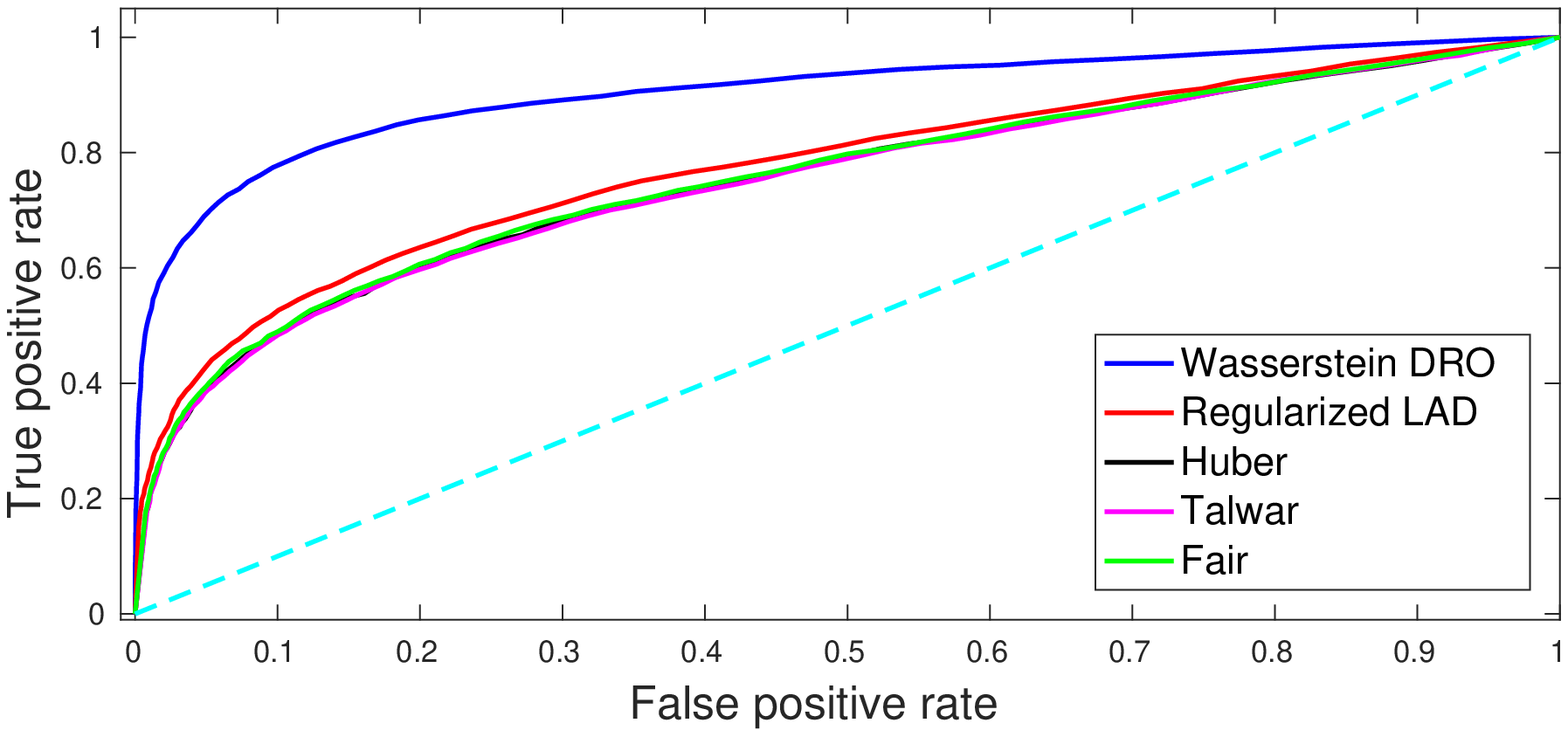}
		\caption{{\small $q=30\%, \delta_R = 5\sigma$}}
	\end{subfigure}
	\caption{ROC curves for outliers in a randomly placed cloud, $N=60, \sigma=0.5$.}
	\label{f15}
\end{figure}

In Fig. \ref{f15} we show the ROC curves for different approaches, where $q$ represents the percentage of outliers, and $\delta_R$ the outlying distance along $y$. We see that the Wasserstein DRO formulation
consistently outperforms all other approaches, with its ROC curve lying
well above others. In general, all approaches have better performance when the percentage of outliers
is lower, and the outlying distance is larger. The approaches that use the AD loss function (e.g., Wasserstein DRO and regularized LAD) tend to outperform those that adopt the SR loss (e.g., M-estimation which uses a variant of the SR loss). The superiority of our
formulation could be attributed to the AD loss function, and the distributional robustness since
we hedge against a family of plausible distributions, including the true
distribution with high confidence. By contrast, M-estimation adopts
an {\em Iteratively Reweighted Least Squares (IRLS)} procedure which
assigns weights to data points based on the residuals from previous
iterations, and then solves a weighted least squares estimation problem. With such an approach, there is a chance of exaggerating the
influence of outliers while downplaying the importance of clean
observations, especially when the initial residuals are obtained through
{\em Ordinary Least Squares (OLS)}. 

\section{Conclusions} \label{s5}
We presented a novel $\ell_1$-loss based robust learning procedure using {\em
  Distributionally Robust Optimization (DRO)} in a linear regression
framework, through which a delicate connection between the metric space on data and the regularization term has been established. The Wasserstein metric was utilized to construct the
ambiguity set and a tractable reformulation was derived. It is worth
noting that the linear law assumption does not necessarily limit the applicability
of our model. In fact, by appropriately pre-processing the data, one can often find a
roughly linear relationship between the response and transformed
explanatory variables. Our Wasserstein formulation incorporates a class
of models whose specific form depends on the norm space that the
Wasserstein metric is defined on. We provide out-of-sample generalization
guarantees, and bound the estimation bias
of the general formulation. Extensive
numerical examples demonstrate the superiority of the Wasserstein
formulation and shed light on the advantages of the $\ell_1$-loss, the implication of the regularizer, and the selection of the norm space for the Wasserstein metric. We also presented an outlier detection example as an application of this robust learning procedure. A
remarkable advantage of our approach rests in its flexibility to adjust the form of the regularizer based on the characteristics of the data.

\section*{Acknowledgments}
Research partially supported by the NSF under grants CCF-1527292,
IIS-1237022, and CNS-1645681, by the ARO under grant
W911NF-12-1-0390. and by the joint Boston University and Brigham \&
Women's Hospital program in Engineering and Radiology. We thank Jenifer
Siegelman and Vladimir Valtchinov for useful motivating discussions.
\appendix
\section{Omitted Definitions and Proofs} \label{A}
This section includes proofs for the theorems and lemmas, in the order they appear in the paper. 
\subsection{Proof of Theorem  \ref{kappa}}
\begin{proof}
	We will adopt the notation $\bz \triangleq (\bx, y), \tilde{\bbeta} \triangleq (-\bbeta, 1)$ for ease of analysis. First rewrite $\kappa(\bbeta)$ as:
	\begin{equation*}
	\kappa(\bbeta) = \sup 
	\Bigl\{\|\btheta\|_*: \sup \limits_{\bz\mid
		\bz'\tilde{\bbeta}\ge 0}\{(\btheta-\tilde{\bbeta})'\bz\} <
	\infty,\ 
	\sup\limits_{\bz\mid \bz'\tilde{\bbeta}\le 0}
	\{(\btheta+\tilde{\bbeta})'\bz\}< \infty\Bigr\}.
	\end{equation*}
	Consider now the two linear optimization problems A and B:	 
	\[ \text{Problem A:} \qquad \begin{array}{rl}
	\max & (\btheta-\tilde{\bbeta})'\bz \\
	\text{s.t.} & \bz'\tilde{\bbeta}\ge 0.
	\end{array}
	\]
	\[ \text{Problem B:} \qquad \begin{array}{rl}
	\max & (\btheta+\tilde{\bbeta})'\bz \\
	\text{s.t.} & \bz'\tilde{\bbeta}\le 0.
	\end{array}
	\]	
	Form the dual problems using dual variables $r_A$ and $r_B$,
	respectively: 
	\[ \text{Dual-A:} \qquad \begin{array}{rl}
	\min & 0\cdot r_A \\
	\text{s.t.} & \tilde{\bbeta} r_A=\btheta-\tilde{\bbeta},\\
	& r_A\le 0,
	\end{array}
	\]
	\[ \text{Dual-B:} \qquad \begin{array}{rl}
	\min & 0\cdot r_B \\
	\text{s.t.} & \tilde{\bbeta} r_B=\btheta+\tilde{\bbeta},\\
	& r_B\ge 0.
	\end{array}
	\]
	We want to find the set of $\btheta$ such that the optimal values of
	problems $A$ and $B$ are finite. Then, Dual-A and Dual-B need to have
	non-empty feasible sets, which implies the following two conditions:
	\begin{gather} 
	\exists \ r_A\le 0, \quad \text{s.t.} \quad \tilde{\bbeta} r_A=
	\btheta-\tilde{\bbeta}, \label{rA} \\
	\exists \ r_B\ge 0, \quad \text{s.t.} \quad \tilde{\bbeta} r_B=
	\btheta+\tilde{\bbeta}.  \label{rB}
	\end{gather}
	For all $i$ with $\tilde{\beta}_i\le 0$, (\ref{rA}) implies
	$\theta_i-\tilde{\beta}_i\ge 0$ and (\ref{rB}) implies $\theta_i\le
	-\tilde{\beta}_i$. On the other hand, for all $j$ with
	$\tilde{\beta}_j\ge 0$, (\ref{rA}) and (\ref{rB}) imply
	$-\tilde{\beta}_j \le \theta_j\le \tilde{\beta}_j$. It is not hard to
	conclude that:
	\[ 
	|\theta_i| \le |\tilde{\beta}_i|, \quad \forall \ i.
	\]	
	It follows, 
	\[ 
	\kappa(\bbeta)=\sup\{\|\btheta\|_*: |\theta_i| \le |\tilde{\beta}_i|,\ \forall i
	\}=
	\|\tilde{\bbeta}\|_* .
	\]
\end{proof}

\subsection{Proof of Lemma \ref{radcom}}
\begin{proof}
	Suppose that $\sigma_1, \ldots, \sigma_N$ are i.i.d.\ uniform random
	variables on $\{1, -1\}$. Then, by the definition of the Rademacher
	complexity and Lemma~\ref{l1},
	\begin{equation*}
	\begin{split}
	\scrR_N(\scrH)& = \mathbb{E}\Biggl[\sup\limits_{h \in \scrH}
	\frac{2}{N}\biggl|\sum\limits_{i=1}^N
	\sigma_ih_{\bbeta}(\bx_i, y_i)\biggr|\Biggl|(\bx_1, y_1),
	\ldots,(\bx_N, y_N)\Biggr]\\ 
	& \le \frac{2\bar{B}R}{N}\mathbb{E}\Biggl[\Biggl|\sum\limits_{i=1}^N
	\sigma_i\Biggr|\Biggr]\\ 
	& \le 
	\frac{2\bar{B}R}{N}\mathbb{E} \Biggl[\sqrt{\sum\limits_{i=1}^N\sigma_i^2}\
	\Biggr]\\  
	& = \frac{2\bar{B}R}{\sqrt{N}}.
	\end{split}
	\end{equation*}
\end{proof}

\subsection{Proof of Theorem \ref{t2}}
\begin{proof}
	We use Theorem 8 in \citet{Peter02}, setting the following
	correspondences with the notation used there: ${\cal L}(\bx, y) =
	\phi(\bx, y) = |y - \bx'\bbeta|$. This yields the
	bound (\ref{exp}) on the expected loss.  For Eq. (\ref{prob}), we
	apply Markov's inequality to obtain:
	\begin{equation*}
	\begin{split}
	\mathbb{P}\biggl(|y - \bx'\hat{\bbeta}|\ge \frac{1}{N}\sum\limits_{i=1}^N |y_i - \bx_i'\hat{\bbeta}|+\zeta\biggr)
	& \le \frac{\mathbb{E}[|y - \bx'\hat{\bbeta}|]}{\frac{1}{N}\sum_{i=1}^N |y_i - \bx_i'\hat{\bbeta}|+\zeta}\\
	&\le \frac{\frac{1}{N}\sum_{i=1}^N |y_i - \bx_i'\hat{\bbeta}|+\frac{2\bar{B}R}{\sqrt{N}}+
		\bar{B}R\sqrt{\frac{8\log(2/\delta)}{N}}}{\frac{1}{N}\sum_{i=1}^N |y_i - \bx_i'\hat{\bbeta}|+\zeta}.
	\end{split}
	\end{equation*}
\end{proof}

\subsection{Proof of Corollary \ref{samplesize1}}

\begin{proof}
	The percentage difference requirement can be translated into:
	\begin{equation*}
	\frac{2}{\sqrt{N}}+
	\sqrt{\frac{8\log(2/\delta)}{N}} \le \tau,
	\end{equation*}
	from which (\ref{N1}) can be easily derived.
\end{proof}

\subsection{Proof of Corollary \ref{samplesize2}}
\begin{proof}
	Based on Theorem 3.3, we just need the following inequality to hold:
	\begin{equation*}
	\frac{\frac{1}{N}\sum_{i=1}^N |y_i - \bx_i'\hat{\bbeta}|+\frac{2\bar{B}R}{\sqrt{N}}+
		\bar{B}R\sqrt{\frac{8\log(2/\delta)}{N}}}{\frac{1}{N}\sum_{i=1}^N |y_i - \bx_i'\hat{\bbeta}|+\gamma\bar{B}R}\le \tau,
	\end{equation*}
	which is equivalent to:
	\begin{equation} \label{A1}
	\frac{\gamma\bar{B}R-\frac{2\bar{B}R}{\sqrt{N}}-
		\bar{B}R\sqrt{\frac{8\log(2/\delta)}{N}}}{\frac{1}{N}\sum_{i=1}^N |y_i - \bx_i'\hat{\bbeta}|+\gamma\bar{B}R}\ge 1-\tau. 
	\end{equation}
	We cannot obtain a lower bound for $N$ by directly solving (\ref{A1}) since $N$ appears in a summation operator. A proper relaxation to (\ref{A1}) is: 
	\begin{equation} \label{A2}
	\frac{\gamma-\frac{2}{\sqrt{N}}-
		\sqrt{\frac{8\log(2/\delta)}{N}}}{1+\gamma}\ge 1-\tau, 
	\end{equation}
	due to the fact that $\frac{1}{N}\sum_{i=1}^N
	|y_i - \bx_i'\hat{\bbeta}|\le \bar{B}R$. By solving (\ref{A2}), we
	obtain (\ref{N2}).
\end{proof}

\subsection{Sub-Gaussian Random Variables and Gaussian Width}

\begin{defi}[Sub-Gaussian random variable]
	A random variable $z$ is sub-Gaussian if the $\psi_2$-norm defined
	below is finite, i.e.,
	\begin{equation*}
	\vertiii{z}_{\psi_2}\triangleq \sup_{q \ge 1}
	\frac{\mbb{E}|z|^q}{\sqrt{q}} < +\infty. 
	\end{equation*}
\end{defi}
An equivalent property for sub-Gaussian random variables is that their
tail distribution decays as fast as a Gaussian, namely, 
\[ 
\mbb{P}(|z|\geq t) \leq 2 \exp\{-t^2/C^2\},\quad \forall t \geq 0, 
\] 
for some constant $C$. 

A random vector $\bz \in \mbb{R}^m$ is sub-Gaussian if $\bz'\bu$ is
sub-Gaussian for any $\bu \in \mbb{R}^m$. The $\psi_2$-norm of a vector
$\bz$ is defined as: 
\[
\vertiii{\bz}_{\psi_2} \triangleq \sup\limits_{\bu\in
	\mbb{S}^{m}}\vertiii{\bz'\bu}_{\psi_2},
\] 
where $\mbb{S}^m$ denotes the unit sphere in the $m$-dimensional
Euclidean space. For the properties of sub-Gaussian random
variables/vectors, please refer to the book by \citet{RV17}.
\begin{defi}[Gaussian width]
	For any set $\scrA \subseteq \mbb{R}^m$, its Gaussian width is defined as:
	\begin{equation} \label{gw}
	w(\scrA) \triangleq \mbb{E}\Bigl[\sup_{\bu \in \scrA} \bu'\bg\Bigr],
	\end{equation}
	where $\bg\sim {\cal N}(\bzero,\bI)$ is an $m$-dimensional standard
	Gaussian random vector.
\end{defi}

\subsection{Proof of Theorem \ref{mainresult}}
In all the following proofs related to Section \ref{limit}, we will adopt the notation $\bz \triangleq (\bx, y), \ \bz_i \triangleq (\bx_i, y_i), \ \tilde{\bbeta} \triangleq (-\bbeta, 1), \ \tilde{\bbeta}_{\text{est}} \triangleq (-\hat{\bbeta}, 1), \ \tilde{\bbeta}_{\text{true}} \triangleq (-\bbeta^*, 1)$ for ease of exposition.

\begin{proof}
	Since both $\hat{\bbeta}$ and $\bbeta^*$ are feasible (the latter due to
	Assumption~\ref{adm}), we have:
	\begin{equation*}
	\begin{split}
	\|\bZ'\tilde{\bbeta}_{\text{est}}\|_1 & \le \gamma_N, \\
	\|\bZ'\tilde{\bbeta}_{\text{true}}\|_1 & \le \gamma_N,
	\end{split}
	\end{equation*}
	from which we derive that $\|\bZ'(\tilde{\bbeta}_{\text{est}} - \tilde{\bbeta}_{\text{true}})\|_1\le
	2\gamma_N$. 
	Since $\hat{\bbeta}$ is an optimal solution to (\ref{infty})
	and $\bbeta^*$ a feasible solution, it follows that
	$\|\tilde{\bbeta}_{\text{est}}\|_* \le \|\tilde{\bbeta}_{\text{true}}\|_*$. This implies that
	$\bnu=\tilde{\bbeta}_{\text{est}} - \tilde{\bbeta}_{\text{true}}$ satisfies the condition
	$\|\tilde{\bbeta}_{\text{true}}+\bv\|_* \le\|\tilde{\bbeta}_{\text{true}}\|_*$ included in the
	definition of $\scrA(\bbeta^*)$ and, furthermore, $(\tilde{\bbeta}_{\text{est}} -
	\tilde{\bbeta}_{\text{true}})/\|\tilde{\bbeta}_{\text{est}} - \tilde{\bbeta}_{\text{true}}\|_2 \in \scrA(\bbeta^*)$. Together
	with Assumption~\ref{RE}, this yields
	\begin{equation} \label{re1}
	(\tilde{\bbeta}_{\text{est}}-\tilde{\bbeta}_{\text{true}})'\bZ \bZ'(\tilde{\bbeta}_{\text{est}}-\tilde{\bbeta}_{\text{true}})\ge
	\underline{\alpha}\|\tilde{\bbeta}_{\text{est}}-\tilde{\bbeta}_{\text{true}}\|_2^2. 
	\end{equation}
	On the other hand, from the Cauchy-Schwarz inequality:
	\begin{equation} \label{re2}
	\begin{split}
	\quad \  (\tilde{\bbeta}_{\text{est}}-\tilde{\bbeta}_{\text{true}})'\bZ \bZ'(\tilde{\bbeta}_{\text{est}}-\tilde{\bbeta}_{\text{true}}) 
	& \le \|\bZ'(\tilde{\bbeta}_{\text{est}}-\tilde{\bbeta}_{\text{true}})\|_1\|\bZ'(\tilde{\bbeta}_{\text{est}}-\tilde{\bbeta}_{\text{true}})\|_\infty\\
	& \le 2\gamma_N \max_i |\bz_i'(\tilde{\bbeta}_{\text{est}}-\tilde{\bbeta}_{\text{true}})|\\
	& \le 2\gamma_N \max_i \|\tilde{\bbeta}_{\text{est}}-\tilde{\bbeta}_{\text{true}}\|_* \|\bz_i\|\\
	& \le 2R\gamma_N \|\tilde{\bbeta}_{\text{est}}-\tilde{\bbeta}_{\text{true}}\|_*.
	\end{split}
	\end{equation}
	Combining (\ref{re1}) and (\ref{re2}), we have:
	\begin{equation*}
	\begin{split}
	\|\hat{\bbeta} - \bbeta^*\|_2 & = \|\tilde{\bbeta}_{\text{est}}-\tilde{\bbeta}_{\text{true}}\|_2\\
	 & \le
	\frac{2R\gamma_N}{\underline{\alpha}}\frac
	{\|\tilde{\bbeta}_{\text{est}}-\tilde{\bbeta}_{\text{true}}\|_*}{\|\tilde{\bbeta}_{\text{est}}-\tilde{\bbeta}_{\text{true}}\|_2}\\ 
	& \le \frac{2R\gamma_N}{\underline{\alpha}}\Psi(\bbeta^*), 
	\end{split}
	\end{equation*}
	where the last step follows from the fact that $(\tilde{\bbeta}_{\text{est}} -
	\tilde{\bbeta}_{\text{true}})/\|\tilde{\bbeta}_{\text{est}} - \tilde{\bbeta}_{\text{true}}\|_2 \in \scrA(\bbeta^*)$.
\end{proof}
	
\subsection{Proof of Lemma \ref{alphalem}}
\begin{proof}
Define $\hat{\bGamma}=\frac{1}{N}\sum_{i=1}^N\bz_i\bz_i'$. Consider the
set of functions $\scrF=\{f_{\bw}(\bz)=\bz'\bGamma^{-1/2}\bw|\bw\in
\scrA_{\bGamma}\}$. Then, for any $f_{\bw}\in \scrF$, 
\begin{equation*}
\begin{split}
\mathbb{E}[f_{\bw}^2]&
 =\mathbb{E}[\bw'\bGamma^{-1/2}\bz\bz'\bGamma^{-1/2}\bw]\\ 
& = \bw'\bGamma^{-1/2}\mathbb{E}[\bz\bz']\bGamma^{-1/2}\bw\\
& = \bw'\bw\\
& = 1,
\end{split}
\end{equation*}
where we used $\bGamma=\mathbb{E}[\bz\bz']$ and the fact that $\bw\in
\scrA_{\bGamma}$. 	

For any $f_{\bw}\in \scrF$ we have
\begin{align*}
\vertiii{f_{\bw}}_{\psi_2} & =\vertiii{\bz'\bGamma^{-1/2}\bw}_{\psi_2}\\
& = \vertiii{\bz'\bGamma^{-1/2}\bw}_{\psi_2}
\frac{\|\bGamma^{-1/2}\bw\|_2}{\|\bGamma^{-1/2}\bw\|_2}\\ 
& = \vertiii{\bz' \frac{\bGamma^{-1/2}\bw}{\|\bGamma^{-1/2}\bw\|_2}}_{\psi_2}
\|\bGamma^{-1/2}\bw\|_2\\
& \le \mu \sqrt{\bw'\bGamma^{-1}\bw}\\
& \le \mu \sqrt{\frac{1}{\lambda_{\text{min}}} \|\bw\|_2^2}\\
& = \mu \sqrt{\frac{1}{\lambda_{\text{min}}}} = \bar{\mu},
\end{align*}
where the first inequality used Assumption~\ref{subgaussian} and the
second inequality used Assumption~\ref{eigen}. 

Applying Theorem D from \citet{mendelson2007reconstruction}, for any
$\theta>0$ and when 
\[
\tilde{C}_1 \bar{\mu}
\gamma_2(\scrF,\vertiii{\cdot}_{\psi_2})\leq \theta \sqrt{N},
\] 
with probability at least $1-\exp(-\tilde{C}_2\theta^2 N/\bar{\mu}^4)$ we have
\begin{align}
\sup\limits_{f_{\bw}\in \scrF}\Bigl|\frac{1}{N}\sum\limits_{i=1}^Nf_{\bw}^2(\bz_i)-\mathbb{E}[f_{\bw}^2]\Bigr|
& = \sup\limits_{f_{\bw}\in
  \scrF}\Bigl|\frac{1}{N} \sum\limits_{i=1}^N\bw'
\bGamma^{-1/2}\bz_i\bz_i'\bGamma^{-1/2}\bw-1\Bigr|\notag\\ 
& = \sup\limits_{\bw\in
  \scrA_{\bGamma}}\Bigl|\bw' \bGamma^{-1/2}
\hat{\bGamma}\bGamma^{-1/2}\bw-1\Bigr|\notag \\  
& \le \theta, \label{thmDm} 
\end{align}
where $\tilde{C}_1$ is some positive constant and
$\gamma_2(\scrF,\vertiii{\cdot}_{\psi_2})$ is defined in
\citet{mendelson2007reconstruction} as a measure of the size of the set
$\scrF$ with respect to the metric $\vertiii{\cdot}_{\psi_2}$.  Using
$\theta=1/2$, and properties of
$\gamma_2(\scrF,\vertiii{\cdot}_{\psi_2})$ outlined in
\citet{chen2016alternating},
we can set $N$ to satisfy
\begin{align*}
\tilde{C}_1 \bar{\mu} \gamma_2(\scrF,\vertiii{\cdot}_{\psi_2}) & \leq   
\tilde{C}_1 \bar{\mu}^2 \gamma_2(\scrA_{\bGamma},\|\cdot\|_2)\\
& \leq \tilde{C}_1 \bar{\mu}^2 C_0 w(\scrA_{\bGamma})\\ 
& \leq \frac{1}{2} \sqrt{N}, 
\end{align*}
for some positive constant $C_0$, where we used Eq.\ (44) in
\citet{chen2016alternating}. This implies 
\[ 
N\geq C_1\bar{\mu}^4 (w(\scrA_{\bGamma}))^2
\]
for some positive constant $C_1$. Thus, for such $N$ and with
probability at least $1-\exp(-C_2N/\bar{\mu}^4)$, for some positive
constant $C_2$, (\ref{thmDm}) holds with $\theta=1/2$. This implies that
for all $\bw\in \scrA_{\bGamma}$,
\[
\Bigl|\bw' \bGamma^{-1/2} \hat{\bGamma} \bGamma^{-1/2}\bw-1\Bigr| 
\le \frac{1}{2}
\] 
or 
\[ 
\bw' \bGamma^{-1/2} \hat{\bGamma} \bGamma^{-1/2}\bw \ge \frac{1}{2} = 
\frac{1}{2}\bw'\bGamma^{-1/2}\bGamma\bGamma^{-1/2}\bw.
\]
By the definition of $\scrA_{\bGamma}$, for any $\bv \in \scrA(\bbeta^*)$,
$$\bv'\hat{\bGamma}\bv \ge \frac{1}{2}\bv'\bGamma\bv.$$
Noting that $\hat{\bGamma} = (1/N) \bZ\bZ'$ yields the desired result.
\end{proof}

\subsection{Proof of Lemma \ref{gaussianwidthlem}}
We follow the proof of Lemma 4 in \citet{chen2016alternating}, adapted
to our setting. We include all key steps for completeness.
\begin{proof}
  Recall the definition of the Gaussian width $w(\scrA_{\bGamma})$
  (cf. (\ref{gw})):
\[ 
w(\scrA_{\bGamma}) = \mbb{E}\Bigl[\sup_{\bu \in \scrA_{\bGamma}} \bu'\bg\Bigr],
\]
where $\bg\sim {\cal N}(\bzero,\bI)$.  
We have: 
\begin{equation*}
\begin{split}
\sup_{\bw\in \scrA_{\bGamma}}\bw'\bg & =\sup\limits_{\bw\in
  \scrA_{\bGamma}}\bw'\bGamma^{-1/2}\bGamma^{1/2}\bg\\
&=  \sup_{\bw\in \scrA_{\bGamma}} \| \bGamma^{-1/2} \bw\|_2 
\frac{\bw'\bGamma^{-1/2}}{\| \bGamma^{-1/2} \bw\|_2} \bGamma^{1/2}\bg \\
& \leq \sqrt{\frac{1}{\lambda_{\text{min}}}} \sup_{\bv\in
  \text{cone}(\scrA(\bbeta^*))\cap \mathbb{B}^m}\bv'\bGamma^{1/2}\bg, 
\end{split}
\end{equation*}
where $\mbb{B}^m$ is the unit ball in the $m$-dimensional Euclidean
space and the inequality used Assumption~\ref{eigen} and the fact that
$\bw' \bGamma^{-1/2}/ \| \bGamma^{-1/2} \bw\|_2 \in \mathbb{B}^m$ and
$\bw\in \scrA_{\bGamma}$.

Define $\scrT=\text{cone}(\scrA(\bbeta^*))\cap \mathbb{B}^m$, and
consider the stochastic process $\{S_{\bv}=\bv'\bGamma^{1/2}\bg\}_{\bv
  \in \scrT}$.  For any $\bv_1, \bv_2 \in \scrT$,
\begin{equation*}
  \begin{split}
 \vertiii{S_{\bv_1}-S_{\bv_2}}_{\psi_2} & =
 \vertiii{(\bv_1-\bv_2)'\bGamma^{1/2}\bg}_{\psi_2}\\ 
& = \|\bGamma^{1/2} (\bv_1-\bv_2) \|_2
\vertiii{\frac{(\bv_1-\bv_2)'\bGamma^{1/2}\bg}{\|\bGamma^{1/2}
    (\bv_1-\bv_2) \|_2}}_{\psi_2} \\
& \leq \|\bGamma^{1/2} (\bv_1-\bv_2) \|_2 \sup_{\bu \in \mbb{S}^m}
\vertiii{\bu'\bg}_{\psi_2} \\
& = \mu_0\|\bGamma^{1/2}(\bv_1-\bv_2)\|_2\\
& \le \mu_0\sqrt{\lambda_{\text{max}}}\|\bv_1-\bv_2\|_2, 
\end{split}
\end{equation*}
where the last step used Assumption~\ref{eigen}. 

Then, by the tail behavior of sub-Gaussian random variables (see
Hoeffding bound, Thm. 2.6.2 in \citep{RV17}), we have:
	\begin{equation*}
\mathbb{P}(|S_{\bv_1}-S_{\bv_2}|\ge \delta) \le 2
\exp\biggl(-\frac{C_{01}\delta^2}{\mu_0^2\lambda_{\text{max}}
  \|\bv_1-\bv_2\|_2^2}\biggr),  
\end{equation*}
for some positive constant $C_{01}$. 

To bound the supremum of $S_{\bv}$, we define the metric
$s(\bv_1,\bv_2)=\mu_0\sqrt{\lambda_{\text{max}}}\|\bv_1-\bv_2\|_2.$
Then, by Lemma B in \citet{chen2016alternating},
\begin{equation*}
\begin{split}
  \mathbb{E}\biggl[\sup\limits_{\bv \in
    \scrT}\bv'\bGamma^{1/2}\bg\biggr] & \le
  C_{02}\gamma_2(\scrT,s)\\
  & = C_{02}\mu_0\sqrt{\lambda_{\text{max}}}\gamma_2(\scrT,\|\cdot\|_2)\\
  & \le C_{3}\mu_0\sqrt{\lambda_{\text{max}}}w(\scrT),
\end{split}
\end{equation*}
for positive constants $C_{02}, C_3$, where $\gamma_2(\scrT, s)$ is
the $\gamma_2$-functional we referred to in the proof of Lemma
\ref{alphalem}. Since $\scrT=\text{cone}(\scrA(\bbeta^*))\cap
\mathbb{B}^m\subseteq \text{conv}(\scrA(\bbeta^*)
\cup\{\boldsymbol{0}\})$, by Lemma 2 in \citet{maurer2014inequality},
	\begin{equation*}
	\begin{split}
	w(\scrT) & \le w(\text{conv}(\scrA(\bbeta^*) \cup\{\boldsymbol{0}\}))\\
	& = w(\scrA(\bbeta^*) \cup\{\boldsymbol{0}\})\\
	& \le \max\{w(\scrA(\bbeta^*)),w(\{\boldsymbol{0}\})\}+2\sqrt{\ln 4}\\
	& \le w(\scrA(\bbeta^*))+3.
	\end{split}
	\end{equation*}
	Thus,
	\begin{equation*}
	\begin{split}
	w(\scrA_{\bGamma})&=\mathbb{E}\biggl[\sup\limits_{\bw\in \scrA_{\bGamma}}\bw'\bg\biggr]\\
	& \le \sqrt{\frac{1}{\lambda_{\text{min}}}}\mathbb{E}\biggl[\sup\limits_{\bv\in \scrT}\bv'\bGamma^{1/2}\bg\biggr]\\
	& \le C_{3}\sqrt{\frac{1}{\lambda_{\text{min}}}}\mu_0\sqrt{\lambda_{\text{max}}}w(\scrT)\\
	& \le C_{3}\mu_0\sqrt{\frac{\lambda_{\text{max}}}{\lambda_{\text{min}}}}\Bigl(w(\scrA(\bbeta^*))+3\Bigr).
	\end{split}
	\end{equation*}
\end{proof}

\subsection{Proof of Corollary \ref{alphacol}}
\begin{proof}
	Combining Lemmas \ref{alphalem} and \ref{gaussianwidthlem}, and using the fact that for any $\bv \in \scrA(\bbeta^*)$,
	$$\frac{N}{2}\bv'\bGamma\bv \ge \frac{N\lambda_{\text{min}}}{2},$$
	we can derive the desired result.
\end{proof}

\subsection{Proof of Lemma \ref{gammalem}}
\begin{proof}
By the definition of dual norm, we know that:
\begin{equation*}
\|\tilde{\bbeta}'\bZ\|_1=\sup\limits_{\bv \in
	\scrB_u}\tilde{\bbeta}'\bZ\bv=\sup\limits_{\bv \in
	\scrB_u}\sum\limits_{i=1}^Nv_i\tilde{\bbeta}'\bz_i. 
\end{equation*}
Since $v_i\tilde{\bbeta}'\bz_i, \ i=1, \ldots, N$ are independent centered
sub-Gaussian random variables, and  
$$\vertiii{v_i\tilde{\bbeta}'\bz_i}_{\psi_2}\le \mu\|v_i\tilde{\bbeta}\|_2,$$
we have that $\sum_{i=1}^Nv_i\tilde{\bbeta}'\bz_i$ is also a centered
sub-Gaussian random variable with 
\begin{equation*}
\begin{split}
\vertiii{\sum\limits_{i=1}^Nv_i\tilde{\bbeta}'\bz_i}_{\psi_2}^2 & \le
C_{03}^2\sum\limits_{i=1}^N\mu^2\|v_i\tilde{\bbeta}\|_2^2\\ 
& = C_{03}^2\mu^2\|\tilde{\bbeta}\|_2^2\|\bv\|_2^2,
\end{split}
\end{equation*} 
for a positive constant $C_{03}$. 

Consider the stochastic process $\{S_{\bv}=\tilde{\bbeta}'\bZ\bv\}_{\bv\in
	\scrB_u}$. As in the proof of Lemma~\ref{gaussianwidthlem}, 
\[
\vertiii{S_{\bv_1}-S_{\bv_2}}_{\psi_2}\le
C_{03}\mu\|\tilde{\bbeta}\|_2\|\bv_1-\bv_2\|_2.
\]
By the tail behavior of sub-Gaussian random variables \citep{RV17}, we
know:
\begin{equation*}
\mathbb{P}(|S_{\bv_1}-S_{\bv_2}|\ge \delta) \le 2
\exp\biggl(-\frac{C_{04}\delta^2}{\mu^2\|\tilde{\bbeta}\|_2^2
	\|\bv_1-\bv_2\|_2^2}\biggr),
\end{equation*}
for a positive constant $C_{04}$.

Define the metric $s(\bv_1,\bv_2)=\mu\|\tilde{\bbeta}\|_2\|\bv_1-\bv_2\|_2.$ Then,
by Lemma B in \citet{chen2016alternating},
\begin{equation*}
\mathbb{P}\biggl(\sup_{\bv_1,\bv_2 \in \scrB_u} |S_{\bv_1}-S_{\bv_2}|
\ge C_{05} \bigl(\gamma_2(\scrB_u,s)+\delta\cdot
\text{diam}(\scrB_u,s)\bigr) \biggr)
\le C_{4}\exp(-\delta^2),
\end{equation*}
for positive constants $C_{05}, C_4$. 
Also,
\begin{equation*}
\begin{split}
\gamma_2(\scrB_u,s) = \ &\mu \|\tilde{\bbeta}\|_2\gamma_2(\scrB_u,\|\cdot\|_2)
\le C_{5}\mu\|\tilde{\bbeta}\|_2w(\scrB_u),\\
\text{diam}(\scrB_u,s)& =\sup_{\bv_1,\bv_2 \in \scrB_u}s(\bv_1,\bv_2)\\
& = \mu\|\tilde{\bbeta}\|_2\sup_{\bv_1,\bv_2 \in \scrB_u}\|\bv_1-\bv_2\|_2\\
& \le 2\mu\|\tilde{\bbeta}\|_2\sup_{\bv \in \scrB_u}\|\bv\|_2\\
& = 2\mu\|\tilde{\bbeta}\|_2 \rho, 
\end{split}
\end{equation*}
for positive constants $C_5$. 
Therefore, noting that $\sup_{\bv_1,\bv_2 \in
	\scrB_u}|S_{\bv_1}-S_{\bv_2}|\ge 2 \sup_{\bv \in \scrB_u} S_{\bv}$, we obtain 
\begin{equation*}
\begin{split}
& \quad \  \mathbb{P}\biggl(\sup\limits_{\bv\in\scrB_u}S_{\bv}\ge
C_{05}\Bigl(\frac{C_{5}}{2}\mu\|\tilde{\bbeta}\|_2w(\scrB_u)+\delta\mu\|\tilde{\bbeta}\|_2\rho
\Bigr)\biggr)\\
& \le \mathbb{P}\biggl(\sup\limits_{\bv_1,\bv_2 \in \scrB_u}|S_{\bv_1}-S_{\bv_2}|\ge C_{05}\bigl(\gamma_2(\scrB_u,s)+\delta \text{diam}(\scrB_u,s)\bigr)\biggr)\\
& \le C_{4}\exp(-\delta^2).
\end{split}
\end{equation*}
Set $\delta=\frac{C_{5}w(\scrB_u)}{2\rho}$; then with probability at least $1-C_{4}\exp(-\frac{C_{5}^2(w(\scrB_u))^2}{4\rho^2})$,
\begin{equation*}
\sup\limits_{\bv\in\scrB_u}S_{\bv}\le C\mu\bar{B}_2w(\scrB_u).
\end{equation*}
The result follows.
\end{proof}

\subsection{Proof of the Result in Section \ref{s4}}
We will show that if the Wasserstein metric is defined by the following metric $s_c$:
$$s_c(\bx, y) = \|(\bx, cy)\|_{\infty},$$
then as $c \rightarrow \infty$, the corresponding Wasserstein DRO formulation becomes:
\begin{equation*}
\inf\limits_{\bbeta\in \scrB} \frac{1}{N}\sum\limits_{i=1}^N|y_i - \bx_i'\bbeta| + \epsilon\|\bbeta\|_1,
\end{equation*}
which is the $\ell_1$-regularized LAD.
\begin{proof}
	We first define a new notion of norm on $(\bx, y)$ where $\bx = (x_1, \ldots, x_{m-1})$:
	\begin{equation*}
		\|(\bx, y)\|_{\bw, p} \triangleq \|(x_1w_1, \ldots, x_{m-1} w_{m-1}, yw_m)\|_{p},
	\end{equation*}
	for some $m$-dimensional weighting vector $\bw = (w_1, \ldots, w_m)$, and $p \ge 1$. Then, $s_c(\bx, y) = \|(\bx, y)\|_{\bw, \infty}$ with $\bw = (1, \ldots, 1, c)$. To obtain the Wasserstein DRO formulation, the key is to derive the dual norm of $\|\cdot\|_{\bw, \infty}$. 
	H\"{o}lder's inequality \citep{rogers1888extension} will be used for the derivation. We state it below for convenience.
	\begin{theorem} [H\"{o}lder's inequality] \label{holder}
		Suppose we have two scalars $p, q >1$ and $1/p + 1/q =1$. For any two vectors $\ba = (a_1, \ldots, a_n)$ and $\bb = (b_1, \ldots, b_n)$, the following holds.
		\begin{equation*}
		\sum_{i=1}^n |a_i b_i| \le \|\ba\|_p \|\bb\|_q.
		\end{equation*}
	\end{theorem}
	We will use the notation $\bz \triangleq (\bx, y)$. Based on the definition of dual norm, we are interested in solving the following optimization problem for $\tilde{\bbeta} \in \mbb{R}^{m}$:
	\begin{equation} \label{dualnorm2}
	\begin{aligned}
	\max\limits_{\bz} & \quad \bz' \tilde{\bbeta} \\
	\text{s.t.} & \quad \|\bz\|_{\bw, \infty} \le 1.
	\end{aligned}
	\end{equation}
	The optimal value of problem (\ref{dualnorm2}), which is a function of $\tilde{\bbeta}$, gives the dual norm evaluated at $\tilde{\bbeta}$. Using H\"{o}lder's inequality, we can write
	\begin{equation*}
	\bz'\tilde{\bbeta} = \sum_{i=1}^m (w_i z_i)\Bigl(\frac{1}{w_i}\tilde{\beta}_i\Bigr) \le \|\bz\|_{\bw, \infty} \|\tilde{\bbeta}\|_{\bw^{-1}, 1} \le \|\tilde{\bbeta}\|_{\bw^{-1}, 1},
	\end{equation*} 
	where $\bw^{-1} \triangleq (\frac{1}{w_1}, \ldots, \frac{1}{w_m})$.
	The last inequality is due to the constraint $\|\bz\|_{\bw, \infty} \le 1$. It follows that the dual norm of $\|\cdot\|_{\bw, \infty}$ is just $\|\cdot\|_{\bw^{-1}, 1}$. Back to our problem setting, using $\bw = (1, \ldots, 1, c)$, and evaluating the dual norm at $(-\bbeta, 1)$, we have the following Wasserstein DRO formulation as $c \rightarrow \infty$:
	\begin{equation*}
	\inf\limits_{\bbeta\in \scrB} \frac{1}{N}\sum\limits_{i=1}^N|y_i - \bx_i'\bbeta| + \epsilon\|(-\bbeta, 1)\|_{\bw^{-1}, 1} = \inf\limits_{\bbeta\in \scrB} \frac{1}{N}\sum\limits_{i=1}^N|y_i - \bx_i'\bbeta| + \epsilon\|\bbeta\|_1.
	\end{equation*}
\end{proof}

\section*{References}



\end{document}